\documentclass{article} 
\usepackage{iclr2021_conference,times}

\usepackage{amsmath,amsfonts,bm}

\def\eqref#1{equation~\ref{#1}}

\def\1{\bm{1}}

\DeclareMathAlphabet{\mathsfit}{\encodingdefault}{\sfdefault}{m}{sl}
\SetMathAlphabet{\mathsfit}{bold}{\encodingdefault}{\sfdefault}{bx}{n}

\iclrfinalcopy

\usepackage{hyperref}
\usepackage{url}

\usepackage{booktabs}
\usepackage{subcaption}
\usepackage{graphicx}
\usepackage{wrapfig}
\usepackage{algorithm}
\usepackage{algorithmic}

\title{Rank the Episodes: \\ A Simple Approach for Exploration in \\ Procedurally-Generated Environments}

\author{Daochen Zha$^1$, Wenye Ma$^2$, Lei Yuan$^2$, Xia Hu$^1$, Ji Liu$^2$ \\
$^1$Department of Computer Science and Engineering, Texas A\&M University\\
$^2$AI Platform, Kwai Inc.\\
\texttt{\{daochen.zha,xiahu\}@tamu.edu}\\
\texttt{\{mawenye,leiyuan03\}@kuaishou.com}, \texttt{jiliu@kwai.com}
}

\begin{document}

\maketitle

\begin{abstract}
Exploration under sparse reward is a long-standing challenge of model-free reinforcement learning. The state-of-the-art methods address this challenge by introducing intrinsic rewards to encourage exploration in novel states or uncertain environment dynamics. Unfortunately, methods based on intrinsic rewards often fall short in procedurally-generated environments, where a different environment is generated in each episode so that the agent is not likely to visit the same state more than once. Motivated by how humans distinguish good exploration behaviors by looking into the entire episode, we introduce RAPID, a simple yet effective episode-level exploration method for procedurally-generated environments. RAPID regards each episode as a whole and gives an episodic exploration score from both per-episode and long-term views. Those highly scored episodes are treated as good exploration behaviors and are stored in a small ranking buffer. The agent then imitates the episodes in the buffer to reproduce the past good exploration behaviors. We demonstrate our method on several procedurally-generated MiniGrid environments, a first-person-view 3D Maze navigation task from MiniWorld, and several sparse MuJoCo tasks. The results show that RAPID significantly outperforms the state-of-the-art intrinsic reward strategies in terms of sample efficiency and final performance. The code is available at \url{https://github.com/daochenzha/rapid}.

\end{abstract}

\section{Introduction}
\vspace{-5pt}
Deep reinforcement learning (RL) is widely applied in various domains~\citep{mnih2015human,silver2016mastering,mnih2016asynchronous,lillicrap2015continuous,andrychowicz2017hindsight,zha2019rlcard,liu2020finrl}. However, RL algorithms often require tremendous number of samples to achieve reasonable performance~\citep{hessel2018rainbow}. This sample efficiency issue becomes more pronounced in sparse reward environments, where the agent may take an extremely long time before bumping into a reward signal~\citep{riedmiller2018learning}. Thus, how to efficiently explore the environment under sparse reward remains an open challenge~\citep{osband2019deep}.

To address the above challenge, many exploration methods have been investigated and demonstrated to be effective on hard-exploration environments. One of the most popular techniques is to use intrinsic rewards to encourage exploration~\citep{schmidhuber1991possibility,oudeyer2009intrinsic,guo2016deep,zheng2018learning,du2019liir}. The key idea is to give intrinsic bonus based on uncertainty, e.g., assigning higher rewards to novel states~\citep{ostrovski2017count}, or using the prediction error of a dynamic model as the intrinsic reward~\citep{pathak2017curiosity}. While many intrinsic reward methods have demonstrated superior performance on hard-exploration environments, such as Montezuma's Revenge~\citep{burda2018exploration} and Pitfall!~\citep{badia2019never}, most of the previous studies use the same singleton environment for training and testing, i.e., the agent aims to solve the same environment in each episode. However, recent studies show that the agent trained in this way is susceptible to overfitting and may fail to generalize to even a slightly different environment~\citep{rajeswaran2017towards,zhang2018dissection}. To deal with this issue, a few procedually-generated environments are designed to test the generalization of RL, such as~\citep{beattie2016deepmind,chevalier2018minimalistic,nichol2018gotta,cote2018textworld,cobbe2019leveraging,kuttler2020nethack}, in which the agent aims to solve the same task, but a different environment is generated in each episode.

\begin{figure}[t]
  \centering
  
  \begin{subfigure}[b]{0.5\textwidth}
    \centering
    \includegraphics[width=0.95\textwidth]{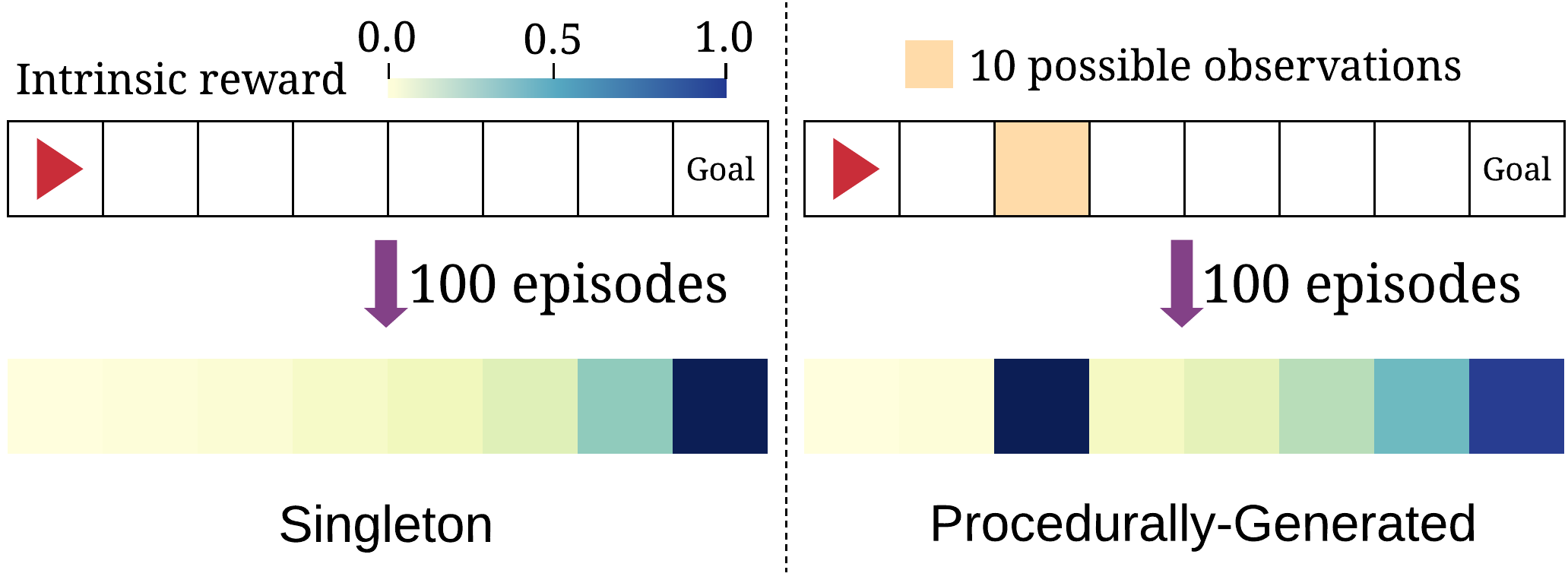}
    \caption{Count-based exploration}
    \label{fig:motivating-1}
  \end{subfigure}%
  \begin{subfigure}[b]{0.5\textwidth}
    \centering
    \includegraphics[width=0.95\textwidth]{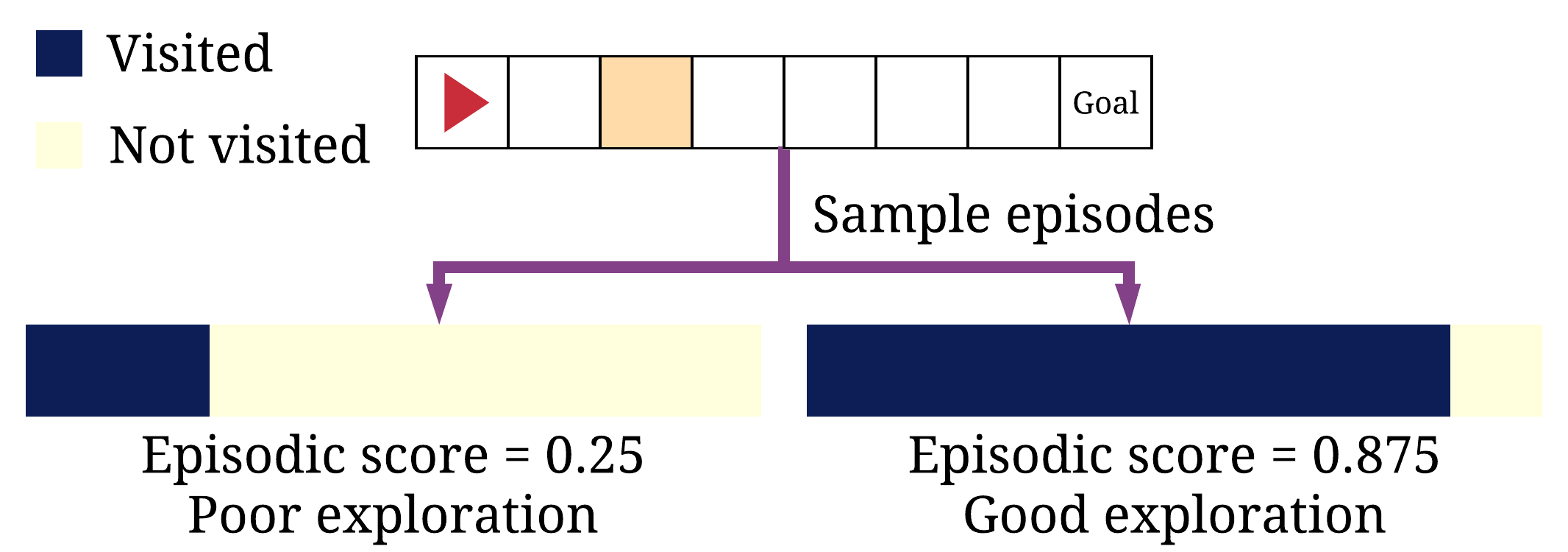}
    \caption{Episodic exploration score}
    \label{fig:motivating-2}
  \end{subfigure}%
  \caption{A motivating example of count-based exploration versus episode-level exploration score. While count-based exploration works well in singleton setting, i.e., the environment is the same in each episode, it may be brittle in procedually-generated setting. In (a), if the observation of the third block in an episode is sampled from ten possible values, the third block will have very high intrinsic reward because the agent is uncertain on the new states. This will reward the agent for exploring the third block and also its neighbors, and hence the agent may get stuck around the third block. In (b), we 
 score behaviors in episode-level with \emph{\# visited states/ \# total states}. The episodic exploration score can effectively distinguish good exploration behaviors in procedually-generated setting.}
 \label{fig:motivating}
 \vspace{-5pt}
\end{figure}

Unfortunately, encouraging exploration in procedually-generated environments is a very challenging task. Many intrinsic reward methods, such as count-based exploration and curiosity-driven exploration, often fall short in procedually-generated environments~\citep{Raileanu2020RIDE,campero2020learning}. Figure~\ref{fig:motivating-1} shows a motivating example of why count-based exploration is less effective in procedually-generated environments. We consider a 1D grid-world, where the agent (red) needs to explore the environment to reach the goal within $7$ timesteps, that is, the agent needs to move right in all the steps. While count-based exploration assigns reasonable intrinsic rewards in singleton setting, it may generate misleading intrinsic rewards in procedually-generated setting because visiting a novel state does not necessarily mean a good exploration behavior. This issue becomes more pronounced in more challenging procedually-generated environments, where the agent is not likely to visit the same state more than once. To tackle the above challenge, this work studies how to reward good exploration behaviors that can generalize to different environments.

When a human judges how well an agent explores the environment, she often views the agent from the episode-level instead of the state-level. For instance, one can easily tell whether an agent has explored a maze well by looking into the coverage rate of the current episode, even if the current environment is different from the previous ones. In Figure~\ref{fig:motivating-2}, we can confidently tell that the episode in the right-hand-side is a good exploration behavior because the agent achieves $0.875$ coverage rate. Similarly, we can safely conclude that the episode in the left-hand-side does not explore the environment well due to its low coverage rate. Motivated by this, we hypothesize that episode-level exploration score could be a more general criterion to distinguish good exploration behaviors than state-level intrinsic rewards in procedually-generated setting.

To verify our hypothesis, we propose exploration via \textbf{Ra}nking the E\textbf{pi}so\textbf{d}es (RAPID). Specifically, we identify good exploration behaviors by assigning episodic exploration scores to past episodes. To efficiently learn the past good exploration behaviors, we use a ranking buffer to store the highly-scored episodes. The agent then learns to reproduce the past good exploration behaviors with imitation learning. We demonstrate that RAPID significantly outperforms the state-of-the-art intrinsic reward methods on several procedually-generated benchmarks. Moreover, we present extensive ablations and analysis for RAPID, showing that RAPID can well explore the environment even without extrinsic rewards and could be generally applied in tasks with continuous state/action spaces.

\vspace{-5pt}
\section{Exploration via Ranking the Episodes}
\begin{figure}[t]
  \centering
  
    \includegraphics[width=0.95\textwidth]{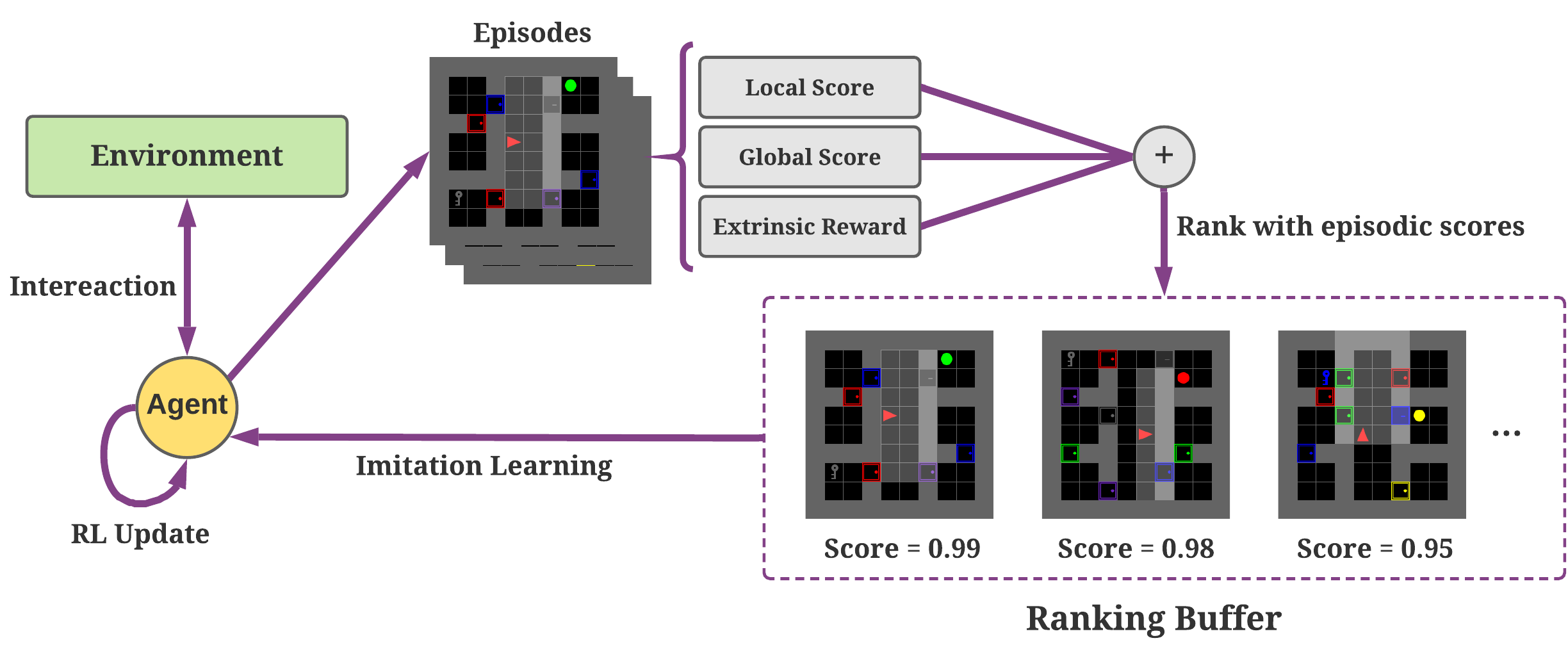}
  \vspace{-8pt}
  \caption{An overview of RAPID. The past episodes are assigned episodic exploration scores based on the local view, the global view, and the extrinsic reward. Those highly scored episodes are stored in a small ranking buffer. The agent is then encouraged to reproduce the past good exploration behaviors, i.e., the episodes in the buffer, with imitation learning.}
  \vspace{-3pt}
  \label{fig:overview}
\end{figure}

\vspace{-5pt}
An overview of RAPID is shown in Figure~\ref{fig:overview}. The key idea is to assign episodic exploration scores to past episodes and store those highly-scored episodes into a small buffer. The agent then treats the episodes in the buffer as demonstrations and learns to reproduce these episodes with imitation learning. RAPID encourages the agent to explore the environment by reproducing the past good exploration behaviors. This section introduces how we define the episodic exploration score for procedurally-generated environments and how the proposed method can be combined with state-of-the-art reinforcement learning agents. 

\vspace{-5pt}
\subsection{Episodic Exploration Score}
\vspace{-5pt}
Each episode is scored from both local and global perspectives. The local score is a per-episode view of the exploration behavior, which measures how well the current episode itself explores the environment. The global score provides a long-term and historical view of exploration, i.e., whether the current episode has explored the regions that are not well explored in the past. Additionally, we consider extrinsic reward, which is the episodic reward received from the environment. It is an external criterion of exploration since a high extrinsic reward often suggests good exploration in sparse environments. The overall episodic exploration score is obtained by the weighted sum of the above three scores. We expect that these scores can model different aspects of exploration behaviors and they are complementary in procedurally-generated environments.


\textbf{Local Score.} The intuition of the local exploration bonus is that the agent is expected to visit as many distinct states as possible in one episode. In our preliminary experiments on MiniGrid, we observe that many poor exploration episodes repeatedly visit the same state. As a result, the agent may get stuck in a well-known region but never reach out to unexplored regions. From a local view, we can quantify this exploration behavior based on the diversity of the visited states. Specifically, we define the local exploration score as
\vspace{-1pt}
\begin{equation}
    S_{\text{local}} = \frac{N_{\text{distinct}}}{N_{\text{total}}},
    \label{eqn:localdiscreate}
\end{equation}
where $N_{\text{total}}$ is the total number of states in the episode, and $N_{\text{distinct}}$ is the number of distinct states in the episode. Intuitively, optimizing the local score will encourage the agent to reach out to the unexplored regions in the current episode. In an ideal case, the episodes that never visit the same state twice will receive a local bonus of $1$. There could be various ways to extend the above definition to continuous state space. In this work, we empirically define the local score in continuous state space as the mean standard deviation of all the states in the episode:

\begin{equation}
    S_{\text{local}} = \frac{\sum_{i=1}^{l}\text{std}(s_i)}{l},
    \label{eqn:localcontinious}
\end{equation}
where $l$ is the dimension of the state, and $\text{std}(s_i)$ is the standard deviation along the $i$-dimension of the states in the episode. Intuitively, this score encourages the episodes that visit diverse states.

\begin{figure}[t]
  \centering
  
  \begin{subfigure}[b]{0.23\textwidth}
    \centering
    \includegraphics[width=0.85\textwidth]{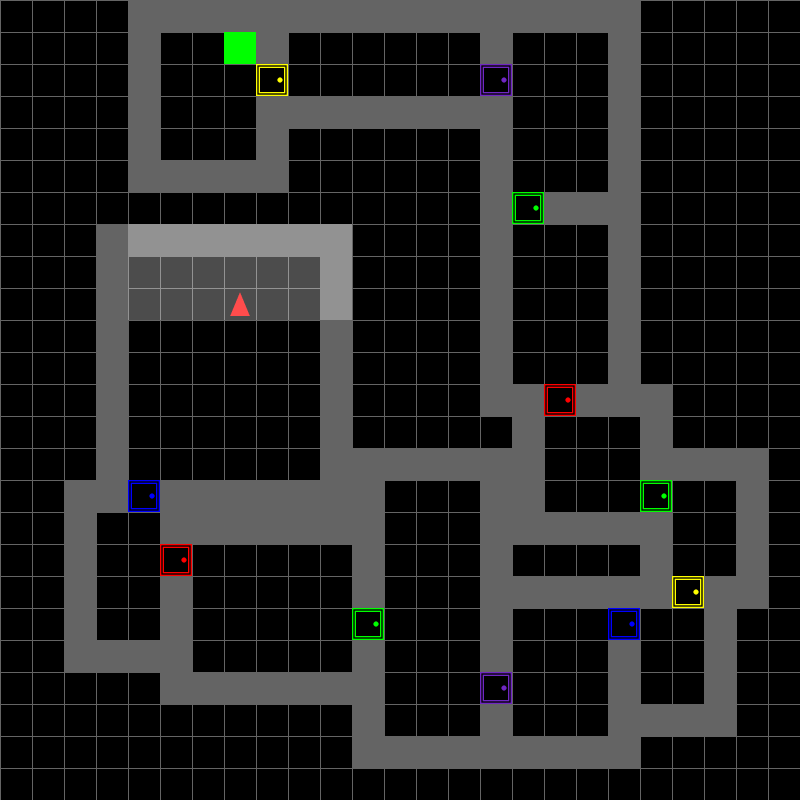}
    \caption{MultiRoom-N12-S10}
    \label{fig:rendering-1}
  \end{subfigure}%
  \begin{subfigure}[b]{0.23\textwidth}
    \centering
    \includegraphics[width=0.85\textwidth]{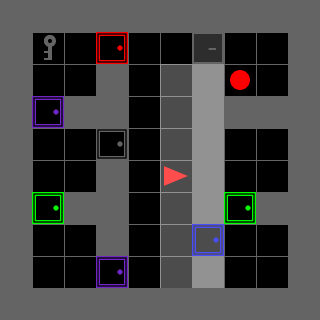}
    \caption{KeyCorridor-S4-R3}
  \end{subfigure}%
  \begin{subfigure}[b]{0.31\textwidth}
    \centering
    \includegraphics[width=0.85\textwidth]{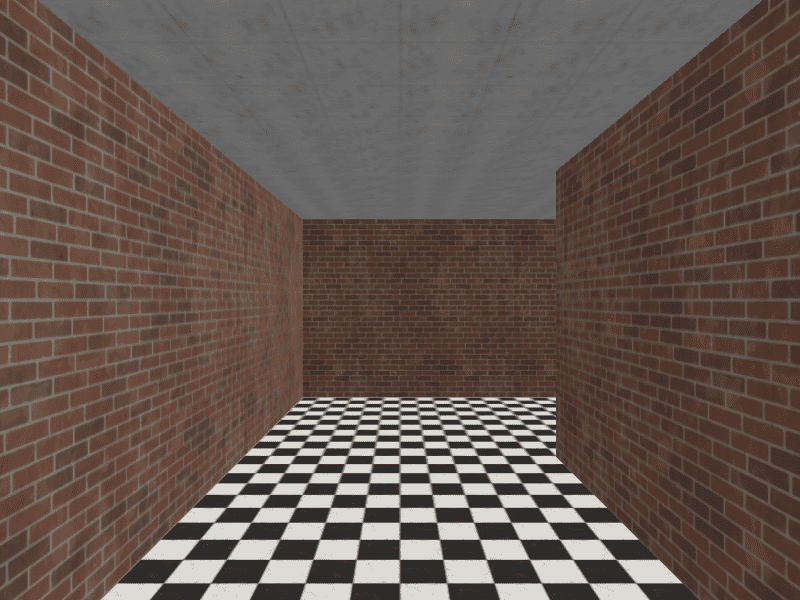}
    \caption{Maze}
  \end{subfigure}%
  \begin{subfigure}[b]{0.23\textwidth}
    \centering
    \includegraphics[width=0.85\textwidth]{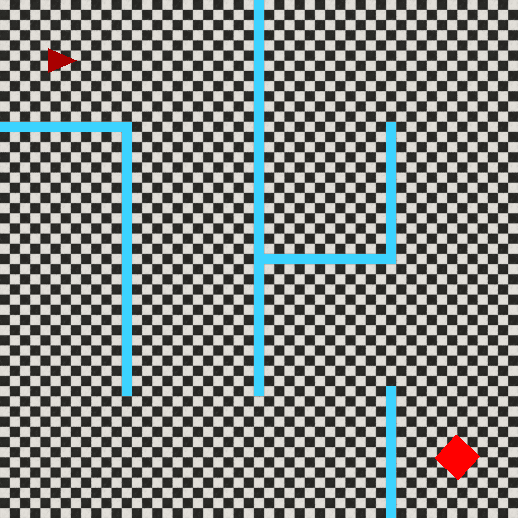}
    \caption{Maze (top view)}
  \end{subfigure}%
  \vspace{-6pt}
  \caption{Rendering of the procedually-generated environments in our experiments.}
  \label{fig:rendering}
  \vspace{-12pt}
\end{figure}

\textbf{Global Score.} The global score is designed to model long-term exploration behavior. While a different environment will be generated in each episode in the procedurally-generated setting, the environments in different episodes usually share some common characteristics. For example, in the MiniGrid MultiRoom environment~(Figure~\ref{fig:rendering-1}), although a different maze will be generated in a new episode, the basic building blocks (e.g., walls, doors, etc.), the number of rooms, and the size of each room will be similar. From a global view, the agent should explore the regions that are not well explored in the past. Motivated by count-based exploration, we define the global score as
\begin{equation}
    S_{\text{global}} = \frac{1}{N_{\text{total}}} \sum_{s}\frac{1}{\sqrt{N(s)}},
    \label{eqn:global}
\end{equation}
where $N(s)$ is the state count of $s$ throughout the training process, that is, the global score is the mean count-based intrinsic reward of all the states in the episode.


\textbf{Episodic Exploration Score.} Suppose $S_{\text{ext}}$ is the total extrinsic reward of an episode received from the environment . The episodic exploration score is obtained by the weighted sum of the extrinsic reward $S_{\text{ext}}$, the local score $S_{\text{local}}$, and the global score $S_{\text{global}}$:


\begin{equation}
\label{eqn:score}
    S = w_0 S_{\text{ext}} + w_1 S_{\text{local}} + w_2 S_{\text{global}},
\end{equation}
where $w_0$, $w_1$ and $w_2$ are hyperparameters. We speculate that the three scores are complementary and the optimal $w_0$, $w_1$ and $w_2$ could vary in different environments. For example, a higher $w_1$ will give a higher weight to the per-episode view of exploration behavior. A higher $w_2$ will focus more on whether the current episode has explored the past unexplored regions. Nevertheless, we find that a single set of $w_0$, $w_1$ and $w_2$ works well across many tasks.

\textbf{Comparison with Prior Work.} Several papers have studied episode-level feedback to improve the sample efficiency of RL. To the best of our knowledge, the existing work mainly focuses on learning a meta-policy based on episodic rewards~\citep{xu2018learning,zha2019experience}, or imitating the episodes with high episodic rewards~\citep{guo2018generative,srivastava2019training}. However, in very sparse environments, it is not likely to receive a non-zero reward without exploration. These methods are not suitable for very sparse environments since episodic rewards could be always zero. In this work, we propose local and global episodic scores to quantify the exploration behaviors and encourage the agent to visit distinct states from per-episode and long-term views.

\begin{algorithm}[t]
\caption{Exploration via Ranking the Episodes (RAPID)}
\label{alg:1}
\setlength{\intextsep}{0pt} 
\begin{algorithmic}[1]
\STATE \textbf{Input:} Training steps $S$, buffer size $D$, RL rollout steps $T$
\STATE Initialize the policy $\pi_{\theta}$, replay buffer $\mathcal{D}$
\FOR{iteration = $1$, $2$, ... until convergence}
    \STATE Execute $\pi_\theta$ for $T$ timesteps
    \STATE Update $\pi_{\theta}$ with RL objective (also update value functions if any)
    \FOR {each generated episode $\tau$}
        \STATE Compute episodic exploration score $S_{\tau}$ based on Eq.~(\ref{eqn:score})
        \STATE Give score $S_{\tau}$ to all the state-action pairs in $\tau$ and store them to the buffer
        \STATE Rank the state-action pairs in $\mathcal{D}$ based on their exploration scores
        \IF {$\mathcal{D}.length > D$}
            \STATE Discard the state-action pairs with low scores so that $\mathcal{D}.length = D$
        \ENDIF
        \FOR{step = $1$, $2$, .., $S$}
        \STATE Sample a batch from $\mathcal{D}$ and train $\pi_\theta$ using the data in $\mathcal{D}$ with behavior cloning
        \ENDFOR
    \ENDFOR
\ENDFOR
\end{algorithmic}
\end{algorithm}

\subsection{Reproducing Past Good Exploration Behaviors with Imitation Learning}
A straightforward idea is to treat the episodic exploration score as a reward signal and assign it to the last timestamp of the episode. In this way, the agent can be updated with any reinforcement learning objective. However, this strategy is not very effective with three potential limitations. First, the reward remains sparse so that the reinforcement learning agents may not be able to effectively exploit the reward signals. Second, the intrinsic exploration bonus may introduce bias in the objective. While the local score and global score may encourage exploration at the beginning stage, they are not true objectives and may degrade the final performance. Third, it only uses a good episode once. In fact, many past good episodes can be exploited many times to improve sample efficiency.

To overcome the above limitations, we propose to use a small ranking buffer to store past good state-action pairs and then use imitation objective to enhance exploration. Specifically, we assign the same episodic exploration score to all the state-action pairs in an episode and rank all the state-action pairs based on scores. In addition to the original RL objective, we employ behavior cloning, the simplest form of imitation learning, to encourage the agent to reproduce the state-action pairs in the buffer. This simple ranking buffer can effectively exploit the past good experiences since a good state-action pair could be sampled more than once. The procedure is summarized in Algorithm~\ref{alg:1}. A possible variant is to keep the entire episodes in the buffer. We have tried this idea and do not observe a clear difference between this variant and Algorithm~\ref{alg:1} (see Appendix~\ref{sec:H} for more discussions).

\section{Experiments}
The experiments are designed to answer the following research questions: \textbf{RQ1:} how is RAPID compared with the state-of-the-art intrinsic reward methods on benchmark procedurally-generated environments~(Section~\ref{sec:minigrid})? \textbf{RQ2:} how will RAPID perform if removing one of the scores (i.e., the local score, the global score, and the extrinsic reward) or the buffer~(Section~\ref{sec:analysis})? \textbf{RQ3:} how will the hyperparameters impact the performance of RAPID~(Section~\ref{sec:analysis})? \textbf{RQ4:} can RAPID explore the environment without extrinsic reward~(Section~\ref{sec:analysis})? \textbf{RQ5:} how will RAPID perform with larger and more challenging grid-world~(Section~\ref{sec:analysis})? \textbf{RQ6:} can RAPID generalize to 3D navigation task and continuous state/action space, such as MuJoCo~(Section~\ref{sec:miniworld-mujoco})?

\subsection{Environments and Baselines}
Figure~\ref{fig:rendering} shows the rendering of procedurally-generated environments used in this work. Following the previous studies of exploration in procedurally-generated environments~\citep{Raileanu2020RIDE,campero2020learning}, we mainly focus on MiniGrid environments~\citep{chevalier2018minimalistic}, which provide procedurally-generated grid-worlds. We consider two types of hard exploration tasks: \emph{MultiRoom-NX-SY}, where X and Y indicate the number of rooms and the room size, respectively, and \emph{KeyCorridor-SX-RY}, where X and Y indicate the room size and the number of rows, respectively. A large X or Y suggests a larger environment, which will be more difficult to explore. The agent has a partial view of the environment and needs to navigate the green block or the red ball under sparse rewards. These tasks are challenging because the agent needs to explore many rooms in order to receive a reward signal so that standard RL algorithms will typically fail. To test whether RAPID can generalize to continuous state space, we consider MiniWorld Maze~\citep{gym_miniworld}, whose observation space is $60 \times 80$ images. Similarly, the agent in MiniWorld Maze has a 3D partial view of the environment and needs to navigate the red box. The maze is procedurally-generated, i.e., a different maze will be generated in each episode. We further study RAPID in continuous state and action spaces on sparse MuJoCo tasks~\citep{todorov2012mujoco,brockman2016openai}. More details of the above environments are described in Appendx~\ref{sec:B}.

While exploration has been extensively studied in the literature, very few methods are designed for procedurally-generated environments. To better understand the performance of RAPID, we consider the baselines in the following categories. \textbf{Traditional Methods:} we consider several popular methods designed for singleton environments, including count-based exploration~(COUNT)~\citep{bellemare2016unifying}, Random Network Distillation Exploration~(RANDOM)~\citep{burda2018exploration}, and curiosity-driven exploration~(CURIOSITY)~\citep{pathak2017curiosity}. \textbf{Methods for Procedurally-Generated Environments:} we consider two state-of-the-art exploration methods, including Impact-Driven Exploration~(RIDE)~\citep{Raileanu2020RIDE}, and exploration with Adversarially Motivated Intrinsic Goals~(AMIGO)~\citep{campero2020learning}. \textbf{Self-Imitation Methods}: self-imitation learning~(SIL)~\citep{oh2018self} also exploits past good experiences to encourage exploration. We further include PPO~\citep{schulman2017proximal} as a reinforcement learning baseline. All the algorithms are run the same number timesteps for a fair comparison. More details are provided in Appendix~\ref{sec:A}.

\begin{figure}[t]
  \centering
  \begin{subfigure}[b]{0.85\textwidth}
    \includegraphics[width=1.0\textwidth]{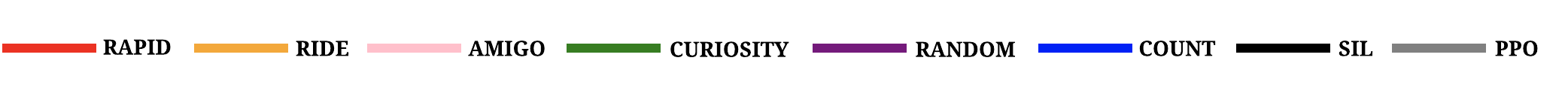}
  \end{subfigure}

  \begin{subfigure}[b]{0.25\textwidth}
    \centering
    \includegraphics[width=0.99\textwidth]{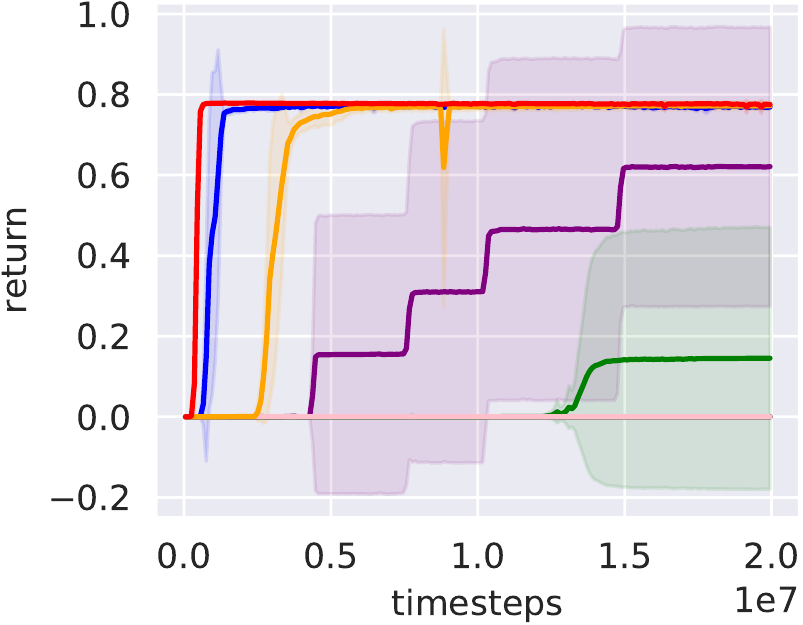}
    \caption{MultiRoom-N7-S4}
  \end{subfigure}%
  \begin{subfigure}[b]{0.25\textwidth}
    \centering
    \includegraphics[width=0.99\textwidth]{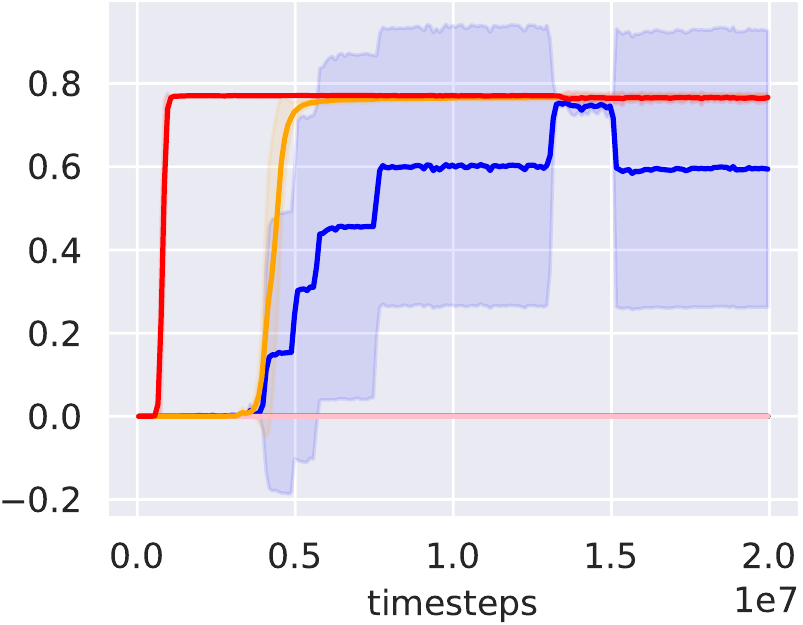}
    \caption{MultiRoom-N10-S4}
  \end{subfigure}%
  \begin{subfigure}[b]{0.25\textwidth}
    \centering
    \includegraphics[width=0.99\textwidth]{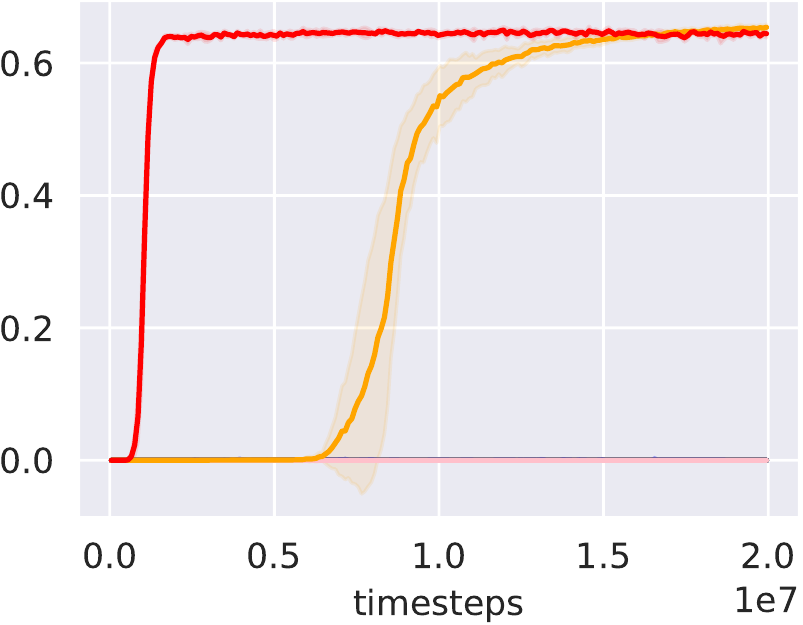}
    \caption{MultiRoom-N7-S8}
  \end{subfigure}%
  \begin{subfigure}[b]{0.25\textwidth}
    \centering
    \includegraphics[width=0.99\textwidth]{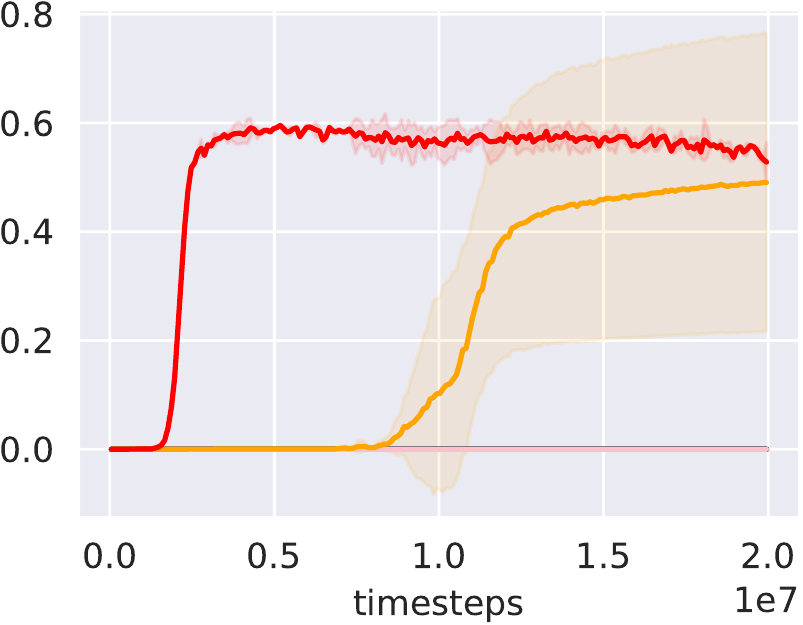}
    \caption{MultiRoom-N10-S10}
  \end{subfigure}%
  
  \begin{subfigure}[b]{0.25\textwidth}
    \centering
    \includegraphics[width=0.99\textwidth]{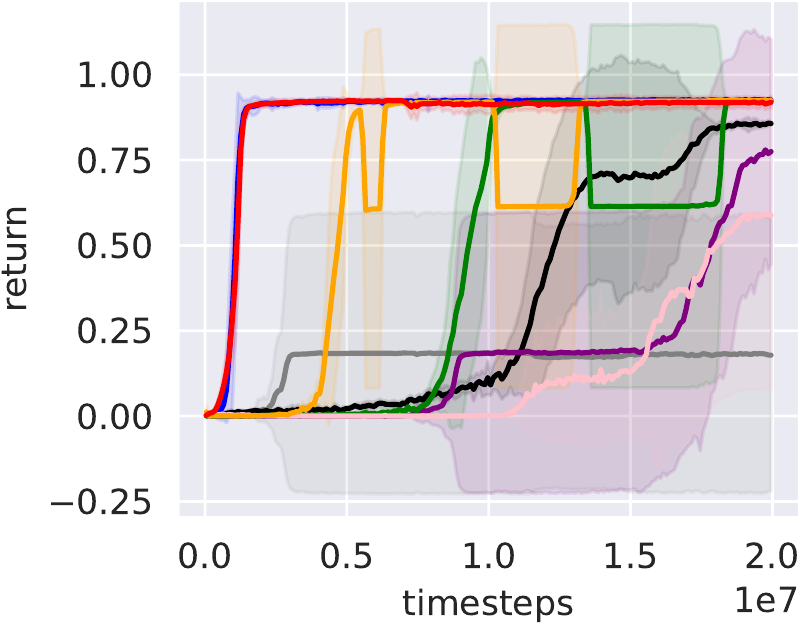}
    \caption{KeyCorridor-S3-R2}
  \end{subfigure}%
  \begin{subfigure}[b]{0.25\textwidth}
    \centering
    \includegraphics[width=0.99\textwidth]{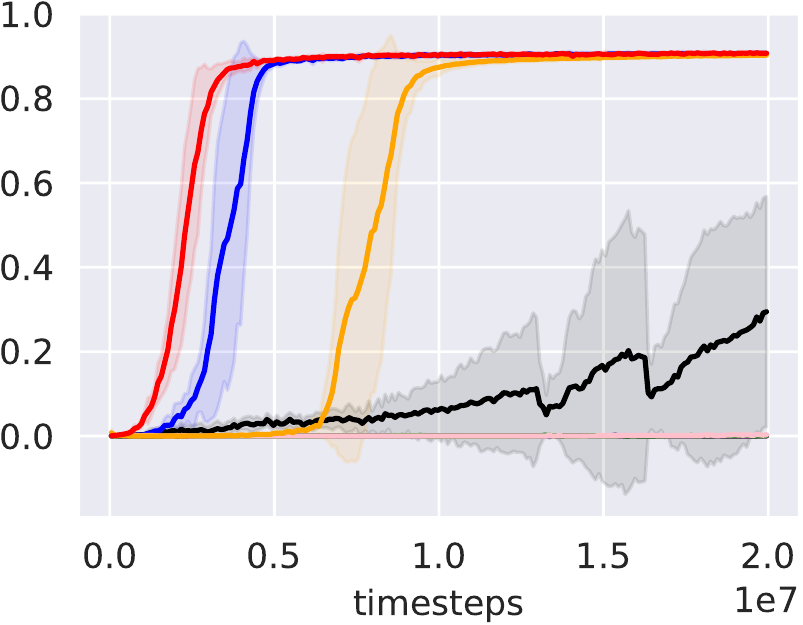}
    \caption{KeyCorridor-S3-R3}
  \end{subfigure}%
  \begin{subfigure}[b]{0.25\textwidth}
    \centering
    \includegraphics[width=0.99\textwidth]{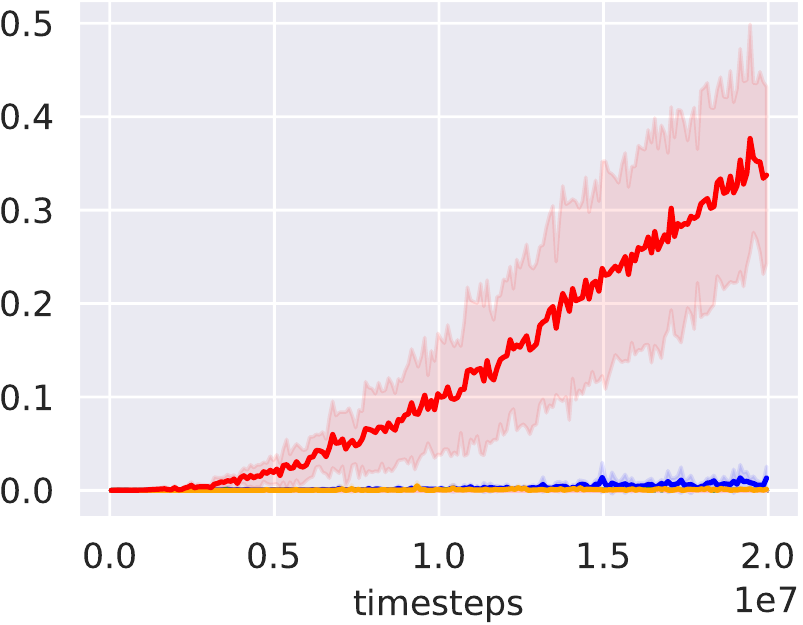}
    \caption{KeyCorridor-S4-R3}
  \end{subfigure}%
  \begin{subfigure}[b]{0.25\textwidth}
    \centering
    \includegraphics[width=0.99\textwidth]{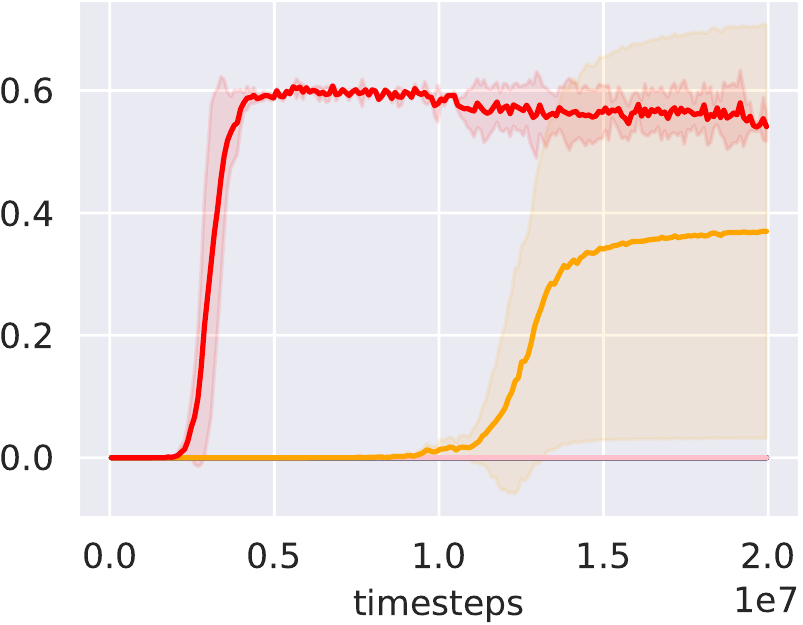}
    \caption{MultiRoom-N12-S10}
  \end{subfigure}%
  \vspace{-7pt}
  \caption{Performance of RAPID against baselines on hard-exploration environments in MiniGrid. All the experiments are run $5$ times. The shaded area represents mean $\pm$ standard deviation.}
  \label{fig:minigrid}
  \vspace{-8pt}
\end{figure}

\begin{table}[t]
    \centering
    \scriptsize
    \setlength{\tabcolsep}{2.9pt}
    \begin{tabular}{l|c|c|c|c|c|c}
    \toprule
     & \textbf{RAPID} & \textbf{w/o local} & \textbf{w/o global} & \textbf{w/o reward} & \textbf{w/o buffer} & w/o ranking\\
    \midrule
     MR-N7S4    &  \bm{$0.787 \pm 0.001$} &\bm{$0.787 \pm 0.000$} & \bm{$0.787 \pm 0.001$} &  $0.781 \pm 0.002$ & $0.000 \pm 0.000$ & $0.002 \pm 0.002$\\
     MR-N10S4    &  \bm{$0.778 \pm 0.000$} & \bm{$0.778 \pm 0.000$} & \bm{$0.778 \pm 0.001$} & $0.775 \pm 0.002$ & $0.000 \pm 0.000$ & $0.000 \pm 0.000$\\
     MR-N7S8    &  \bm{$0.678 \pm 0.001$} & $0.677 \pm 0.002$ & $0.677 \pm 0.002$ & $0.652 \pm 0.004$ & $0.000 \pm 0.000$ & $0.000 \pm 0.000$\\
     MR-N10S10    &  \bm{$0.632 \pm 0.001$} & $0.238 \pm 0.288$ & $0.630 \pm 0.002$ & $0.604 \pm 0.010$ & $0.000 \pm 0.000$ & $0.000 \pm 0.000$\\
     MR-N12S10    &  \bm{$0.644 \pm 0.001$} & $0.001 \pm 0.001$ & $0.633 \pm 0.005$ & $0.613 \pm 0.007$ & $0.000 \pm 0.000$ & $0.000 \pm 0.000$\\
     
     KC-S3R2    & \bm{$0.934 \pm 0.004$} & $0.933 \pm 0.002$ & \bm{$0.934 \pm 0.000$} & $0.929 \pm 0.003$& $0.018 \pm 0.008$ & $0.527 \pm 0.380$\\
     KC-S3R3    & \bm{$0.922 \pm 0.001$} & $0.885 \pm 0.022$ & $0.912 \pm 0.003$ & $0.903 \pm 0.002$ & $0.012 \pm 0.007$ & $0.013 \pm 0.006$\\
     KC-S4R3    & \bm{$0.473 \pm 0.087$} & $0.150 \pm 0.095$ & $0.244 \pm 0.144$ & $0.035 \pm 0.035$ & $0.000 \pm 0.000$ & $0.001 \pm 0.001$\\
   
     \bottomrule
    \end{tabular}
    \vspace{-6pt}
    \caption{Performance of RAPID and the ablations on MiniGrid environments. The mean maximum returns and standard deviations are reported. MR stands for MultiRoom; KC stands for KeyCorridor. Learning curves are in Appendix~\ref{sec:C}.}.
    \label{tab:minigrid-hyper}
    \vspace{-15pt}
\end{table}

\subsection{MiniGrid Results}
\label{sec:minigrid}
To study \textbf{RQ1}, we plot the learning curves of RAPID and the baselines on MiniGrid benchmarks in Figure~\ref{fig:minigrid}. We can make three observations. First, RAPID performs significantly better than the state-of-the-art methods designed for singleton or procedurally-generated environments in terms of sample efficiency and final performance. For example, on MultiRoom-N7-S8, RAPID is at least 10 times more sample efficient than RIDE. On KeyCorridor-S4-R3, RAPID is the only algorithm that successfully navigates the goal within 20 million timesteps. Second, the traditional intrinsic reward methods perform well on easy procedurally-generated environments but perform poorly on hard-exploration tasks. For example, COUNT performs as well as RAPID on KeyCorridor-S3-R2 but fails on more challenging tasks, such as MultiRoom-N7-S8. This suggests that traditional methods may fall short in challenging procedurally-generated environments, which is consistent with the results in previous work~\citep{Raileanu2020RIDE}. Third, RAPID outperforms SIL across all the tasks. Recall that both RAPID and SIL learn to reproduce past good experiences. While SIL reproduces experiences with high rewards, RAPID additionally considers local and global exploration scores. This verifies the effectiveness of the episodic exploration scores.

\subsection{Ablations and Analysis}
\label{sec:analysis}
To study \textbf{RQ2}, we perform ablation studies on RAPID. Specifically, we remove one of the local score, the global score, and the extrinsic reward, and compare each ablation with RAPID. Besides, we consider a variant that directly uses the episodic exploration score as the reward signal without buffer. To demonstrate the effectiveness of the ranking, we further include a baseline that uses the replay buffer without ranking. Table~\ref{tab:minigrid-hyper} shows the results. We observe that the local score contributes the most to the performance. While most existing exploration methods reward exploration globally, the global score could be not aware of generalization, which may explain the unsatisfactory performance of COUNT and CURIOSITY. While the global score seems to play a relatively small role as compared to local score in most MiniGrid environments, it plays an important role in KeyCorridor-S4-R3. A possible reason is that the local score is a too weak signal in KeyCorridor-S4-R3, and the global score may help alleviate this issue since it provides some historical information. We also observe that the agent fails in almost all the tasks without buffer or ranking. This suggests that the ranking buffer is an effective way to exploit past good exploration behaviors.

\begin{figure}[t]
  \centering

  \begin{subfigure}[b]{0.25\textwidth}
    \centering
    \includegraphics[width=0.99\textwidth]{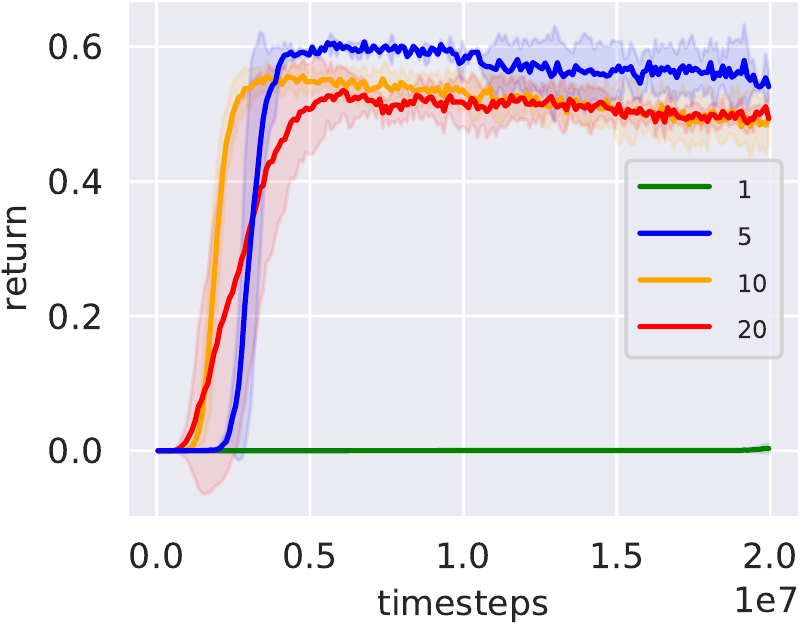}
    \caption{Impact of training steps}
    \label{fig:analysis-1}
  \end{subfigure}%
  \begin{subfigure}[b]{0.25\textwidth}
    \centering
    \includegraphics[width=0.99\textwidth]{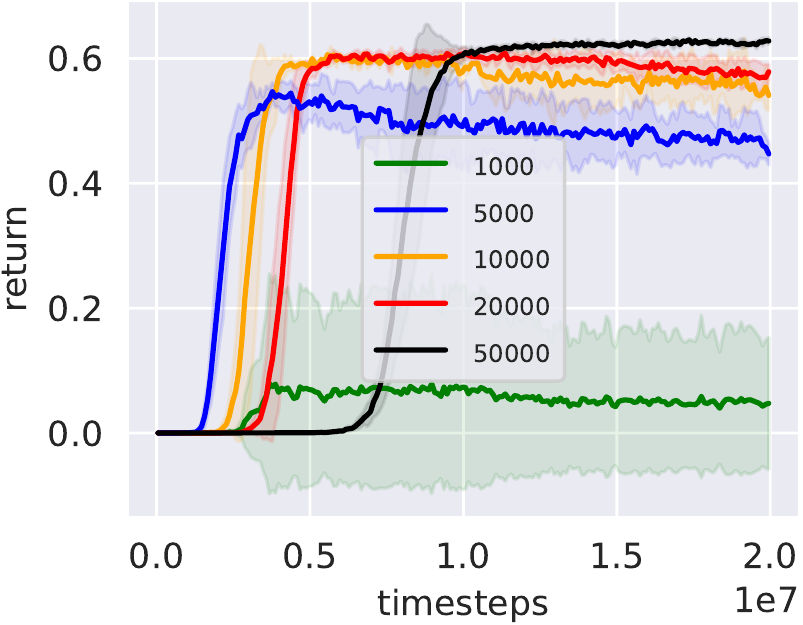}
    \caption{Imapct of buffer size}
    \label{fig:analysis-2}
  \end{subfigure}%
  \begin{subfigure}[b]{0.25\textwidth}
    \centering
    \includegraphics[width=0.99\textwidth]{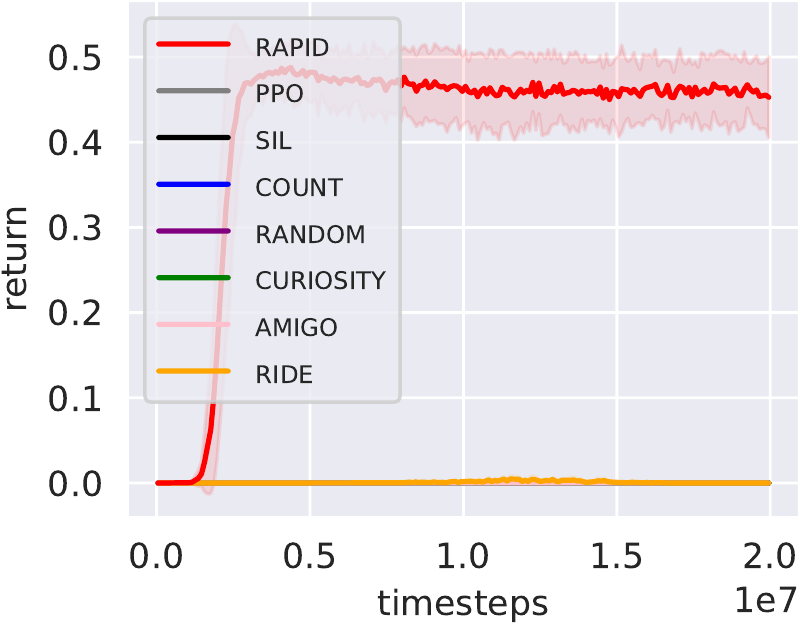}
    \caption{Pure exploration}
    \label{fig:analysis-3}
  \end{subfigure}%
  \begin{subfigure}[b]{0.25\textwidth}
    \centering
    \includegraphics[width=0.99\textwidth]{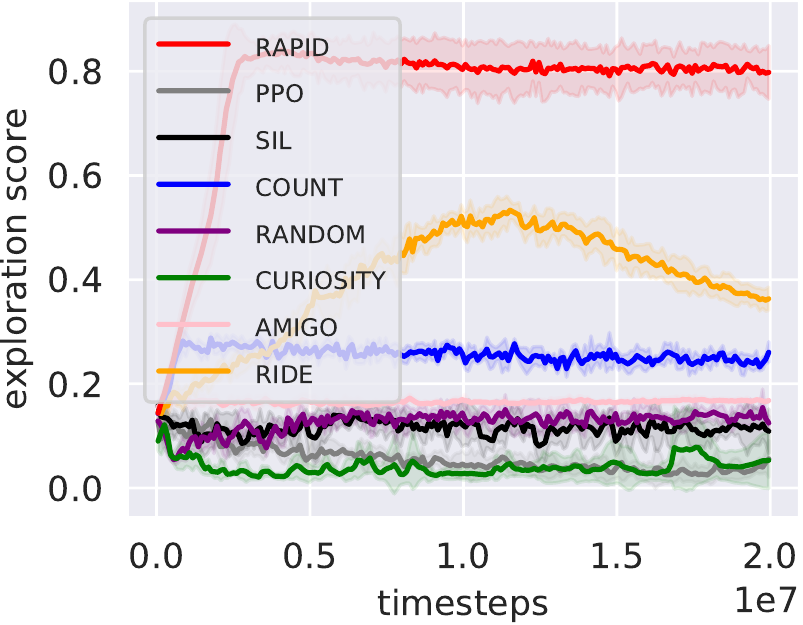}
    \caption{Local exploration score}
    \label{fig:analysis-4}
  \end{subfigure}%
  \vspace{-5pt}
  \caption{Analysis of RAPID on MultiRoom-N12-S10. Results for other environments are in Appendix~\ref{sec:D}. \textbf{(a)(b):} the impact of the hyperparameters. \textbf{(c)(d):} the returns without extrinsic rewards (pure exploration) and the corresponding local exploration scores.}
  \label{fig:analysis}
  \vspace{-10pt}
\end{figure}

\begin{figure}[t]
  \centering
  \begin{subfigure}[b]{0.28\textwidth}
    \centering
    \includegraphics[width=0.80\textwidth]{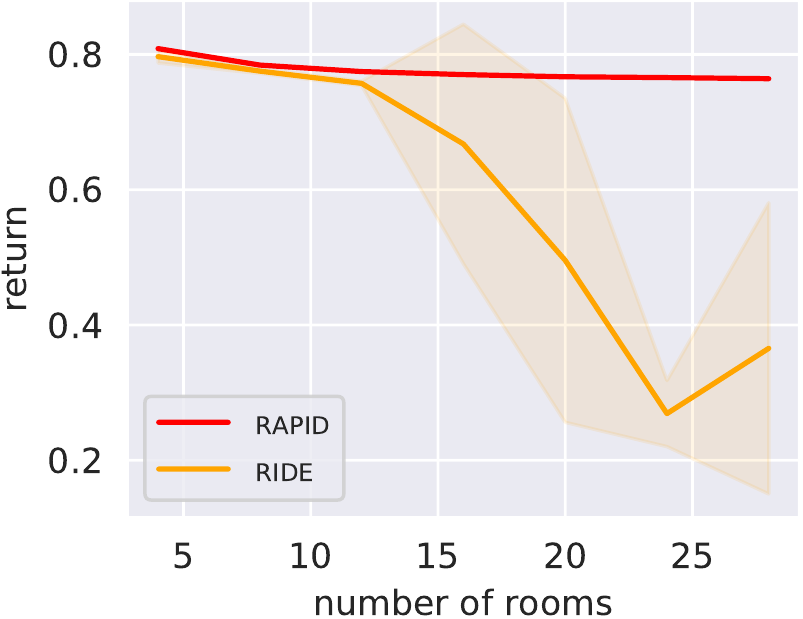}
  \end{subfigure}%
  \begin{subfigure}[b]{0.28\textwidth}
    \centering
    \includegraphics[width=0.80\textwidth]{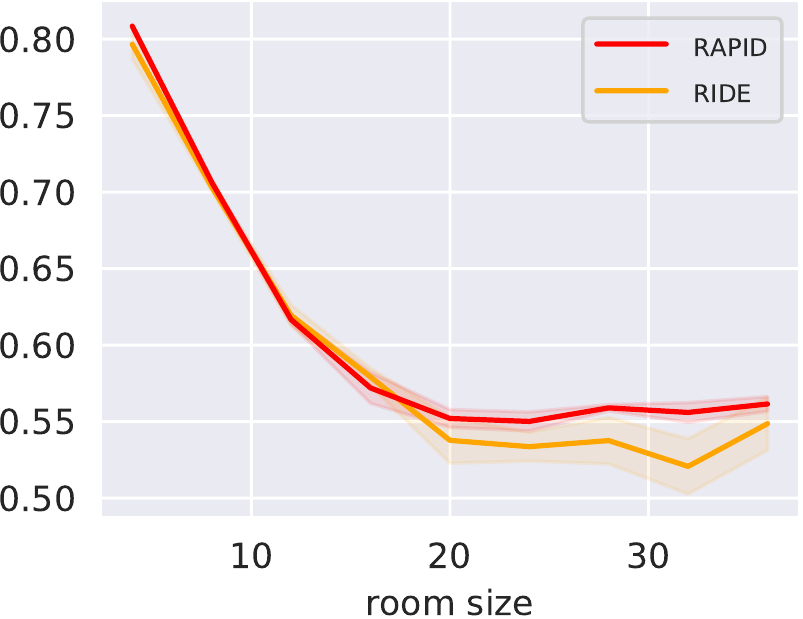}
  \end{subfigure}%
  \begin{subfigure}[b]{0.28\textwidth}
    \centering
    \includegraphics[width=0.80\textwidth]{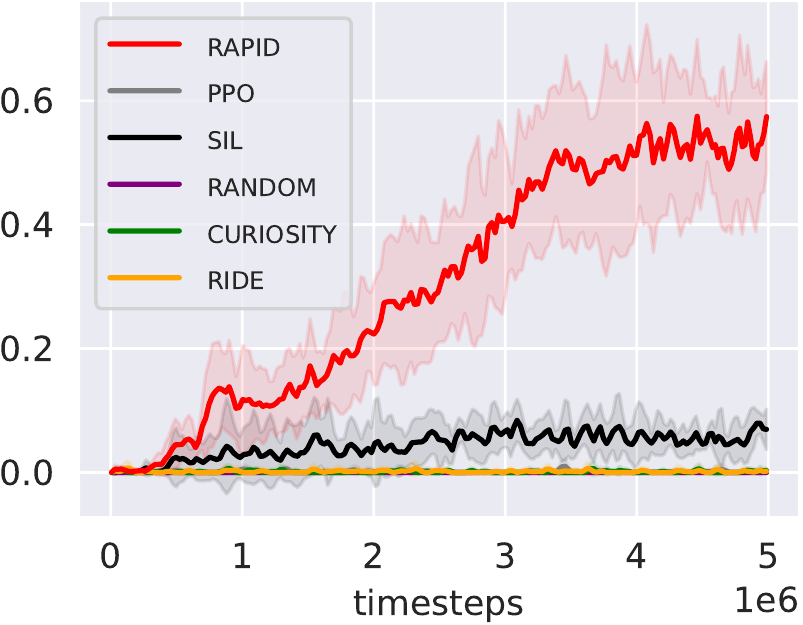}
  \end{subfigure}%
  \vspace{-10pt}
  
  \caption{\textbf{Left:} the maximum returns w.r.t. the number of rooms under room size $4$ (full curves in Appendix~\ref{sec:E}). \textbf{Middle:} the maximum returns w.r.t. the room sizes with $4$ rooms (full curves in Appendix~\ref{sec:F}). \textbf{Right:} learning curves on MiniWolrd Maze (more results in Appendix~\ref{sec:G}).}
  \label{fig:others}
  \vspace{-15pt}
\end{figure}

To study \textbf{RQ3}, we plot the learning curves with different training steps $S$ and buffer size $D$ in Figure~\ref{fig:analysis-1} and \ref{fig:analysis-2}. We observe that larger number of training steps usually leads to faster learning speed at the beginning stage but may result in sub-optimal performance. For the buffer size, we observe that a large buffer will lead to slow learning speed with more stable performance. We speculate that there is a trade-off between learning speed and the final performance. In our experiments, we set training steps $S$ to be $5$ and buffer size $D$ to be $10000$, respectively, across all the tasks.

To investigate \textbf{RQ4}, we remove extrinsic rewards and show the learning curves of RAPID and the baselines in Figure~\ref{fig:analysis-3} with the corresponding local exploration scores in Figure~\ref{fig:analysis-4}. We make two interesting observations. First, without extrinsic rewards, RAPID can deliver reasonable performance. For example, on MultiRoom-N12-S10, RAPID achieves nearly $0.5$ average return with pure exploration, compared with around $0.6$ if using extrinsic rewards. We look into the episodes generated by RAPID with pure exploration and observe that, in most cases, the agent is able to reach the goal but struggles to find the optimal path to the goal. This is expected because the extrinsic reward is obtained by whether the agent can reach the goal subtracting some penalties of the timesteps spent. The extrinsic reward can encourage the agent to choose the shortest path to reach the goal. For other baselines, we observe that only RIDE and AMIGO can occasionally reach the goal. Second, we observe that some baselines, such as RIDE, are indirectly maximizing the local exploration scores. For example, we observe that the local exploration score increases for RIDE on MultiRoom-N12-S10. However, the local score of RAPID drops after about $10$ million timesteps. This suggests the episode-level score may better align with good exploration behaviors. A possible explanation for the superiority of RAPID is that it is designed to directly optimize episodic exploration score. 

To answer \textbf{RQ5}, we focus on \emph{MultiRoom-NX-SY} and ambitiously vary $X$ and $Y$ to make the environments more challenging. Specifically, we fix $Y$ to be $4$ and vary $X$ in the left-hand-side of Figure~\ref{fig:others}, and fix $X$ to be $4$ and vary $Y$ in the middle of Figure~\ref{fig:others}. Compared with RIDE, our RAPID delivers stronger performance on more challenging mini-worlds.

\vspace{-5pt}
\subsection{MiniWorld Maze and Sparse MuJoCo Results}
\label{sec:miniworld-mujoco}
\begin{figure}[t]
  \centering
  \begin{subfigure}[b]{0.3\textwidth}
    \includegraphics[width=\textwidth]{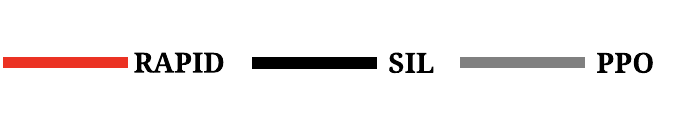}
  \end{subfigure}
  
  \begin{subfigure}[b]{0.25\textwidth}
    \centering
    \includegraphics[width=0.99\textwidth]{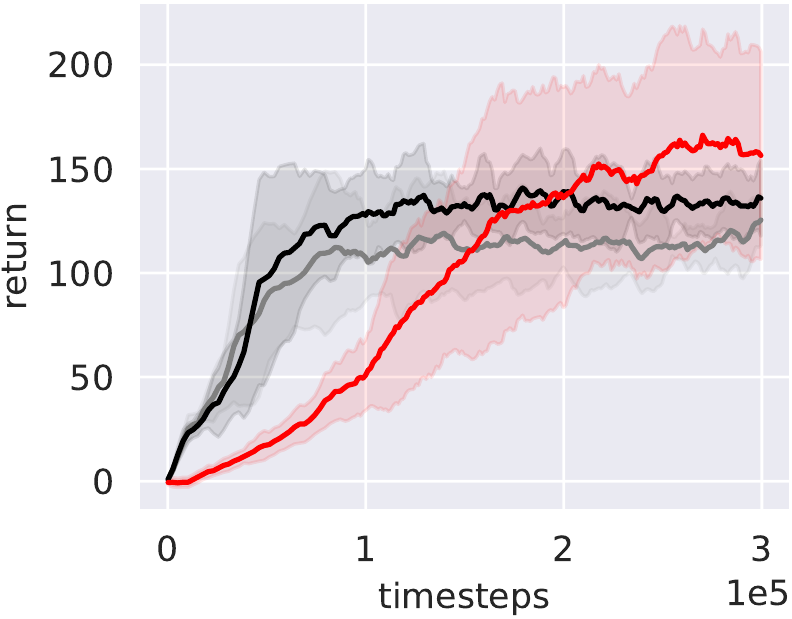}
    \caption{Walker2d}
  \end{subfigure}%
  \begin{subfigure}[b]{0.25\textwidth}
    \centering
    \includegraphics[width=0.99\textwidth]{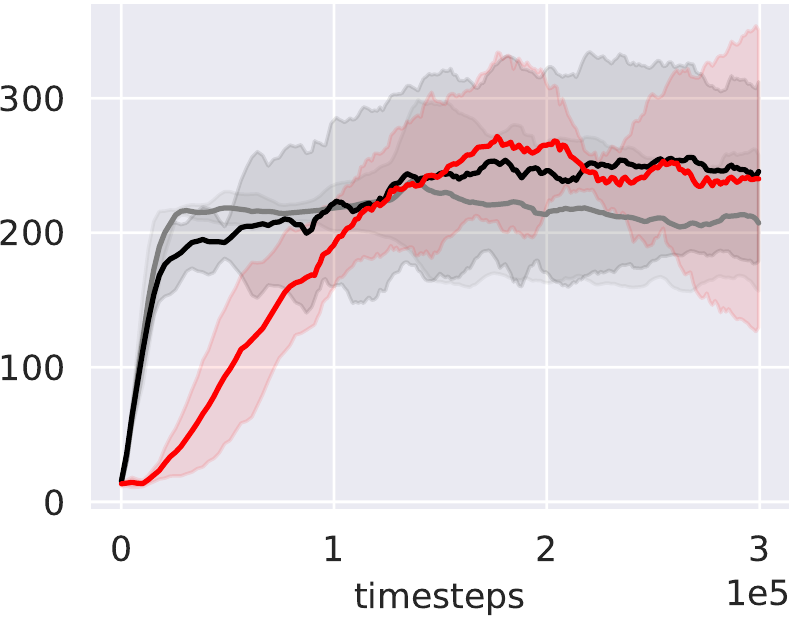}
    \caption{Hopper}
  \end{subfigure}%
  \begin{subfigure}[b]{0.25\textwidth}
    \centering
    \includegraphics[width=0.99\textwidth]{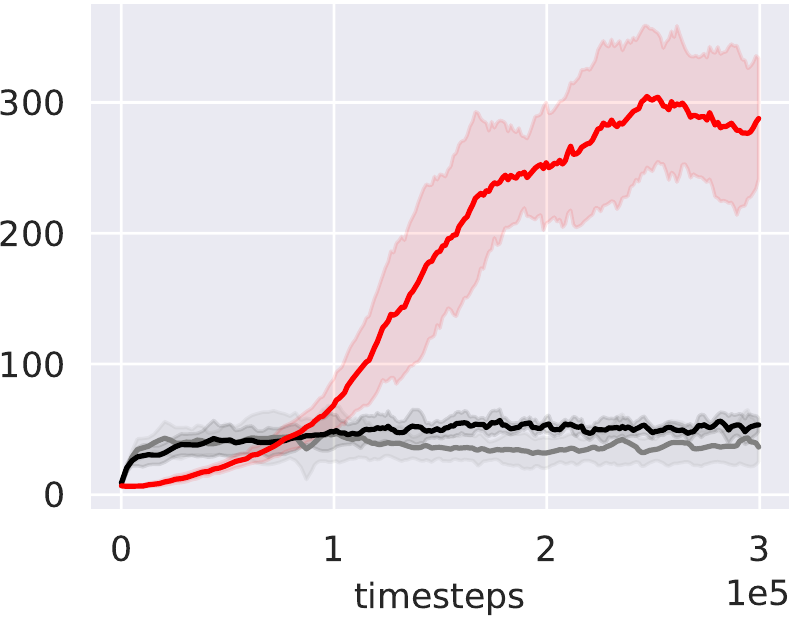}
    \caption{InvertedPendulum}
  \end{subfigure}%
  \begin{subfigure}[b]{0.25\textwidth}
    \centering
    \includegraphics[width=0.99\textwidth]{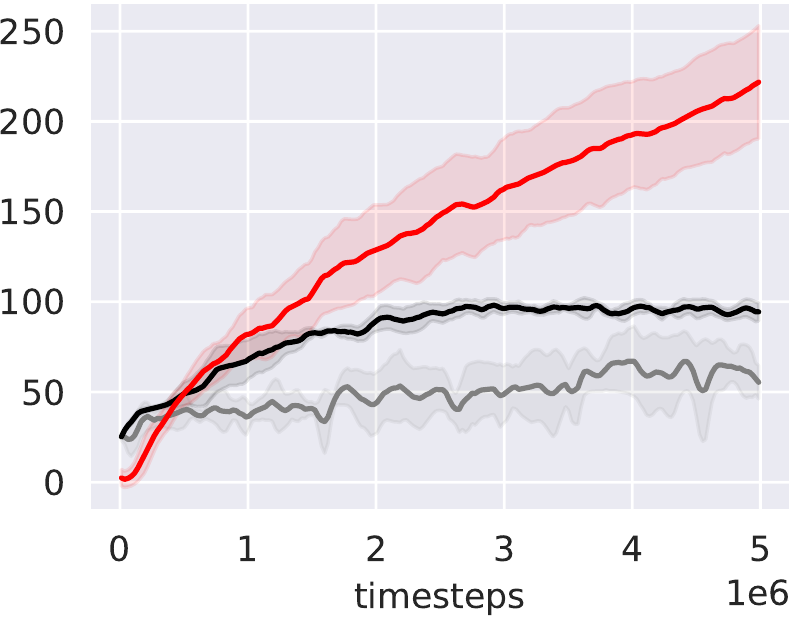}
    \caption{Swimmer}
  \end{subfigure}%
  \vspace{-10pt}
  \caption{Performance on MuJoCo tasks with episodic reward. i.e., a non-zero reward is only provided at the end of an episode. All the experiments are run $5$ times with different random seeds.}
  \label{fig:7}
  \vspace{-10pt}
\end{figure}

For \textbf{RQ6}, the learning curves on 3D Maze are shown in the right-hand-side of Figure~\ref{fig:others}. RAPID performs significantly better than the baselines. We also conduct experiments on a subset of sparse MuJoCo tasks~(Figure~\ref{fig:7}). The results show that RAPID improves over PPO and SIL. Interestingly, RAPID achieves more than $200$ average return on Swimmer, which surpasses some previous results reported on the dense reward counterpart~\citep{duan2016benchmarking,henderson2018deep,lai2020dual}. Note that the MuJoCo environments are singleton. We use these environments solely to test whether RAPID can deal with continuous state/action spaces. We have also tried adapting Swimmer to be procedurally-generated and observe similar results (Appendix~\ref{sec:I})

\vspace{-7pt}
\section{Related Work}
\vspace{-5pt}

\textbf{Count-Based and Curiosity-Driven Exploration.} These two classes of intrinsic motivations are the most popular and proven to be effective in the literature. Count-based exploration encourages the agent to visit the novel states, which is first studied in tabular setting~\citep{strehl2008analysis} and has been recently extended to more complex state space~\citep{bellemare2016unifying,ostrovski2017count,tang2017exploration,martin2017count,machado2018count,osband2019deep}. Curiosity-driven exploration learns the environment dynamics to encourage exploration~\citep{stadie2015incentivizing,pathak2017curiosity,sukhbaatar2018intrinsic,burda2018large}. However, these methods are designed for singleton environments and often struggle to generalize to procedurally-generated environments~\citep{Raileanu2020RIDE}.

\textbf{Exploration for Procedurally-Generated Environments.} Several recent studies have discussed the generalization of reinforcement learning~\citep{rajeswaran2017towards,zhang2018dissection,zhang2018study,choi2018contingency} and designed procedurally-generated environments to test the generalization of reinforcement learning~\citep{beattie2016deepmind,nichol2018gotta,kuttler2020nethack}. More recent papers show that traditional exploration methods fall short in procedurally-generated environments and address this issue with new exploration methods~\citep{Raileanu2020RIDE,campero2020learning}. This work studies a new perspective of exploration bonus in episode-level and achieves significantly better performance than the previous methods on procedurally-generated benchmarks.

\textbf{Experience Replay Buffer and Self-Imitation Learning.} Experience replay mechanism is widely used and is initially designed to stabilize deep reinforcement learning~\citep{lin1992self,lin1993reinforcement,mnih2015human,zhang2017deeper,zha2019experience,fedus2020revisiting}. Self-Imitation Learning extends the experience replay buffer and uses it to store past good experiences~\citep{oh2018self,guo2018generative,guo2019efficient,gangwani2018learning}. However, Self-Imitation only exploits highly rewarded experiences and thus is not suitable for very sparse environments. Our work incorporates local and global scores to encourage past good exploration behaviors in sparse procedurally-generated environments.

\textbf{Episodic Memory.} Inspired by the episodic memory of the animals, some studies propose to use episodic memory to enhance sample efficiency~\citep{blundell2016model,pritzel2017neural}. The key idea is to memorize and replay the good episodic experiences. More recently, episodic memory is used to generate intrinsic rewards to guide exploration~\citep{savinov2018episodic}. Never Give Up~(NGU)~\citep{badia2019never} further considers local and global novelties in the intrinsic rewards. Our work differs from NGU in that we approach the sample efficiency from a different angle by treating each episode as a whole and using a ranking mechanism to select good experiences. Thus, good experiences can be exploited multiple times to improve sample efficiency.

\section{Limitations and Discussions}
RAPID is designed to study our hypothesis that imitating episode-level good exploration behaviors can encourage exploration in procedurally-generated environments. In the presented algorithm, we have made some simple choices. In order to make RAPID more competitive to the state-of-the-art methods in more complicated environments, in this section, we highlight some limitations of RAPID and the potential research directions.

First, in our work, the local and global scores are mainly calculated on low-dimensional state space. We have not yet tested the scores on more complex procedurally-generated environments, such as ProcGen benchmark~\citep{cobbe2019leveraging}, TextWorld~\citep{cote2018textworld}, and NetHack Learning Environment~\citep{kuttler2020nethack}, some of which require high-dimensional image inputs. In this context, count-based scores may not be well-suited. Learning the state embedding~\citep{Raileanu2020RIDE} and pseudo-counts~\citep{bellemare2016unifying,ostrovski2017count} could be the potential ways to address this issue. Moreover, high extrinsic rewards may not necessarily mean good exploration behaviors in these environments. The applicability of RAPID in these environments needs future research efforts.

Second, the ranking buffer greedily selects the highly-reward state-action pairs, which is the key to boosting the performance. However, with this greedy strategy, some highly-ranked state-action pairs may never be forgotten even if they are out-of-date, which may introduce bias. As shown in Figure~\ref{fig:analysis-2}, a larger buffer size will lead to more stable performance but will sacrifice the sample efficiency. In more complicated environments, selecting the proper buffer size could be non-trivial. Some forgetting mechanism~\citep{novati2019remember} and meta-learning strategies~\citep{zha2019experience} could be the potential directions to better manage the buffer.

Third, in our implementation, we employ behavior cloning, the simplest form of imitation learning, to train the policy. In recent years, many more advanced imitation learning methods have been proposed~\citep{ho2016generative,hester2017deep}. In particular, the idea of inverse reinforcement learning~\citep{abbeel2004apprenticeship} is to learn reward functions based on the demonstrations. An interesting future direction is to study whether we can derive intrinsic rewards based on the experiences in the buffer via inverse reinforcement learning. Specifically, is there a connection between RAPID and intrinsic rewards from the view of inverse reinforcement learning? Can RAPID be combined with intrinsic rewards to improve the performance? We hope 
that RAPID can facilitate future research to understand these questions. 

\section{Conclusions and Future Work}

This work studies how we can reward good exploration behaviors in procedurally-generated environments. We hypothesize that episode-level exploration score could be a more general criterion to distinguish good exploration behaviors. To verify our hypothesis, we design RAPID, a practical algorithm that learns to reproduce past good exploration behaviors with a ranking buffer. Experiments on benchmark hard-exploration procedurally-generated environments show that RAPID significantly outperforms state-of-the-art intrinsic reward methods in terms of sample efficiency and final performance. Further analysis shows that RAPID has robust performance across different configurations, even without extrinsic rewards. Moreover, experiments on a 3D navigation task and a subset of sparse MuJoCo tasks show that RAPID could generalize to continuous state/action spaces, showing the promise in rewarding good episode-level exploration behaviors to encourage exploration. In the future, we will extend our idea and test it in more challenging procedurally-generated environments. We will also explore the possibility of combining the episode-level exploration bonus with intrinsic rewards. Last, we will try other imitation learning techniques beyond behavior cloning.

\bibliography{iclr2021_conference}

\begin{thebibliography}{64}
\providecommand{\natexlab}[1]{#1}
\providecommand{\url}[1]{\texttt{#1}}
\expandafter\ifx\csname urlstyle\endcsname\relax
  \providecommand{\doi}[1]{doi: #1}\else
  \providecommand{\doi}{doi: \begingroup \urlstyle{rm}\Url}\fi

\bibitem[Abbeel \& Ng(2004)Abbeel and Ng]{abbeel2004apprenticeship}
Pieter Abbeel and Andrew~Y Ng.
\newblock Apprenticeship learning via inverse reinforcement learning.
\newblock In \emph{Proceedings of the twenty-first international conference on
  Machine learning}, pp.\ ~1, 2004.

\bibitem[Andrychowicz et~al.(2017)Andrychowicz, Wolski, Ray, Schneider, Fong,
  Welinder, McGrew, Tobin, Abbeel, and Zaremba]{andrychowicz2017hindsight}
Marcin Andrychowicz, Filip Wolski, Alex Ray, Jonas Schneider, Rachel Fong,
  Peter Welinder, Bob McGrew, Josh Tobin, OpenAI~Pieter Abbeel, and Wojciech
  Zaremba.
\newblock Hindsight experience replay.
\newblock In \emph{Advances in Neural Information Processing Systems}, 2017.

\bibitem[Badia et~al.(2019)Badia, Sprechmann, Vitvitskyi, Guo, Piot,
  Kapturowski, Tieleman, Arjovsky, Pritzel, Bolt, et~al.]{badia2019never}
Adri{\`a}~Puigdom{\`e}nech Badia, Pablo Sprechmann, Alex Vitvitskyi, Daniel
  Guo, Bilal Piot, Steven Kapturowski, Olivier Tieleman, Martin Arjovsky,
  Alexander Pritzel, Andrew Bolt, et~al.
\newblock Never give up: Learning directed exploration strategies.
\newblock In \emph{International Conference on Learning Representations}, 2019.

\bibitem[Beattie et~al.(2016)Beattie, Leibo, Teplyashin, Ward, Wainwright,
  K{\"u}ttler, Lefrancq, Green, Vald{\'e}s, Sadik, et~al.]{beattie2016deepmind}
Charles Beattie, Joel~Z Leibo, Denis Teplyashin, Tom Ward, Marcus Wainwright,
  Heinrich K{\"u}ttler, Andrew Lefrancq, Simon Green, V{\'\i}ctor Vald{\'e}s,
  Amir Sadik, et~al.
\newblock Deepmind lab.
\newblock \emph{arXiv preprint arXiv:1612.03801}, 2016.

\bibitem[Bellemare et~al.(2016)Bellemare, Srinivasan, Ostrovski, Schaul,
  Saxton, and Munos]{bellemare2016unifying}
Marc Bellemare, Sriram Srinivasan, Georg Ostrovski, Tom Schaul, David Saxton,
  and Remi Munos.
\newblock Unifying count-based exploration and intrinsic motivation.
\newblock In \emph{Advances in Neural Information Processing Systems}, 2016.

\bibitem[Blundell et~al.(2016)Blundell, Uria, Pritzel, Li, Ruderman, Leibo,
  Rae, Wierstra, and Hassabis]{blundell2016model}
Charles Blundell, Benigno Uria, Alexander Pritzel, Yazhe Li, Avraham Ruderman,
  Joel~Z Leibo, Jack Rae, Daan Wierstra, and Demis Hassabis.
\newblock Model-free episodic control.
\newblock \emph{arXiv preprint arXiv:1606.04460}, 2016.

\bibitem[Brockman et~al.(2016)Brockman, Cheung, Pettersson, Schneider,
  Schulman, Tang, and Zaremba]{brockman2016openai}
Greg Brockman, Vicki Cheung, Ludwig Pettersson, Jonas Schneider, John Schulman,
  Jie Tang, and Wojciech Zaremba.
\newblock Openai gym.
\newblock \emph{arXiv preprint arXiv:1606.01540}, 2016.

\bibitem[Burda et~al.(2018{\natexlab{a}})Burda, Edwards, Pathak, Storkey,
  Darrell, and Efros]{burda2018large}
Yuri Burda, Harri Edwards, Deepak Pathak, Amos Storkey, Trevor Darrell, and
  Alexei~A Efros.
\newblock Large-scale study of curiosity-driven learning.
\newblock In \emph{International Conference on Learning Representations},
  2018{\natexlab{a}}.

\bibitem[Burda et~al.(2018{\natexlab{b}})Burda, Edwards, Storkey, and
  Klimov]{burda2018exploration}
Yuri Burda, Harrison Edwards, Amos Storkey, and Oleg Klimov.
\newblock Exploration by random network distillation.
\newblock In \emph{International Conference on Learning Representations},
  2018{\natexlab{b}}.

\bibitem[Campero et~al.(2020)Campero, Raileanu, K{\"u}ttler, Tenenbaum,
  Rockt{\"a}schel, and Grefenstette]{campero2020learning}
Andres Campero, Roberta Raileanu, Heinrich K{\"u}ttler, Joshua~B Tenenbaum, Tim
  Rockt{\"a}schel, and Edward Grefenstette.
\newblock Learning with amigo: Adversarially motivated intrinsic goals.
\newblock \emph{arXiv preprint arXiv:2006.12122}, 2020.

\bibitem[Chevalier-Boisvert(2018)]{gym_miniworld}
Maxime Chevalier-Boisvert.
\newblock gym-miniworld environment for openai gym.
\newblock \url{https://github.com/maximecb/gym-miniworld}, 2018.

\bibitem[Chevalier-Boisvert et~al.(2018)Chevalier-Boisvert, Willems, and
  Pal]{chevalier2018minimalistic}
Maxime Chevalier-Boisvert, Lucas Willems, and Suman Pal.
\newblock Minimalistic gridworld environment for openai gym.
\newblock \emph{GitHub repository}, 2018.

\bibitem[Choi et~al.(2018)Choi, Guo, Moczulski, Oh, Wu, Norouzi, and
  Lee]{choi2018contingency}
Jongwook Choi, Yijie Guo, Marcin Moczulski, Junhyuk Oh, Neal Wu, Mohammad
  Norouzi, and Honglak Lee.
\newblock Contingency-aware exploration in reinforcement learning.
\newblock In \emph{International Conference on Learning Representations}, 2018.

\bibitem[Cobbe et~al.(2019)Cobbe, Hesse, Hilton, and
  Schulman]{cobbe2019leveraging}
Karl Cobbe, Christopher Hesse, Jacob Hilton, and John Schulman.
\newblock Leveraging procedural generation to benchmark reinforcement learning.
\newblock \emph{arXiv preprint arXiv:1912.01588}, 2019.

\bibitem[C{\^o}t{\'e} et~al.(2018)C{\^o}t{\'e}, K{\'a}d{\'a}r, Yuan, Kybartas,
  Barnes, Fine, Moore, Hausknecht, El~Asri, Adada, et~al.]{cote2018textworld}
Marc-Alexandre C{\^o}t{\'e}, {\'A}kos K{\'a}d{\'a}r, Xingdi Yuan, Ben Kybartas,
  Tavian Barnes, Emery Fine, James Moore, Matthew Hausknecht, Layla El~Asri,
  Mahmoud Adada, et~al.
\newblock Textworld: A learning environment for text-based games.
\newblock In \emph{Workshop on Computer Games}, 2018.

\bibitem[Du et~al.(2019)Du, Han, Fang, Liu, Dai, and Tao]{du2019liir}
Yali Du, Lei Han, Meng Fang, Ji~Liu, Tianhong Dai, and Dacheng Tao.
\newblock Liir: Learning individual intrinsic reward in multi-agent
  reinforcement learning.
\newblock In \emph{Advances in Neural Information Processing Systems}, 2019.

\bibitem[Duan et~al.(2016)Duan, Chen, Houthooft, Schulman, and
  Abbeel]{duan2016benchmarking}
Yan Duan, Xi~Chen, Rein Houthooft, John Schulman, and Pieter Abbeel.
\newblock Benchmarking deep reinforcement learning for continuous control.
\newblock In \emph{International Conference on Machine Learning}, 2016.

\bibitem[Fedus et~al.(2020)Fedus, Ramachandran, Agarwal, Bengio, Larochelle,
  Rowland, and Dabney]{fedus2020revisiting}
William Fedus, Prajit Ramachandran, Rishabh Agarwal, Yoshua Bengio, Hugo
  Larochelle, Mark Rowland, and Will Dabney.
\newblock Revisiting fundamentals of experience replay.
\newblock In \emph{International Conference on Machine Learning}, 2020.

\bibitem[Gangwani et~al.(2018)Gangwani, Liu, and Peng]{gangwani2018learning}
Tanmay Gangwani, Qiang Liu, and Jian Peng.
\newblock Learning self-imitating diverse policies.
\newblock In \emph{International Conference on Learning Representations}, 2018.

\bibitem[Guo et~al.(2016)Guo, Singh, Lewis, and Lee]{guo2016deep}
Xiaoxiao Guo, Satinder Singh, Richard Lewis, and Honglak Lee.
\newblock Deep learning for reward design to improve monte carlo tree search in
  atari games.
\newblock In \emph{International Joint Conference on Artificial Intelligence},
  2016.

\bibitem[Guo et~al.(2018)Guo, Oh, Singh, and Lee]{guo2018generative}
Yijie Guo, Junhyuk Oh, Satinder Singh, and Honglak Lee.
\newblock Generative adversarial self-imitation learning.
\newblock \emph{arXiv preprint arXiv:1812.00950}, 2018.

\bibitem[Guo et~al.(2019)Guo, Choi, Moczulski, Bengio, Norouzi, and
  Lee]{guo2019efficient}
Yijie Guo, Jongwook Choi, Marcin Moczulski, Samy Bengio, Mohammad Norouzi, and
  Honglak Lee.
\newblock Efficient exploration with self-imitation learning via
  trajectory-conditioned policy.
\newblock \emph{arXiv preprint arXiv:1907.10247}, 2019.

\bibitem[Henderson et~al.(2018)Henderson, Islam, Bachman, Pineau, Precup, and
  Meger]{henderson2018deep}
Peter Henderson, Riashat Islam, Philip Bachman, Joelle Pineau, Doina Precup,
  and David Meger.
\newblock Deep reinforcement learning that matters.
\newblock In \emph{AAAI Conference on Artificial Intelligence}, 2018.

\bibitem[Hessel et~al.(2018)Hessel, Modayil, Van~Hasselt, Schaul, Ostrovski,
  Dabney, Horgan, Piot, Azar, and Silver]{hessel2018rainbow}
Matteo Hessel, Joseph Modayil, Hado Van~Hasselt, Tom Schaul, Georg Ostrovski,
  Will Dabney, Dan Horgan, Bilal Piot, Mohammad Azar, and David Silver.
\newblock Rainbow: Combining improvements in deep reinforcement learning.
\newblock In \emph{AAAI Conference on Artificial Intelligence}, 2018.

\bibitem[Hester et~al.(2017)Hester, Vecerik, Pietquin, Lanctot, Schaul, Piot,
  Horgan, Quan, Sendonaris, Dulac-Arnold, et~al.]{hester2017deep}
Todd Hester, Matej Vecerik, Olivier Pietquin, Marc Lanctot, Tom Schaul, Bilal
  Piot, Dan Horgan, John Quan, Andrew Sendonaris, Gabriel Dulac-Arnold, et~al.
\newblock Deep q-learning from demonstrations.
\newblock 2017.

\bibitem[Ho \& Ermon(2016)Ho and Ermon]{ho2016generative}
Jonathan Ho and Stefano Ermon.
\newblock Generative adversarial imitation learning.
\newblock In \emph{Advances in neural information processing systems}, 2016.

\bibitem[K{\"u}ttler et~al.(2020)K{\"u}ttler, Nardelli, Miller, Raileanu,
  Selvatici, Grefenstette, and Rockt{\"a}schel]{kuttler2020nethack}
Heinrich K{\"u}ttler, Nantas Nardelli, Alexander~H Miller, Roberta Raileanu,
  Marco Selvatici, Edward Grefenstette, and Tim Rockt{\"a}schel.
\newblock The nethack learning environment.
\newblock \emph{arXiv preprint arXiv:2006.13760}, 2020.

\bibitem[Lai et~al.(2020)Lai, Zha, Li, and Hu]{lai2020dual}
Kwei-Herng Lai, Daochen Zha, Yuening Li, and Xia Hu.
\newblock Dual policy distillation.
\newblock In \emph{International Joint Conference on Artificial Intelligence},
  2020.

\bibitem[Lillicrap et~al.(2015)Lillicrap, Hunt, Pritzel, Heess, Erez, Tassa,
  Silver, and Wierstra]{lillicrap2015continuous}
Timothy~P Lillicrap, Jonathan~J Hunt, Alexander Pritzel, Nicolas Heess, Tom
  Erez, Yuval Tassa, David Silver, and Daan Wierstra.
\newblock Continuous control with deep reinforcement learning.
\newblock \emph{arXiv preprint arXiv:1509.02971}, 2015.

\bibitem[Lin(1992)]{lin1992self}
Long-Ji Lin.
\newblock Self-improving reactive agents based on reinforcement learning,
  planning and teaching.
\newblock \emph{Machine learning}, 8\penalty0 (3-4):\penalty0 293--321, 1992.

\bibitem[Lin(1993)]{lin1993reinforcement}
Long-Ji Lin.
\newblock Reinforcement learning for robots using neural networks.
\newblock Technical report, Carnegie-Mellon Univ Pittsburgh PA School of
  Computer Science, 1993.

\bibitem[Liu et~al.(2020)Liu, Yang, Chen, Zhang, Yang, Xiao, and
  Wang]{liu2020finrl}
Xiao-Yang Liu, Hongyang Yang, Qian Chen, Runjia Zhang, Liuqing Yang, Bowen
  Xiao, and Christina~Dan Wang.
\newblock Finrl: A deep reinforcement learning library for automated stock
  trading in quantitative finance.
\newblock \emph{arXiv preprint arXiv:2011.09607}, 2020.

\bibitem[Machado et~al.(2018)Machado, Bellemare, and Bowling]{machado2018count}
Marlos~C Machado, Marc~G Bellemare, and Michael Bowling.
\newblock Count-based exploration with the successor representation.
\newblock \emph{arXiv preprint arXiv:1807.11622}, 2018.

\bibitem[Martin et~al.(2017)Martin, Narayanan, Everitt, and
  Hutter]{martin2017count}
Jarryd Martin, S~Suraj Narayanan, Tom Everitt, and Marcus Hutter.
\newblock Count-based exploration in feature space for reinforcement learning.
\newblock In \emph{International Joint Conference on Artificial Intelligence},
  2017.

\bibitem[Mnih et~al.(2015)Mnih, Kavukcuoglu, Silver, Rusu, Veness, Bellemare,
  Graves, Riedmiller, Fidjeland, Ostrovski, et~al.]{mnih2015human}
Volodymyr Mnih, Koray Kavukcuoglu, David Silver, Andrei~A Rusu, Joel Veness,
  Marc~G Bellemare, Alex Graves, Martin Riedmiller, Andreas~K Fidjeland, Georg
  Ostrovski, et~al.
\newblock Human-level control through deep reinforcement learning.
\newblock \emph{nature}, 518\penalty0 (7540):\penalty0 529--533, 2015.

\bibitem[Mnih et~al.(2016)Mnih, Badia, Mirza, Graves, Lillicrap, Harley,
  Silver, and Kavukcuoglu]{mnih2016asynchronous}
Volodymyr Mnih, Adria~Puigdomenech Badia, Mehdi Mirza, Alex Graves, Timothy
  Lillicrap, Tim Harley, David Silver, and Koray Kavukcuoglu.
\newblock Asynchronous methods for deep reinforcement learning.
\newblock In \emph{International Conference on Machine Learning}, 2016.

\bibitem[Nichol et~al.(2018)Nichol, Pfau, Hesse, Klimov, and
  Schulman]{nichol2018gotta}
Alex Nichol, Vicki Pfau, Christopher Hesse, Oleg Klimov, and John Schulman.
\newblock Gotta learn fast: A new benchmark for generalization in rl.
\newblock \emph{arXiv preprint arXiv:1804.03720}, 2018.

\bibitem[Novati \& Koumoutsakos(2019)Novati and
  Koumoutsakos]{novati2019remember}
Guido Novati and Petros Koumoutsakos.
\newblock Remember and forget for experience replay.
\newblock In \emph{International Conference on Machine Learning}, 2019.

\bibitem[Oh et~al.(2018)Oh, Guo, Singh, and Lee]{oh2018self}
Junhyuk Oh, Yijie Guo, Satinder Singh, and Honglak Lee.
\newblock Self-imitation learning.
\newblock In \emph{International Conference on Machine Learning}, 2018.

\bibitem[Osband et~al.(2019)Osband, Van~Roy, Russo, and Wen]{osband2019deep}
Ian Osband, Benjamin Van~Roy, Daniel~J Russo, and Zheng Wen.
\newblock Deep exploration via randomized value functions.
\newblock \emph{Journal of Machine Learning Research}, 20\penalty0
  (124):\penalty0 1--62, 2019.

\bibitem[Ostrovski et~al.(2017)Ostrovski, Bellemare, Oord, and
  Munos]{ostrovski2017count}
Georg Ostrovski, Marc~G Bellemare, A{\"a}ron Oord, and R{\'e}mi Munos.
\newblock Count-based exploration with neural density models.
\newblock In \emph{International Conference on Machine Learning}, 2017.

\bibitem[Oudeyer \& Kaplan(2009)Oudeyer and Kaplan]{oudeyer2009intrinsic}
Pierre-Yves Oudeyer and Frederic Kaplan.
\newblock What is intrinsic motivation? a typology of computational approaches.
\newblock \emph{Frontiers in neurorobotics}, 1:\penalty0 6, 2009.

\bibitem[Pathak et~al.(2017)Pathak, Agrawal, Efros, and
  Darrell]{pathak2017curiosity}
Deepak Pathak, Pulkit Agrawal, Alexei~A Efros, and Trevor Darrell.
\newblock Curiosity-driven exploration by self-supervised prediction.
\newblock In \emph{IEEE Conference on Computer Vision and Pattern Recognition
  Workshops}, 2017.

\bibitem[Pritzel et~al.(2017)Pritzel, Uria, Srinivasan, Puigdomenech, Vinyals,
  Hassabis, Wierstra, and Blundell]{pritzel2017neural}
Alexander Pritzel, Benigno Uria, Sriram Srinivasan, Adria Puigdomenech, Oriol
  Vinyals, Demis Hassabis, Daan Wierstra, and Charles Blundell.
\newblock Neural episodic control.
\newblock \emph{arXiv preprint arXiv:1703.01988}, 2017.

\bibitem[Raileanu \& Rocktäschel(2020)Raileanu and
  Rocktäschel]{Raileanu2020RIDE}
Roberta Raileanu and Tim Rocktäschel.
\newblock Ride: Rewarding impact-driven exploration for procedurally-generated
  environments.
\newblock In \emph{International Conference on Learning Representations}, 2020.

\bibitem[Rajeswaran et~al.(2017)Rajeswaran, Lowrey, Todorov, and
  Kakade]{rajeswaran2017towards}
Aravind Rajeswaran, Kendall Lowrey, Emanuel~V Todorov, and Sham~M Kakade.
\newblock Towards generalization and simplicity in continuous control.
\newblock In \emph{Advances in Neural Information Processing Systems}, 2017.

\bibitem[Riedmiller et~al.(2018)Riedmiller, Hafner, Lampe, Neunert, Degrave,
  Wiele, Mnih, Heess, and Springenberg]{riedmiller2018learning}
Martin Riedmiller, Roland Hafner, Thomas Lampe, Michael Neunert, Jonas Degrave,
  Tom Wiele, Vlad Mnih, Nicolas Heess, and Jost~Tobias Springenberg.
\newblock Learning by playing solving sparse reward tasks from scratch.
\newblock In \emph{International Conference on Machine Learning}, 2018.

\bibitem[Savinov et~al.(2018)Savinov, Raichuk, Vincent, Marinier, Pollefeys,
  Lillicrap, and Gelly]{savinov2018episodic}
Nikolay Savinov, Anton Raichuk, Damien Vincent, Raphael Marinier, Marc
  Pollefeys, Timothy Lillicrap, and Sylvain Gelly.
\newblock Episodic curiosity through reachability.
\newblock In \emph{International Conference on Learning Representations}, 2018.

\bibitem[Schmidhuber(1991)]{schmidhuber1991possibility}
J{\"u}rgen Schmidhuber.
\newblock A possibility for implementing curiosity and boredom in
  model-building neural controllers.
\newblock In \emph{International conference on simulation of adaptive behavior:
  From animals to animats}, 1991.

\bibitem[Schulman et~al.(2017)Schulman, Wolski, Dhariwal, Radford, and
  Klimov]{schulman2017proximal}
John Schulman, Filip Wolski, Prafulla Dhariwal, Alec Radford, and Oleg Klimov.
\newblock Proximal policy optimization algorithms.
\newblock \emph{arXiv preprint arXiv:1707.06347}, 2017.

\bibitem[Silver et~al.(2016)Silver, Huang, Maddison, Guez, Sifre, Van
  Den~Driessche, Schrittwieser, Antonoglou, Panneershelvam, Lanctot,
  et~al.]{silver2016mastering}
David Silver, Aja Huang, Chris~J Maddison, Arthur Guez, Laurent Sifre, George
  Van Den~Driessche, Julian Schrittwieser, Ioannis Antonoglou, Veda
  Panneershelvam, Marc Lanctot, et~al.
\newblock Mastering the game of go with deep neural networks and tree search.
\newblock \emph{nature}, 529\penalty0 (7587):\penalty0 484--489, 2016.

\bibitem[Srivastava et~al.(2019)Srivastava, Shyam, Mutz, Ja{\'s}kowski, and
  Schmidhuber]{srivastava2019training}
Rupesh~Kumar Srivastava, Pranav Shyam, Filipe Mutz, Wojciech Ja{\'s}kowski, and
  J{\"u}rgen Schmidhuber.
\newblock Training agents using upside-down reinforcement learning.
\newblock \emph{arXiv preprint arXiv:1912.02877}, 2019.

\bibitem[Stadie et~al.(2015)Stadie, Levine, and
  Abbeel]{stadie2015incentivizing}
Bradly~C Stadie, Sergey Levine, and Pieter Abbeel.
\newblock Incentivizing exploration in reinforcement learning with deep
  predictive models.
\newblock \emph{arXiv preprint arXiv:1507.00814}, 2015.

\bibitem[Strehl \& Littman(2008)Strehl and Littman]{strehl2008analysis}
Alexander~L Strehl and Michael~L Littman.
\newblock An analysis of model-based interval estimation for markov decision
  processes.
\newblock \emph{Journal of Computer and System Sciences}, 74\penalty0
  (8):\penalty0 1309--1331, 2008.

\bibitem[Sukhbaatar et~al.(2018)Sukhbaatar, Lin, Kostrikov, Synnaeve, Szlam,
  and Fergus]{sukhbaatar2018intrinsic}
Sainbayar Sukhbaatar, Zeming Lin, Ilya Kostrikov, Gabriel Synnaeve, Arthur
  Szlam, and Rob Fergus.
\newblock Intrinsic motivation and automatic curricula via asymmetric
  self-play.
\newblock In \emph{International Conference on Learning Representations}, 2018.

\bibitem[Tang et~al.(2017)Tang, Houthooft, Foote, Stooke, Chen, Duan, Schulman,
  DeTurck, and Abbeel]{tang2017exploration}
Haoran Tang, Rein Houthooft, Davis Foote, Adam Stooke, OpenAI~Xi Chen, Yan
  Duan, John Schulman, Filip DeTurck, and Pieter Abbeel.
\newblock \# exploration: A study of count-based exploration for deep
  reinforcement learning.
\newblock In \emph{Advances in Neural Information Processing Systems}, 2017.

\bibitem[Todorov et~al.(2012)Todorov, Erez, and Tassa]{todorov2012mujoco}
Emanuel Todorov, Tom Erez, and Yuval Tassa.
\newblock Mujoco: A physics engine for model-based control.
\newblock In \emph{2012 IEEE/RSJ International Conference on Intelligent Robots
  and Systems}, 2012.

\bibitem[Xu et~al.(2018)Xu, Liu, Zhao, and Peng]{xu2018learning}
Tianbing Xu, Qiang Liu, Liang Zhao, and Jian Peng.
\newblock Learning to explore with meta-policy gradient.
\newblock In \emph{International Conference on Machine Learning}, 2018.

\bibitem[Zha et~al.(2019{\natexlab{a}})Zha, Lai, Cao, Huang, Wei, Guo, and
  Hu]{zha2019rlcard}
Daochen Zha, Kwei-Herng Lai, Yuanpu Cao, Songyi Huang, Ruzhe Wei, Junyu Guo,
  and Xia Hu.
\newblock Rlcard: A toolkit for reinforcement learning in card games.
\newblock \emph{arXiv preprint arXiv:1910.04376}, 2019{\natexlab{a}}.

\bibitem[Zha et~al.(2019{\natexlab{b}})Zha, Lai, Zhou, and
  Hu]{zha2019experience}
Daochen Zha, Kwei-Herng Lai, Kaixiong Zhou, and Xia Hu.
\newblock Experience replay optimization.
\newblock In \emph{International Joint Conference on Artificial Intelligence},
  2019{\natexlab{b}}.

\bibitem[Zhang et~al.(2018{\natexlab{a}})Zhang, Ballas, and
  Pineau]{zhang2018dissection}
Amy Zhang, Nicolas Ballas, and Joelle Pineau.
\newblock A dissection of overfitting and generalization in continuous
  reinforcement learning.
\newblock \emph{arXiv preprint arXiv:1806.07937}, 2018{\natexlab{a}}.

\bibitem[Zhang et~al.(2018{\natexlab{b}})Zhang, Vinyals, Munos, and
  Bengio]{zhang2018study}
Chiyuan Zhang, Oriol Vinyals, Remi Munos, and Samy Bengio.
\newblock A study on overfitting in deep reinforcement learning.
\newblock \emph{arXiv preprint arXiv:1804.06893}, 2018{\natexlab{b}}.

\bibitem[Zhang \& Sutton(2017)Zhang and Sutton]{zhang2017deeper}
Shangtong Zhang and Richard~S Sutton.
\newblock A deeper look at experience replay.
\newblock \emph{NIPS Deep Reinforcement Learning Symposium}, 2017.

\bibitem[Zheng et~al.(2018)Zheng, Oh, and Singh]{zheng2018learning}
Zeyu Zheng, Junhyuk Oh, and Satinder Singh.
\newblock On learning intrinsic rewards for policy gradient methods.
\newblock In \emph{Advances in Neural Information Processing Systems}, 2018.

\end{thebibliography}
\bibliographystyle{iclr2021_conference}

\newpage
\appendix

\section{Hyperparameters and Neural Network Architecture}
\label{sec:A}

For the proposed RAPID, the common hyperparameters are summarized in Table~\ref{tab:hyper}. RAPID is based on the PPO implementation from OpenAI baselines\footnote{\url{https://github.com/openai/baselines}}. The nstep is $128$ for all MiniGrid environments and MuJoCo environments, and $512$ for MiniWorld-Maze-S5. For MuJoCo environments, we report the results with only imitation loss since we find that the policy loss of PPO will harm the performance in the episodic reward setting. The learning rate is $10^{-4}$ for all MiniGrid environments and MiniWorld-Maze-S5, and $5 \times 10^{-4}$ for MuJoCo environments. Since MiniWorld-Maze-S5 and MuJoCo environments have continuous state space, we set $w_2=0$. In practice, in MiniWorld-Maze-S5, we observe that counting the states will usually lead to memory error because it is not likely to encounter the same state again in MiniWorld-Maze-S5 (COUNT is excluded in comparison due to memory issues). Note that it is possible to use pseudo-counts~\citep{bellemare2016unifying,ostrovski2017count},
which we will study in our future work. For MiniGrid-KeyCorridorS2R3-v0, MiniGrid-KeyCorridorS3R3-v0 and MiniWorld-Maze-S5, the update frequency of imitation learning is linearly annealed to $0$ throughout the training process. For other environments, the imitation learning is performed after each episode. We keep all other hyperparameters of PPO as default and use the default CNN architecture provided in OpenAI baselines for MiniWorld-Maze-S5 and MLP with 64-64 for other environments. The PPO also uses the same hyperparameters. We summarize the state space of each environment and how the local and global scores are computed in Table~\ref{tab:statescpace}.

\begin{table}[h!]
    \centering
    \begin{tabular}{l|l}
    \toprule
    \textbf{Hyperparameter} & \textbf{Value} \\
    \midrule
     $w_0$    &  $1$\\
     $w_1$    &  $0.1$\\
     $w_2$    &  $0.001$\\
     buffer Size    &  $10000$\\
     batch Size    &  $256$\\
     number of update steps    &  $5$\\
     \midrule
     entropy coefficient    &  $0.01$\\
     value function coefficient    &  $0.5$\\
     $\gamma$    &  $0.99$\\
     $\lambda$    &  $0.95$\\
     clip range    &  $0.2$\\

     \bottomrule
    \end{tabular}
    \caption{Common hyperparameters of the proposed RAPID (top) and the hyperparameters of PPO baseline (bottom).}
    \label{tab:hyper}
\end{table}

\begin{table}[h!]
    \centering
    \begin{tabular}{l|l|l|l}
    \toprule
    \textbf{Environment} & \textbf{State Space} & \textbf{Local Score} & \textbf{Global Score} \\
    \midrule
     8 MiniGrid Environments & $7 \times 7 \times 3$, discrete & Eq.~\ref{eqn:localdiscreate} & Eq.~\ref{eqn:global} \\
     MiniWorld Maze & $60 \times 80 \times 3$, continuous & Eq.~\ref{eqn:localcontinious} & No \\
     MuJoCo Walker2d & $17$, continuous & Eq.~\ref{eqn:localcontinious} & No \\
     MuJoCo Walker2d & $11$, continuous & Eq.~\ref{eqn:localcontinious} & No \\
     MuJoCo Walker2d & $4$, continuous & Eq.~\ref{eqn:localcontinious} & No \\
     MuJoCo Walker2d & $8$, continuous & Eq.~\ref{eqn:localcontinious} & No \\

     \bottomrule
    \end{tabular}
    \caption{The calculation of local and global scores for all the environments. Note that if the state space is continuous. We disable the global score since counting continuous states is meaningless. A possible solution is to use pseudo-counts~\citep{bellemare2016unifying,ostrovski2017count}.}
    \label{tab:statescpace}
\end{table}

For RANDOM, we implement the two networks with 64-64 MLP. For CURIOSITY and RIDE, we use 64-64 MLP for state embedding model, forward dynamics model and inverse dynamics model, respectively. However, with extensive hyperparameters search, we are not able to reproduce the RIDE results in MiniGrid environments even with the exactly same neural architecture and hyperparameters as suggested in~\citep{Raileanu2020RIDE}. Thus, we use the authors' implementation\footnote{\url{https://github.com/facebookresearch/impact-driven-exploration}}~\citep{Raileanu2020RIDE}, which is well-tuned on MiniGrid tasks. Our reported result is on par with that reported in~\citep{Raileanu2020RIDE}. For SIL and AMIGO, we use the implementations from the authors\footnote{\url{https://github.com/junhyukoh/self-imitation-learning}}\footnote{\url{https://github.com/facebookresearch/adversarially-motivated-intrinsic-goals}} with the recommended hyperparameters.

For the methods based on intrinsic rewards, i.e., RIDE,, AMIGO CURIOSITY,  RANDOM, and COUNT, we search intrinsic reward coefficient from $\{0.1, 0.05, 0.01, 0.005, 0.001, 0.0005, 0.0001$\}. For RIDE, the intrinsic reward coefficient is $0.1$ for MiniGrid-KeyCorridorS2R3-v0, MiniGrid-KeyCorridorS3R3-v0, MiniGrid-KeyCorridorS4R3-v0, MiniGrid-MultiRoom-N10S4-v0, MiniGrid-MultiRoom-N7S4-v0 and MiniWorld-Maze-S5, and $0.5$ for MiniGrid-MultiRoom-N7S8-v0, MiniGrid-MultiRoom-N10S10-v0 and MiniGrid-MultiRoom-N12S10-v0. For AMIGO and CURIOSITY, the intrinsic reward coefficient is $0.1$ for all the environments. For COUNT, the intrinsic reward coefficient is $0.005$ for all the environments. For RANDOM, the intrinsic reward coefficient is $0.001$ for all the environemnts. We also tune the entropy coefficient of RIDE from $\{0.001, 0.0005\}$, as suggested by the original paper. The entropy coefficient is $0.0005$ for MiniGrid-KeyCorridorS2R3-v0, MiniGrid-KeyCorridorS3R3-v0, MiniGrid-KeyCorridorS4R3-v0, MiniGrid-MultiRoom-N10S4-v0, MiniGrid-MultiRoom-N7S4-v0 and MiniWorld-Maze-S5, and $0.001$ for MiniGrid-MultiRoom-N7S8-v0, MiniGrid-MultiRoom-N10S10-v0 and MiniGrid-MultiRoom-N12S10-v0.

\begin{figure}[h!]
  \centering
    \includegraphics[width=0.5\textwidth]{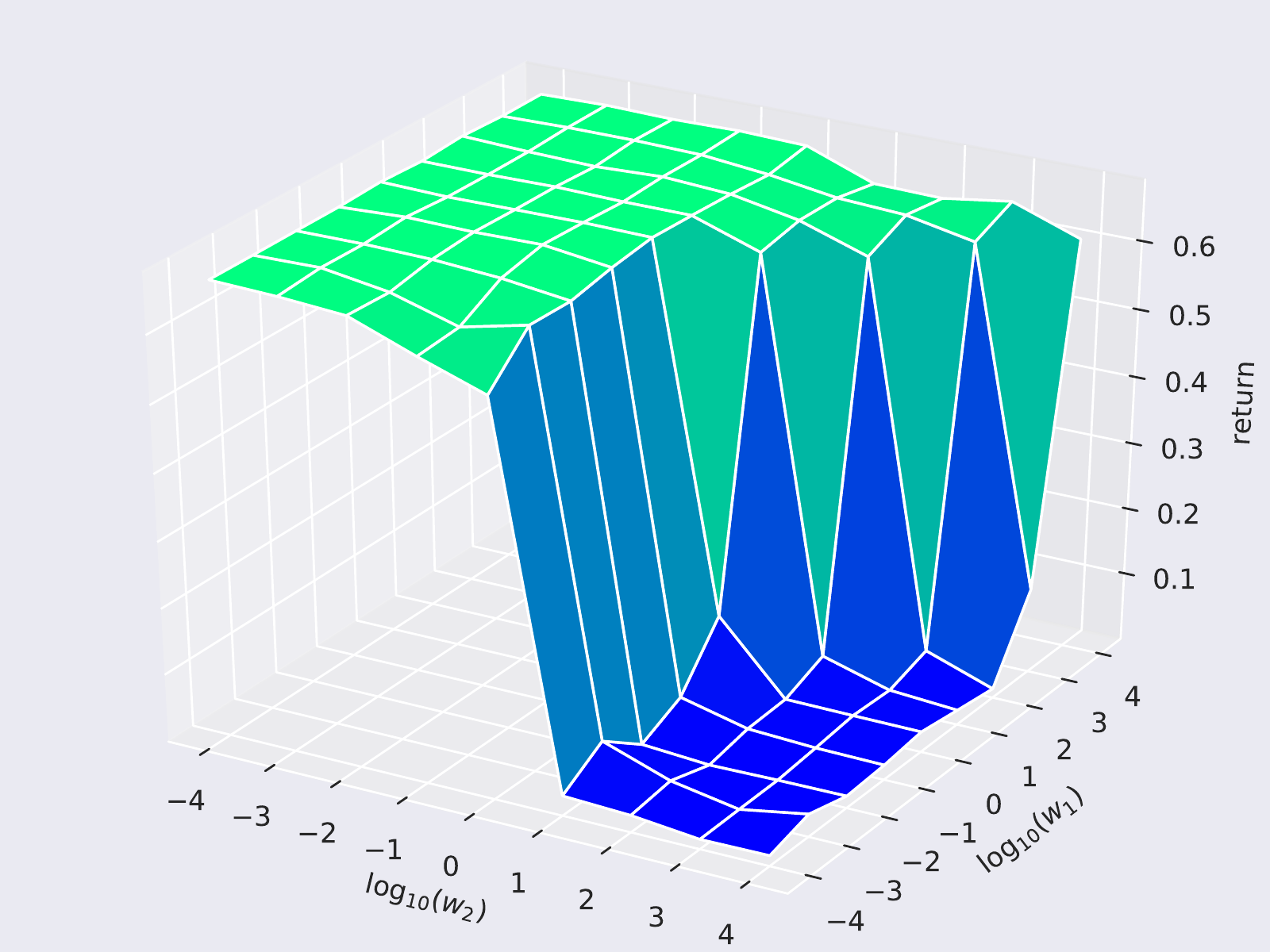}
  \caption{Sensitivity analysis of $w_1$ and $w_2$ ($w_0$ is fixed to 1) in MiniGrid-MultiRoom-N7S8-v0. All the experiments are run $3 \times 10^6$ timesteps. Note that $3 \times 10^6$ is more than enough for RAPID to converge in this environment~(see Figure~\ref{fig:minigrid}). The average results over 5 independent runs are plotted.}
  \label{fig:sensitivity}
\end{figure}

The $w_0$, $w_1$, and $w_2$ are selected as follows. We first run RAPID on MiniGrid-MultiRoom-N7S8-v0. We choose this environment because it is not too hard nor too easy, so we expect that it is representative among all the considered MiniGrid environments. We fix $w_0$ to be $1$, and select $w_1$ and $w_2$ from $\{10^{-4}, 10^{-3}, 10^{-2}, 10^{-1}, 10^{0}, 10^{1}, 10^{2}, 10^{3}, 10^{4}\}$, that is, there are $9 \times 9=81$ combinations. Note that we can fix $w_0$ because only the relative ranking matters in the ranking buffer. The selected $w_1$ and $w_2$ are 
then directly used in other environments without tuning. We further conduct a sensitivity analysis on MiniGrid-MultiRoom-N7S8-v0 in Figure~\ref{fig:sensitivity}. We observe that RAPID has good performance in a very wide range of hyperparameter choices.

\section{Environments}
\label{sec:B}
\subsection{MiniGrid Environments}

\begin{figure}[h!]
  \centering

  \begin{subfigure}[b]{0.25\textwidth}
    \centering
    \includegraphics[width=0.85\textwidth]{imgs/screenshots/minigrid/MiniGrid-MultiRoom-N12-S10-v0_0.png}
  \end{subfigure}%
  \begin{subfigure}[b]{0.25\textwidth}
    \centering
    \includegraphics[width=0.85\textwidth]{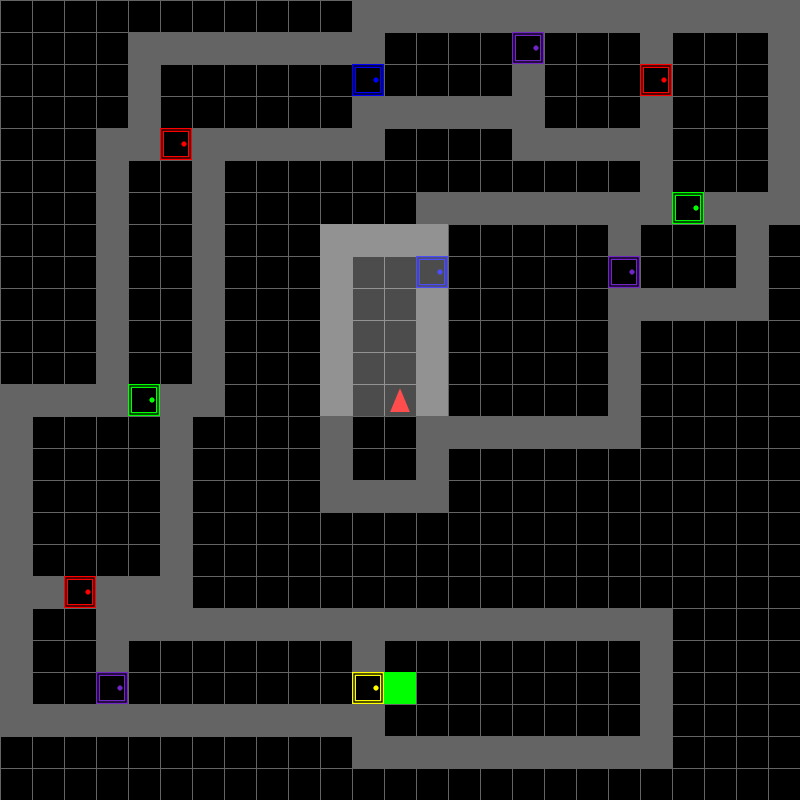}
  \end{subfigure}%
   \begin{subfigure}[b]{0.25\textwidth}
    \centering
    \includegraphics[width=0.85\textwidth]{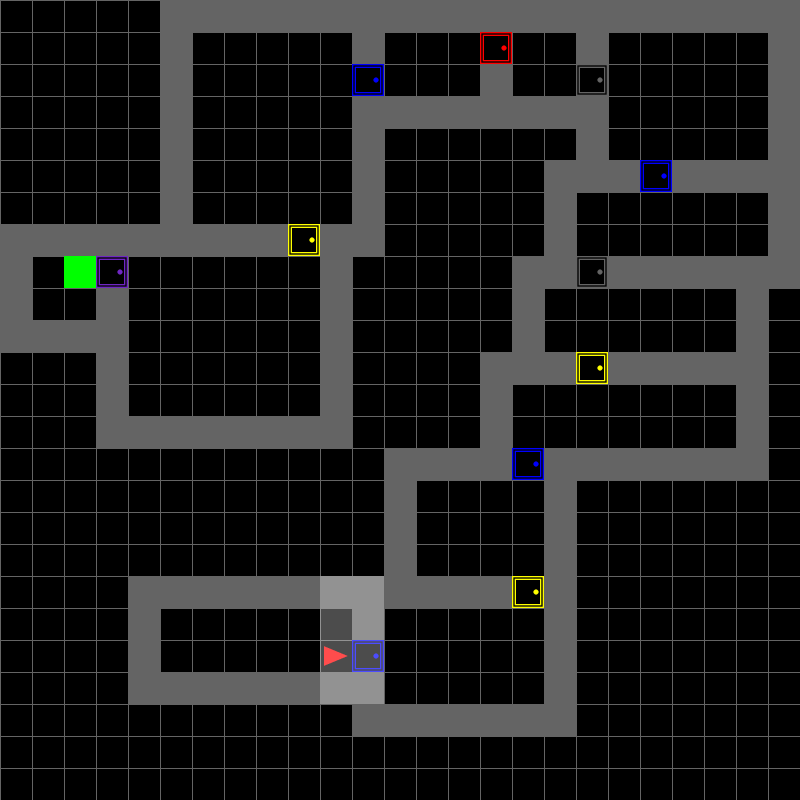}
  \end{subfigure}%
   \begin{subfigure}[b]{0.25\textwidth}
    \centering
    \includegraphics[width=0.85\textwidth]{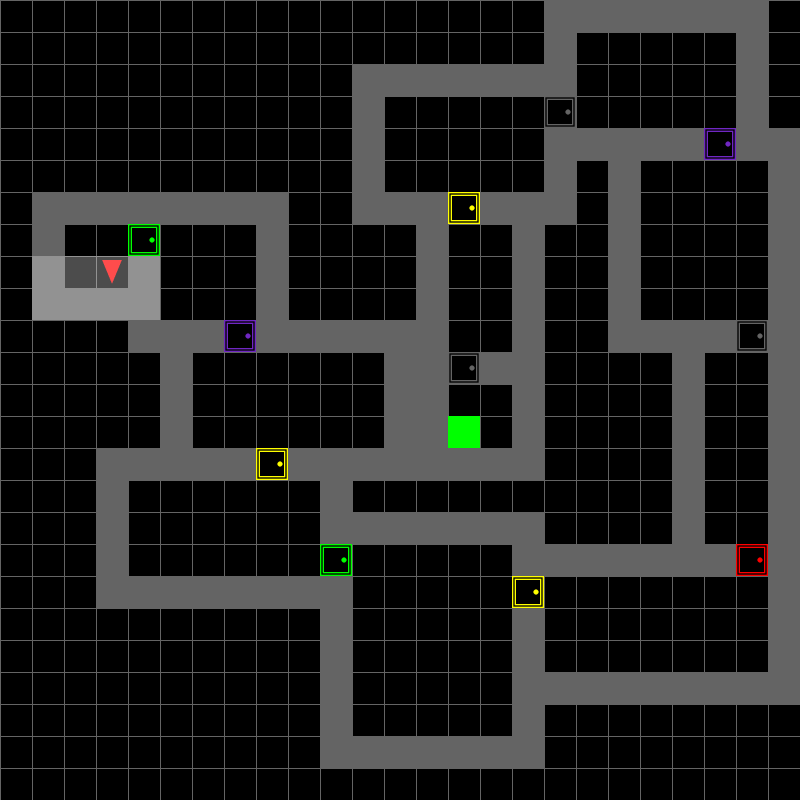}
  \end{subfigure}%
  
  \begin{subfigure}[b]{0.25\textwidth}
    \centering
    \includegraphics[width=0.85\textwidth]{imgs/screenshots/minigrid/MiniGrid-KeyCorridorS4R3-v0_0.png}
  \end{subfigure}%
  \begin{subfigure}[b]{0.25\textwidth}
    \centering
    \includegraphics[width=0.85\textwidth]{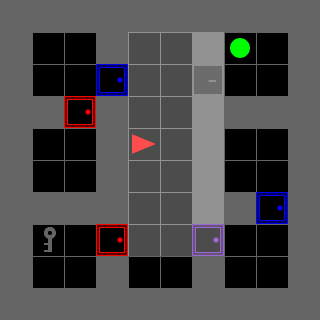}
  \end{subfigure}%
  \begin{subfigure}[b]{0.25\textwidth}
    \centering
    \includegraphics[width=0.85\textwidth]{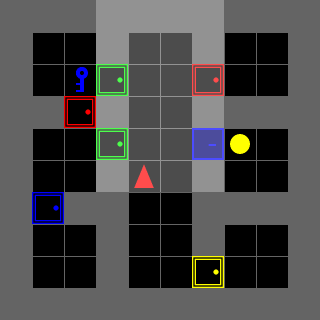}
  \end{subfigure}%
  \begin{subfigure}[b]{0.25\textwidth}
    \centering
    \includegraphics[width=0.85\textwidth]{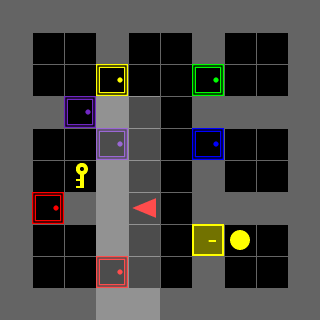}
  \end{subfigure}%
  \caption{Rendering of MultiRoom-N12-S10 (top row) and KeyCorridor-S4-R3(bottom row) in $4$ different episodes. The environments are procedually-generated, i.e., a different room is generated in a new episode.}
  \label{fig:minigridprocedurally}
\end{figure}

MiniGrid\footnote{\url{https://github.com/maximecb/gym-minigrid}} is a suite of light-weighted and fast gridworld environments with OpenAI gym interfaces. The environments are partially observable with a grid size of $N \times N$. Each tile in the gird can have at most one object, that is, $0$ or $1$ object. The possible objects are wall, floor, lava, door, key, ball, box and goal. Each object will have an associated color. The agent can pick up at most one object (ball or key). To open a locked door, the agent must use a key. Most of the environments in MiniGrid are procedurally-generated, i.e. a different grid will be sampled in each episode. Figure~\ref{fig:minigridprocedurally} shows the grids in four different episodes. The procedurally-generated nature makes training RL challenging because the RL agents need to learn the skills that can generalize to different grids. The environments can be easily modified so that we can create different levels of the environments. For example, we can modify the number of rooms and the size of each room to create hard-exploration or easy-exploration environments. The flexibility of MiniGrid environments enables us to conduct systematic comparison of algorithms under different difficulties.

The original observations of MiniGrid are dictionaries, which consist of a \emph{image} field providing a partially observable view of the environment, and a \emph{mission} field describing the goal with texts. In our experiments, we use \texttt{ImgObsWrapper} which only keeps the \emph{image} field. The environment uses a compact encoding with $3$ input values per visible grid cell. Note that the cells behind the wall or unopened doors are invisible.

The possible actions in MiniGrid are (1) turn left, (2) turn right, (3) move forward, (4) pick up an object, (5) drop the carried object, (6) open doors/interact with objects, and (7) done. The agent will remain in the same state if the action is not legal. For example, if there is no object in front of the agent, the agent will remain in the same state when performing action (4) pick up an object.

\begin{figure}[h!]
  \centering

  \begin{subfigure}[b]{0.25\textwidth}
    \centering
    \includegraphics[width=0.85\textwidth]{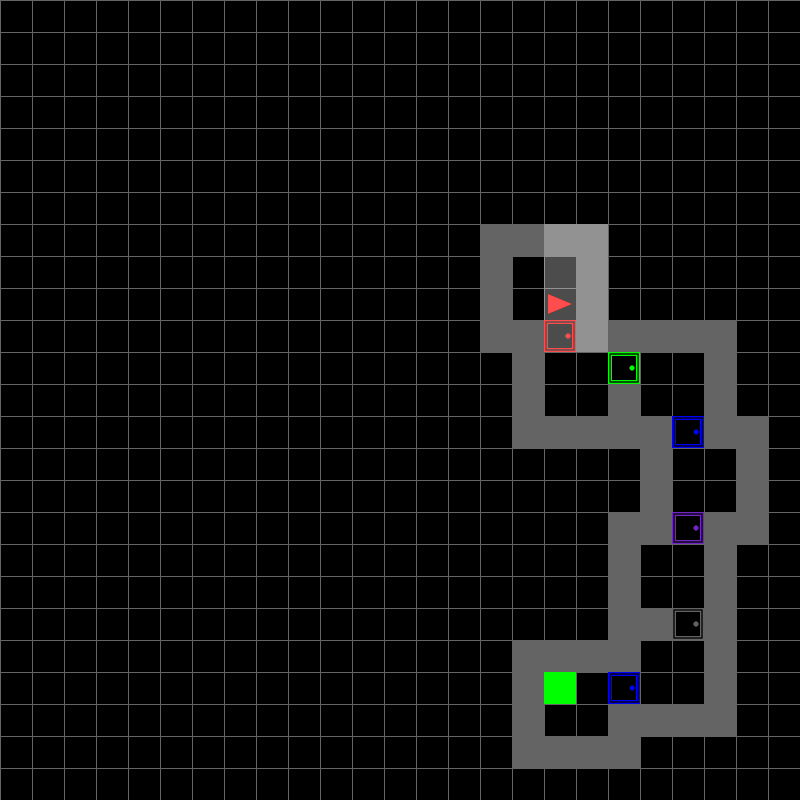}
    \caption{MultiRoom-N7-S4}
  \end{subfigure}%
  \begin{subfigure}[b]{0.25\textwidth}
    \centering
    \includegraphics[width=0.85\textwidth]{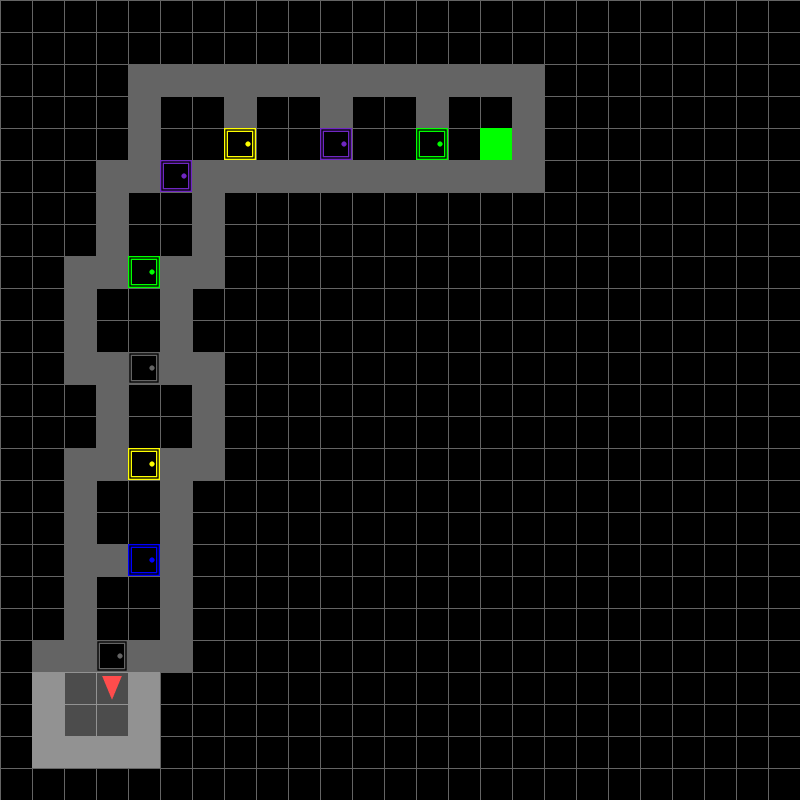}
    \caption{MultiRoom-N10-S4}
  \end{subfigure}%
   \begin{subfigure}[b]{0.25\textwidth}
    \centering
    \includegraphics[width=0.85\textwidth]{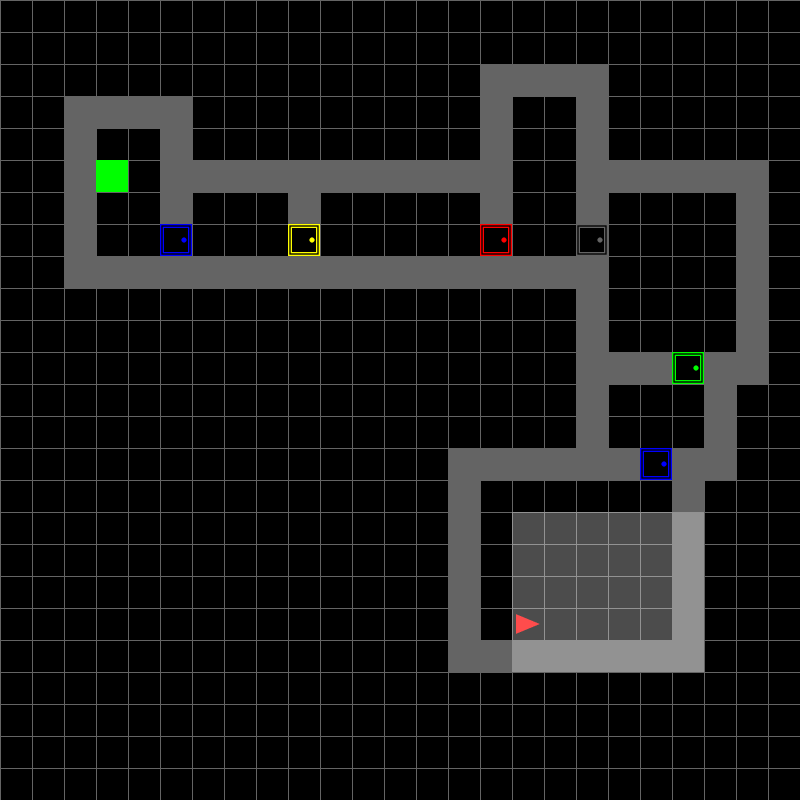}
    \caption{MultiRoom-N7-S8}
  \end{subfigure}%
   \begin{subfigure}[b]{0.25\textwidth}
    \centering
    \includegraphics[width=0.85\textwidth]{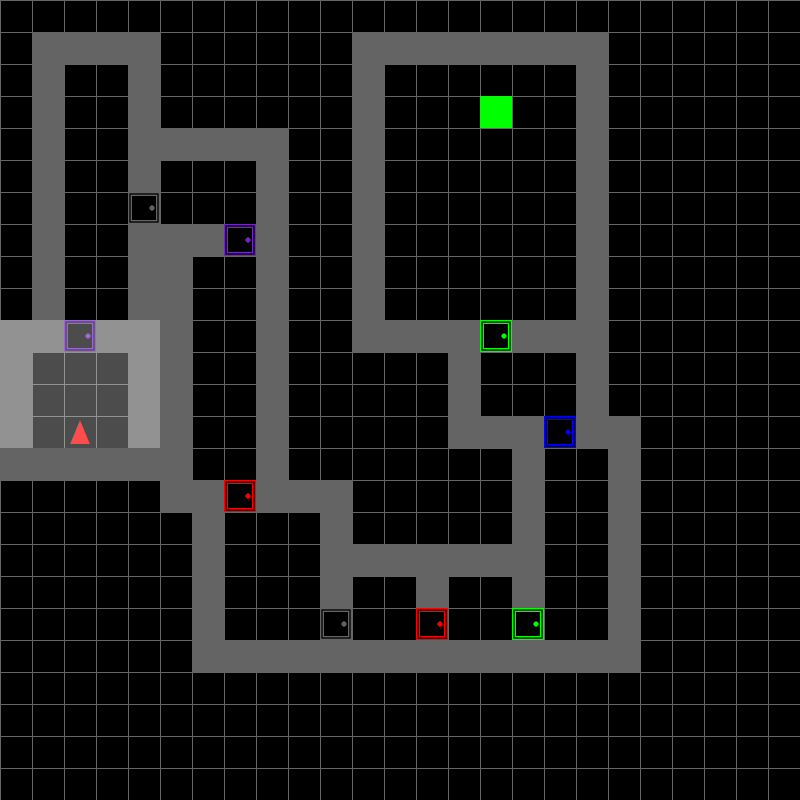}
    \caption{MultiRoom-N10-S10}
  \end{subfigure}%
  
  \begin{subfigure}[b]{0.25\textwidth}
    \centering
    \includegraphics[width=0.85\textwidth]{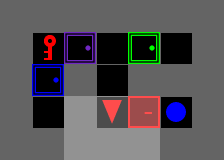}
    \caption{KeyCorridor-S3-R2}
  \end{subfigure}%
  \begin{subfigure}[b]{0.25\textwidth}
    \centering
    \includegraphics[width=0.85\textwidth]{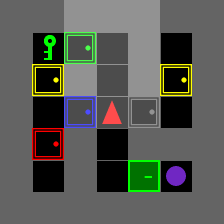}
    \caption{KeyCorridor-S3-R3}
  \end{subfigure}%
  \begin{subfigure}[b]{0.25\textwidth}
    \centering
    \includegraphics[width=0.85\textwidth]{imgs/screenshots/minigrid/MiniGrid-KeyCorridorS4R3-v0_0.png}
    \caption{KeyCorridor-S4-R3}
  \end{subfigure}%
  \begin{subfigure}[b]{0.25\textwidth}
    \centering
    \includegraphics[width=0.85\textwidth]{imgs/screenshots/minigrid/MiniGrid-MultiRoom-N12-S10-v0_0.png}
    \caption{MultiRoom-N12-S10}
  \end{subfigure}%
  \caption{Rendering of Minigird environments used in the work.}
  \label{fig:minigridrender}
\end{figure}

We focus on MultiRoom-NX-SY and KeyCorridor-SX-RY, where X and Y are hyperparameters specifying the number of rooms or the room sizes in the environments. With larger X and Y, the environments will become more difficult to solve due to the sparse rewards. Figure~\ref{fig:minigridrender} shows the environments (with different difficulties) in this work. In MultiRoom-NX-SY, the agent needs to navigate the green tile in the last room. If the agent successfully navigates the goal, the agent will receive a reward of $1$ subtracting some penalties of the timesteps spent. In KeyCorridor-SX-RY, the agent needs to pick up the key, use the key to open the locked door, and pick up the ball. Similarly, the agent will receive a reward of $1$ subtracting some penalties of the timesteps spent if it picks up the ball. Both environments are extremely difficult to solve using RL alone due to the sparse rewards.

\subsection{MiniWorld Maze Environment}
MiniWorld\footnote{\url{https://github.com/maximecb/gym-miniworld}} is a minimalistic 3D interior environment simulator. Similar to Minigrid, we can easily create environments with different levels. We focus on MiniWorld-Maze, where the agent is asked to navigate the goal through a procedurally-generated maze. The agent has a first-person partially observable view of the environment. This environment is extremely difficult to solve due to sparse reward, long-horizon, and the procedurally-generated nature of the environments.

\begin{figure}[h!]
  \centering
  
  \begin{subfigure}[b]{0.25\textwidth}
    \centering
    \includegraphics[width=0.85\textwidth]{imgs/screenshots/miniworld/top_0.png}
  \end{subfigure}%
  \begin{subfigure}[b]{0.25\textwidth}
    \centering
    \includegraphics[width=0.85\textwidth]{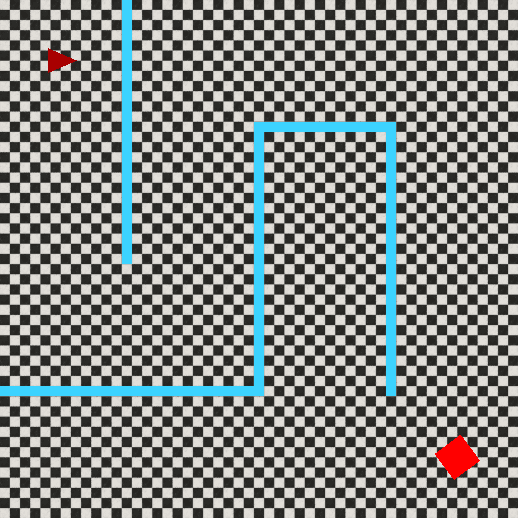}
  \end{subfigure}%
  \begin{subfigure}[b]{0.25\textwidth}
    \centering
    \includegraphics[width=0.85\textwidth]{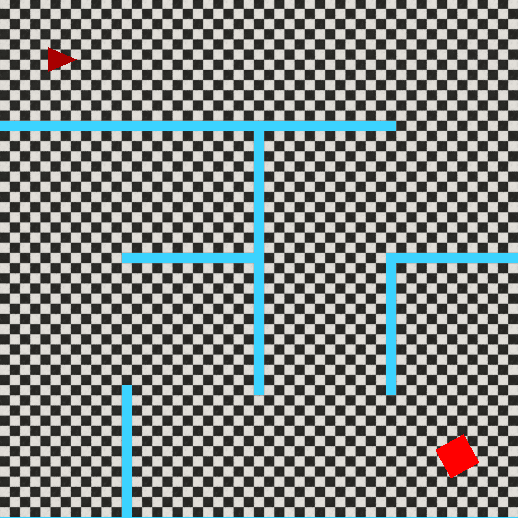}
  \end{subfigure}%
  \begin{subfigure}[b]{0.25\textwidth}
    \includegraphics[width=0.85\textwidth]{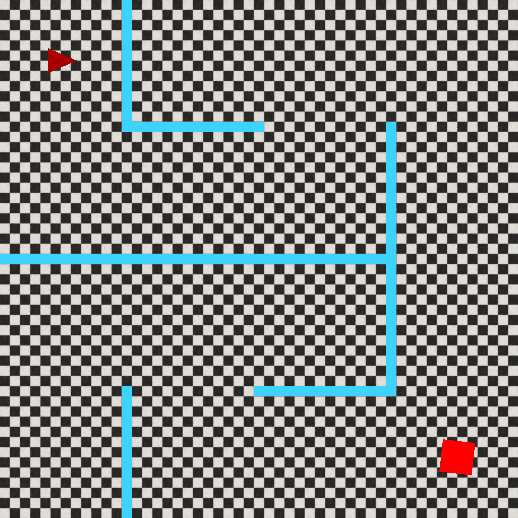}
  \end{subfigure}%
  \caption{Example generated mazes in $4$ diffident episodes.}
  \label{fig:mazeprocedurally}
\end{figure}

In this work, we focus on MiniWorld-Maze-S5, a variant of the MiniWorld Maze environment with $5 \times 5$ tiles. Figure~\ref{fig:mazeprocedurally} shows the top view of the mazes in four different episodes. In each episode, the agent and the goal are initialized in the top-left and bottom-right corners of the map, respectively. In this way, we ensure that the agent and the goal are far away enough so that the agent can not easily see the goal without exploration. A random maze will be generated in each episode. There are three possible actions: (1) move forward, (2) turn left, and (3) turn right. The agent will move $0.4 \times tile\_length$ if moving forward, where $tile\_length$ is the length of one side of a tile. The agent will rotate $22.5$ degrees if turning right/left. The time budget of each episode is $600$. The agent will not receive any positive reward if it can not navigate the goal under the time budget.

\subsection{Sparse MuJoCo Environments}
\begin{figure}[h!]
  \centering
  
  \begin{subfigure}[b]{0.25\textwidth}
    \centering
    \includegraphics[width=0.85\textwidth]{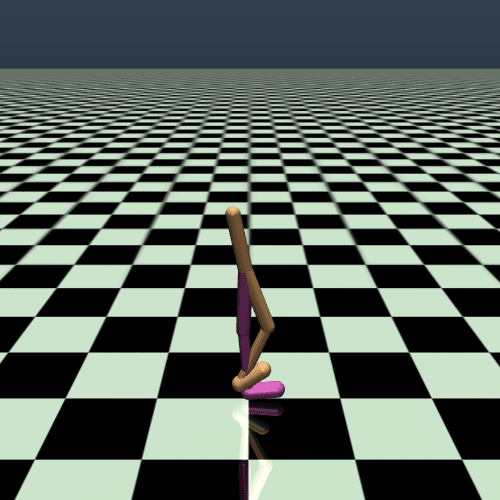}
    \caption{Walker2d}
  \end{subfigure}%
  \begin{subfigure}[b]{0.25\textwidth}
    \centering
    \includegraphics[width=0.85\textwidth]{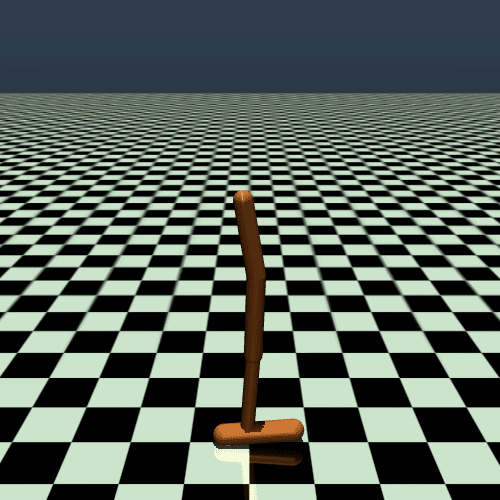}
    \caption{Hopper}
  \end{subfigure}%
  \begin{subfigure}[b]{0.25\textwidth}
    \centering
    \includegraphics[width=0.85\textwidth]{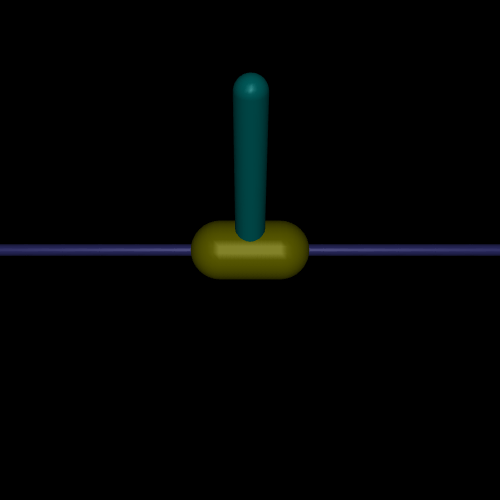}
    \caption{InvertedPendulum}
  \end{subfigure}%
  \begin{subfigure}[b]{0.25\textwidth}
    \centering
    \includegraphics[width=0.85\textwidth]{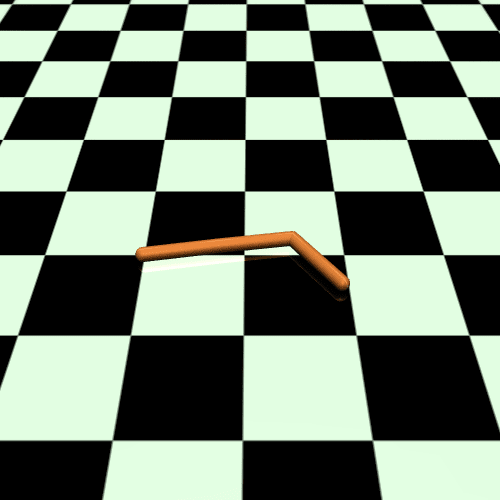}
    \caption{Swimmer}
  \end{subfigure}%
  \caption{Rendering of Mujoco environments.}
  \label{fig:mujocorender}
\end{figure}

Mujoco is a physical engine for continuous control~\citep{todorov2012mujoco}. Figure~\ref{fig:mujocorender} shows the rendering of the four MuJoCo tasks used in our work. The original MuJoCo environments use dense rewards, i.e., a well-defined reward is given in each timestep. However, in many scenarios, well-defined dense rewards may be not available. We consider a variant of MuJoCo task using only episodic reward. Specifically, the original rewards are accumulated and only an episodic reward is given in the final timestep. The agent will receive a $0$ reward in all the other timesteps. This sparse setting makes the tasks much more challenging to solve. Contemporary RL algorithms will usually suffer from the sparse rewards and deliver unsatisfactory performance. We aim to use these sparse variants to study whether our RAPID can effectively capture the sparse rewards.

\section{Ablations}
\label{sec:C}
\begin{figure}[h!]
  \centering
  \begin{subfigure}[b]{0.65\textwidth}
    \includegraphics[width=1.0\textwidth]{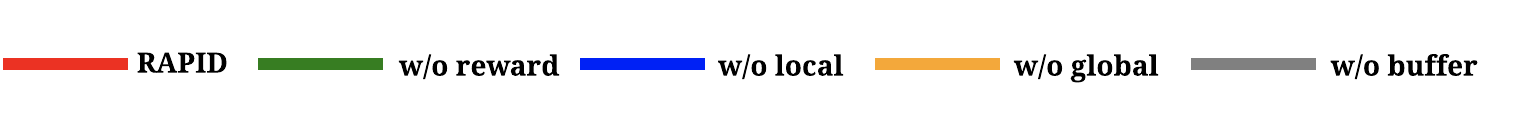}
  \end{subfigure}

  \begin{subfigure}[b]{0.25\textwidth}
    \centering
    \includegraphics[width=0.99\textwidth]{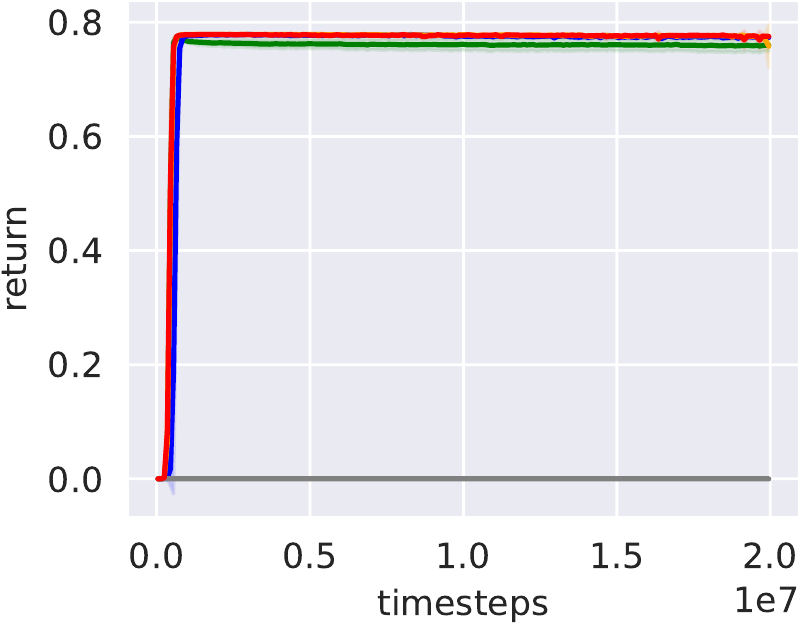}
    \caption{MultiRoom-N7-S4}
  \end{subfigure}%
  \begin{subfigure}[b]{0.25\textwidth}
    \centering
    \includegraphics[width=0.99\textwidth]{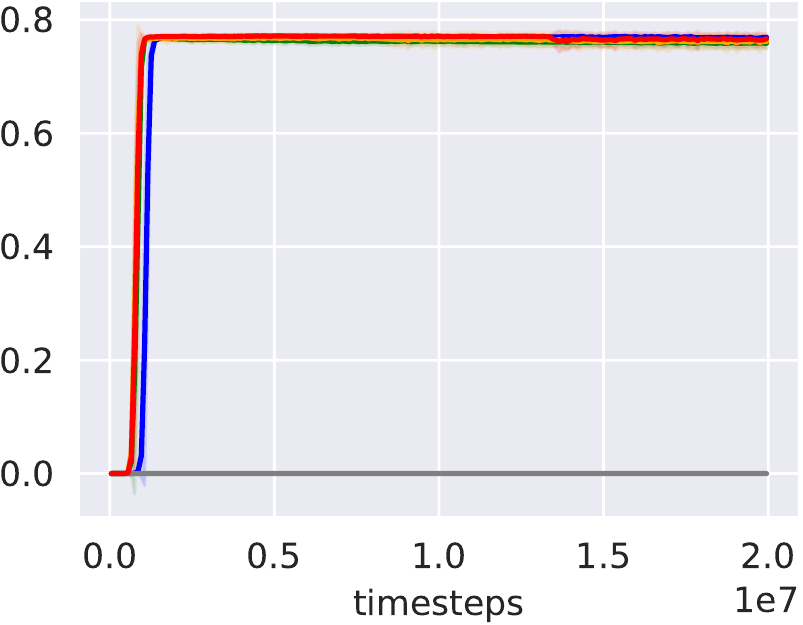}
    \caption{MultiRoom-N10-S4}
  \end{subfigure}%
  \begin{subfigure}[b]{0.25\textwidth}
    \centering
    \includegraphics[width=0.99\textwidth]{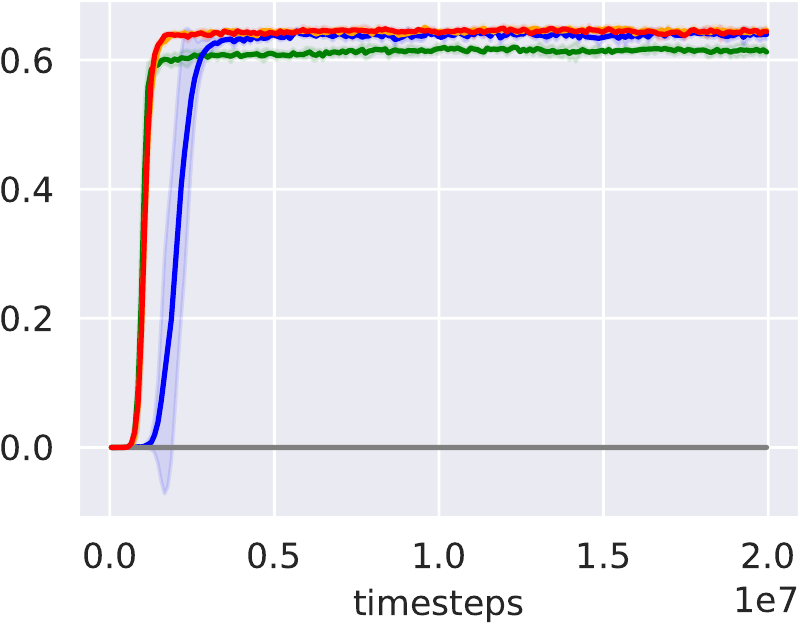}
    \caption{MultiRoom-N7-S8}
  \end{subfigure}%
  \begin{subfigure}[b]{0.25\textwidth}
    \centering
    \includegraphics[width=0.99\textwidth]{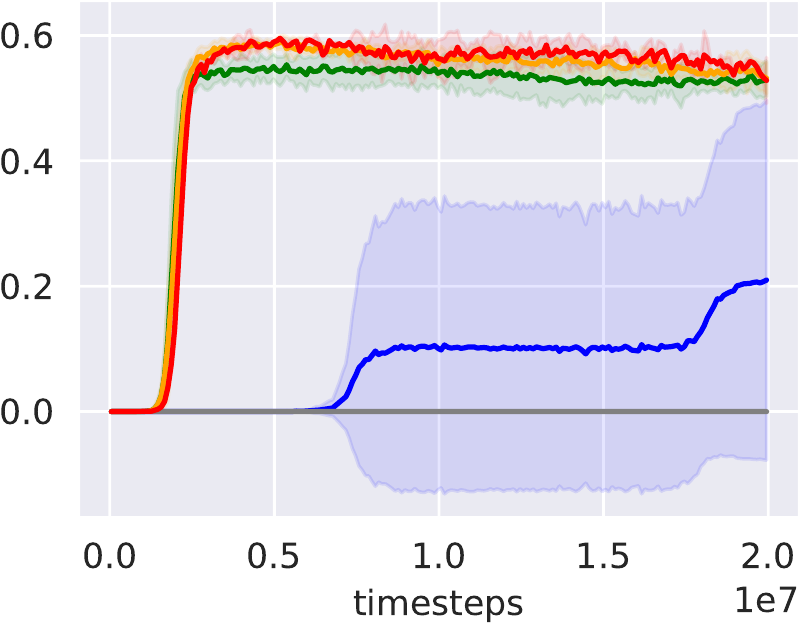}
    \caption{MultiRoom-N10-S10}
  \end{subfigure}%
  
  \begin{subfigure}[b]{0.25\textwidth}
    \centering
    \includegraphics[width=0.99\textwidth]{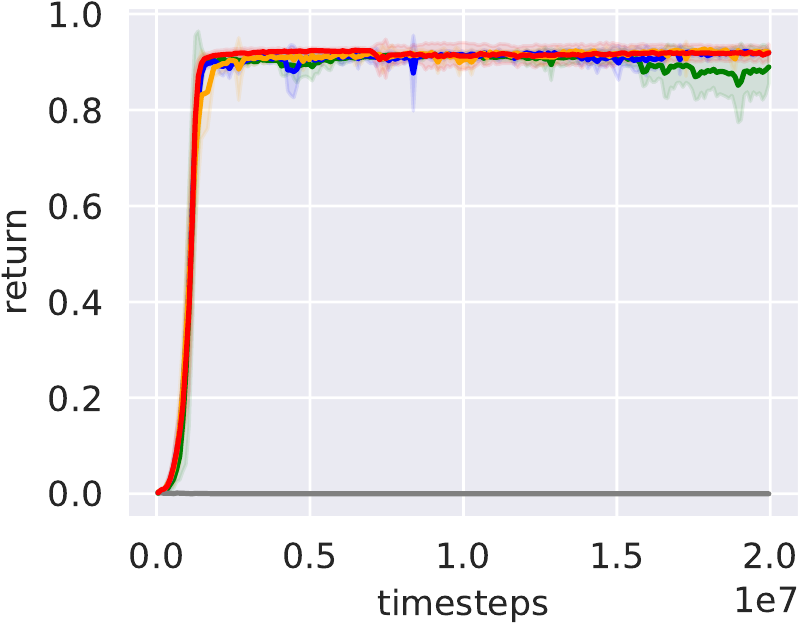}
    \caption{KeyCorridor-S3-R2}
  \end{subfigure}%
  \begin{subfigure}[b]{0.25\textwidth}
    \centering
    \includegraphics[width=0.99\textwidth]{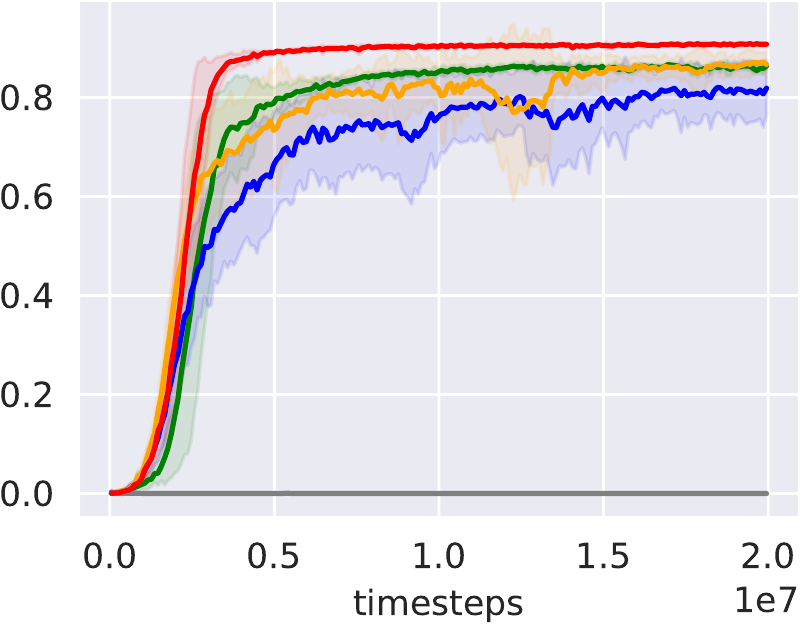}
    \caption{KeyCorridor-S3-R3}
  \end{subfigure}%
  \begin{subfigure}[b]{0.25\textwidth}
    \centering
    \includegraphics[width=0.99\textwidth]{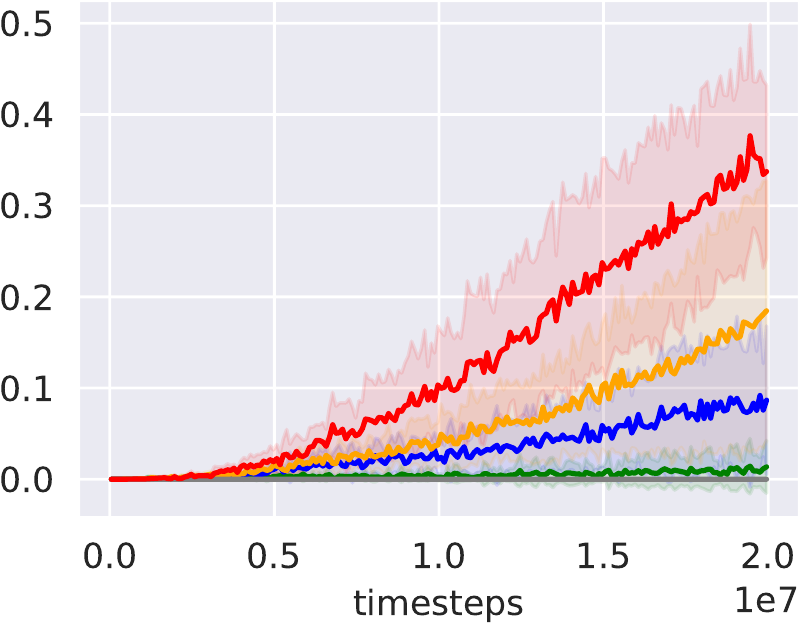}
    \caption{KeyCorridor-S4-R3}
  \end{subfigure}%
  \begin{subfigure}[b]{0.25\textwidth}
    \centering
    \includegraphics[width=0.99\textwidth]{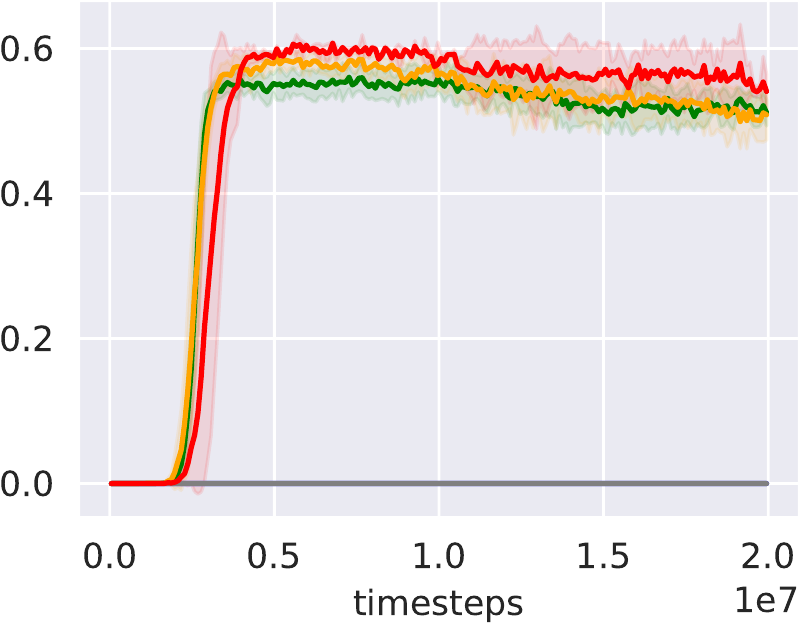}
    \caption{MultiRoom-N12-S10}
  \end{subfigure}%
  \caption{Learning curves of RAPID and the ablations}
  \label{fig:ablationcurve}
\end{figure}

Figure~\ref{fig:ablationcurve} shows the learning curves of RAPID against different ablations. Without local scores, we observe significant performance drop on challenging environments in terms of sample efficiency and final performance, i.e., on MultiRoom-N7-S8, MultiRoom-N10-S10, MultiRoom-N12-S10, KeyCorridor-S3-R3, and KeyCorridor-S4-R3. This suggests that the proposed local score plays an important role in encouraging exploration. We also observe that without buffer, the agent fails to learn a good policy. Naively using the proposed scores as intrinsic reward will harm the performance. Using the extrinsic reward and the global score to rank the episodes also contribute in the challenging environments, such as KeyCorridor-S4-R3. Therefore, the three proposed scoring methods may be complementary and play different roles in different environments.

\section{Full Analysis Results}
\label{sec:D}
\subsection{Impact of the Number of Update Steps}

\begin{figure}[h!]
  \centering
  \begin{subfigure}[b]{0.45\textwidth}
    \includegraphics[width=1.0\textwidth]{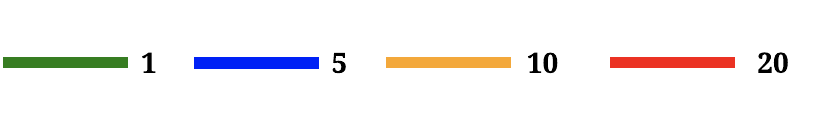}
  \end{subfigure}

  \begin{subfigure}[b]{0.25\textwidth}
    \centering
    \includegraphics[width=0.99\textwidth]{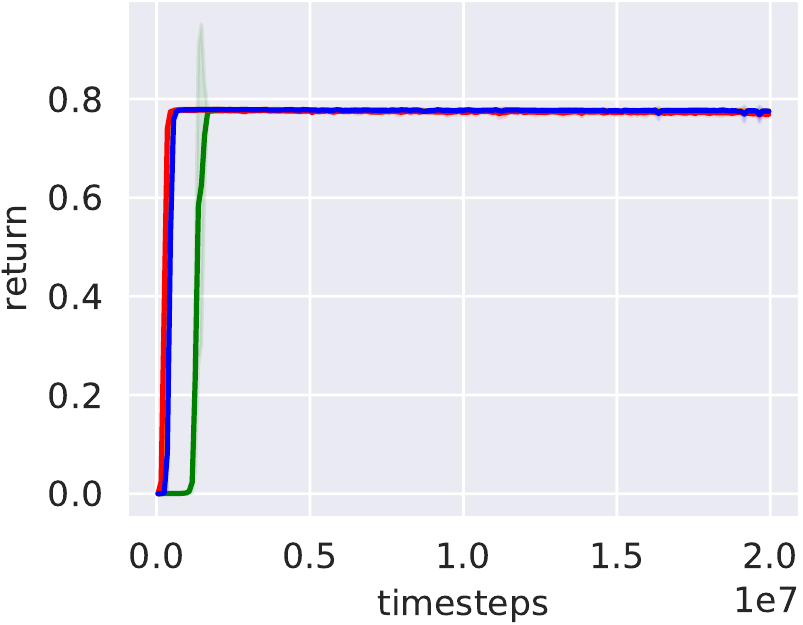}
    \caption{MultiRoom-N7-S4}
  \end{subfigure}%
  \begin{subfigure}[b]{0.25\textwidth}
    \centering
    \includegraphics[width=0.99\textwidth]{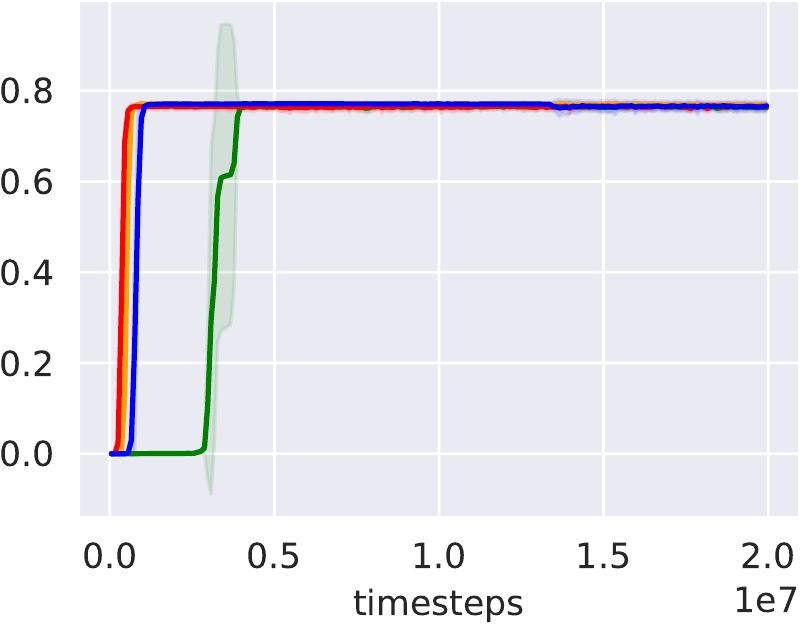}
    \caption{MultiRoom-N10-S4}
  \end{subfigure}%
  \begin{subfigure}[b]{0.25\textwidth}
    \centering
    \includegraphics[width=0.99\textwidth]{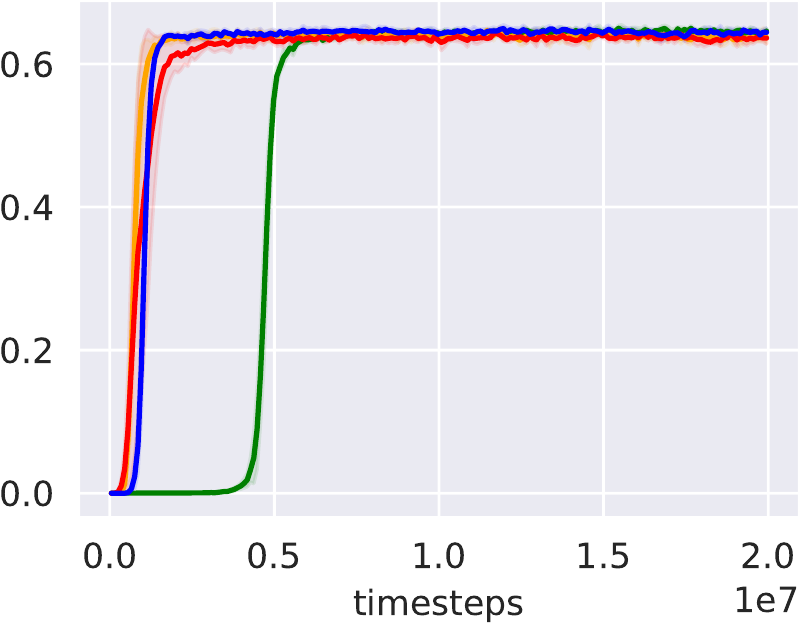}
    \caption{MultiRoom-N7-S8}
  \end{subfigure}%
  \begin{subfigure}[b]{0.25\textwidth}
    \centering
    \includegraphics[width=0.99\textwidth]{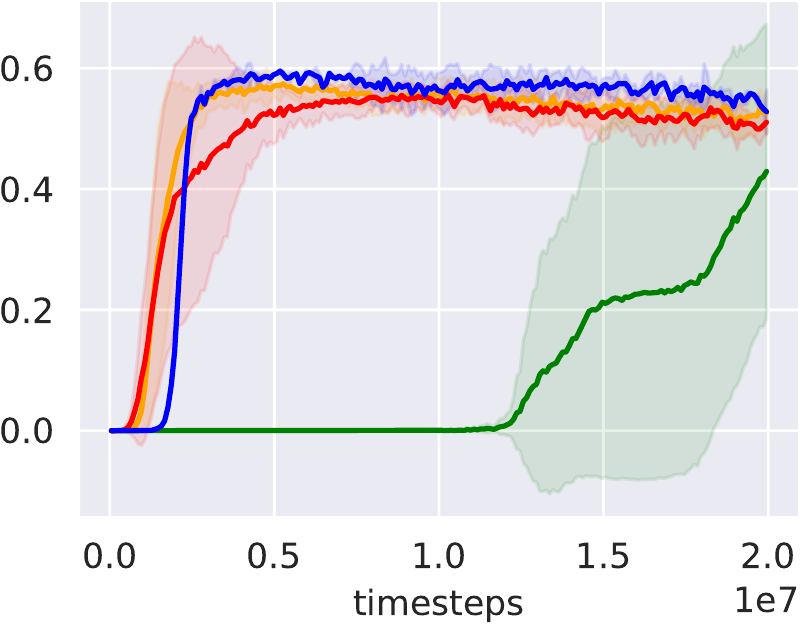}
    \caption{MultiRoom-N10-S10}
  \end{subfigure}%
  
  \begin{subfigure}[b]{0.25\textwidth}
    \centering
    \includegraphics[width=0.99\textwidth]{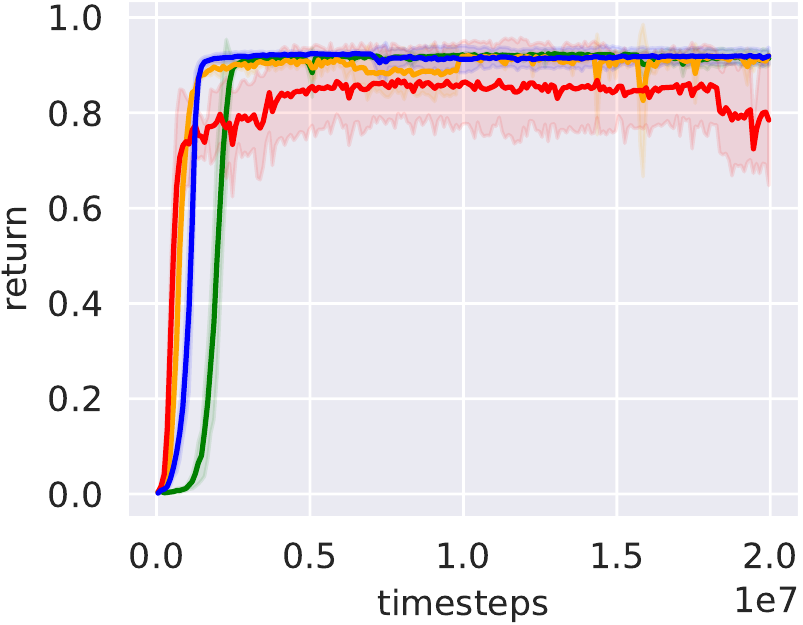}
    \caption{KeyCorridor-S3-R2}
  \end{subfigure}%
  \begin{subfigure}[b]{0.25\textwidth}
    \centering
    \includegraphics[width=0.99\textwidth]{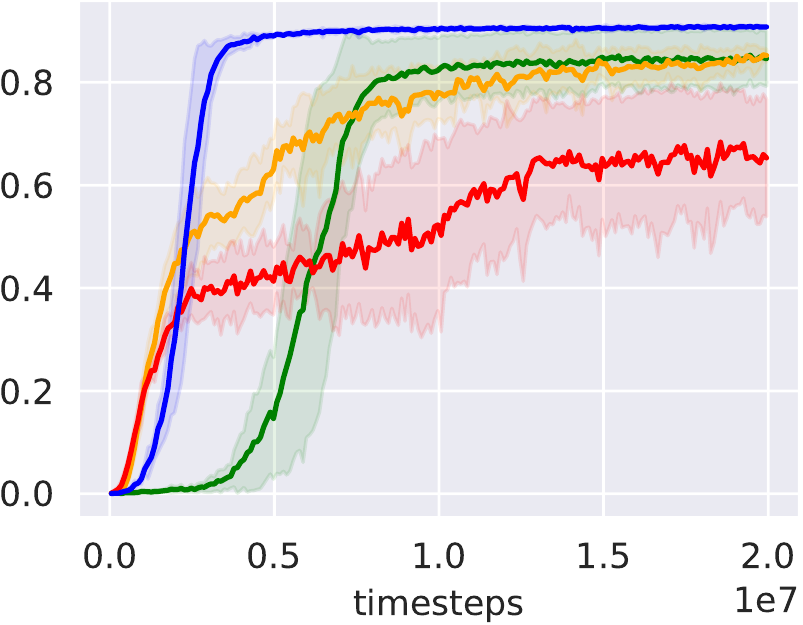}
    \caption{KeyCorridor-S3-R3}
  \end{subfigure}%
  \begin{subfigure}[b]{0.25\textwidth}
    \centering
    \includegraphics[width=0.99\textwidth]{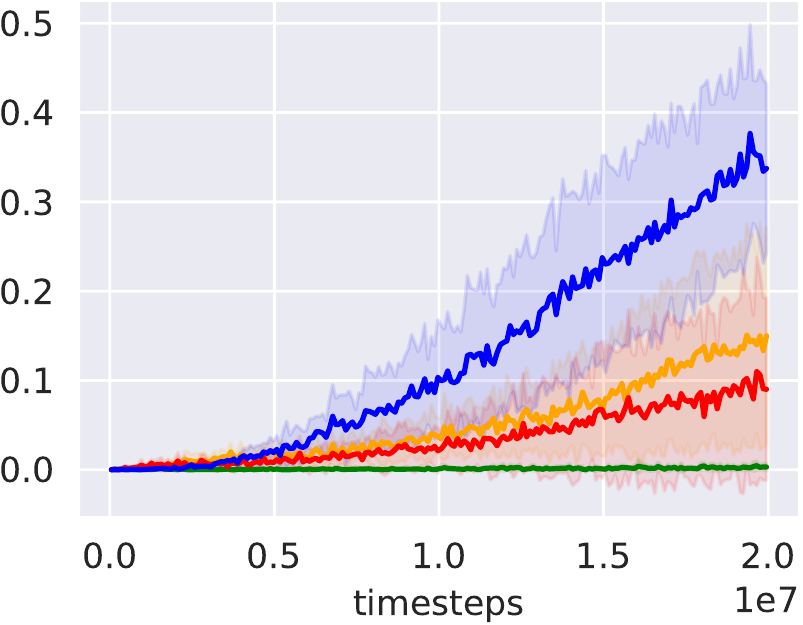}
    \caption{KeyCorridor-S4-R3}
  \end{subfigure}%
  \begin{subfigure}[b]{0.25\textwidth}
    \centering
    \includegraphics[width=0.99\textwidth]{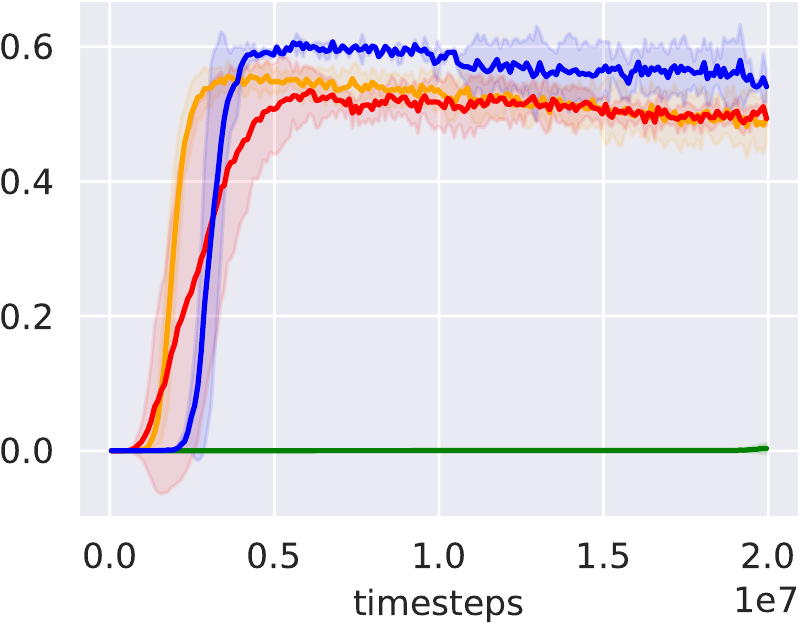}
    \caption{MultiRoom-N12-S10}
  \end{subfigure}%
  \caption{Impact of training steps $S$ on MiniGrid environments.}
\end{figure}

\newpage
\subsection{Impact of Buffer Size}

\begin{figure}[h!]
  \centering
  \begin{subfigure}[b]{0.55\textwidth}
    \includegraphics[width=1.0\textwidth]{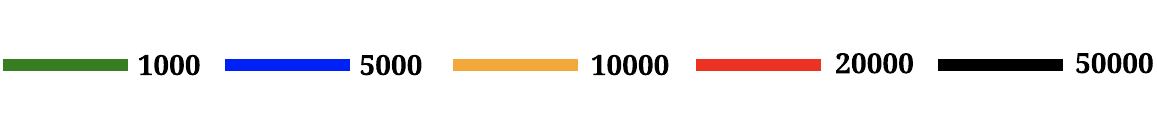}
  \end{subfigure}

  \begin{subfigure}[b]{0.25\textwidth}
    \centering
    \includegraphics[width=0.99\textwidth]{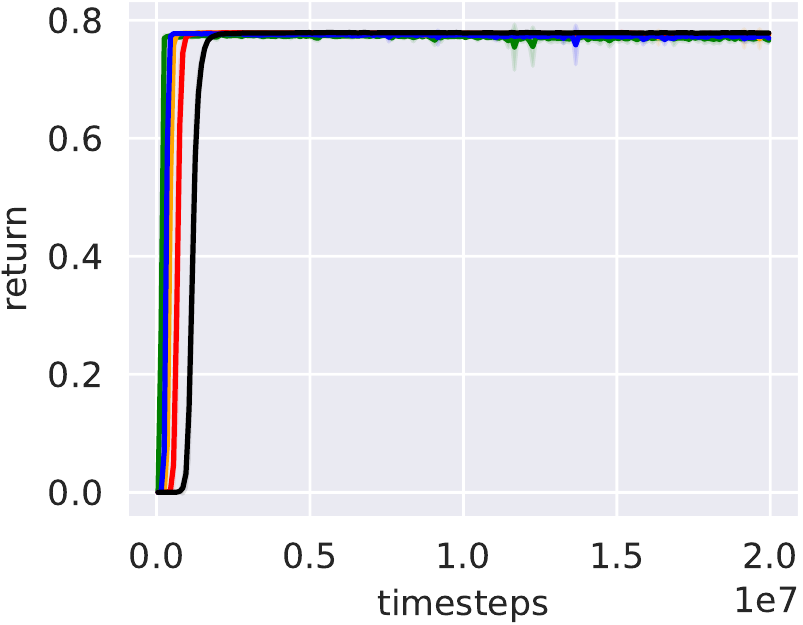}
    \caption{MultiRoom-N7-S4}
  \end{subfigure}%
  \begin{subfigure}[b]{0.25\textwidth}
    \centering
    \includegraphics[width=0.99\textwidth]{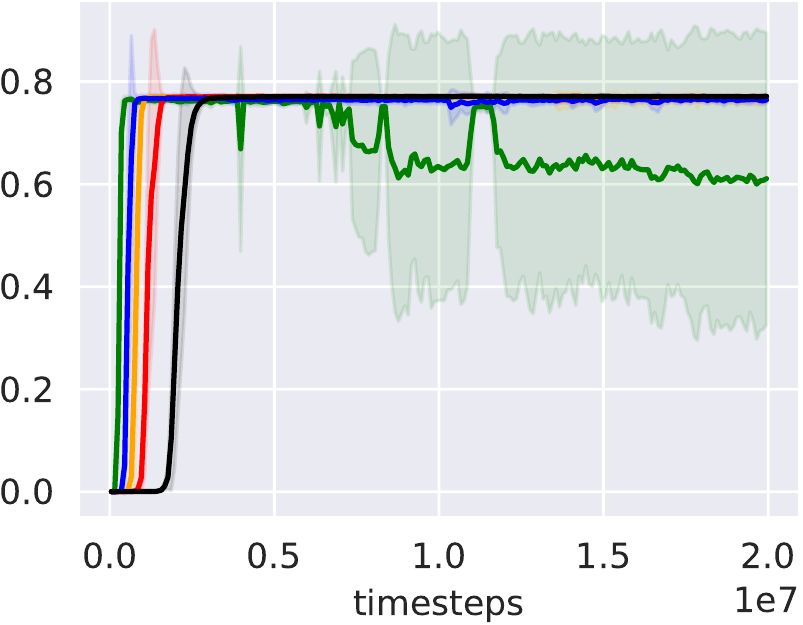}
    \caption{MultiRoom-N10-S4}
  \end{subfigure}%
  \begin{subfigure}[b]{0.25\textwidth}
    \centering
    \includegraphics[width=0.99\textwidth]{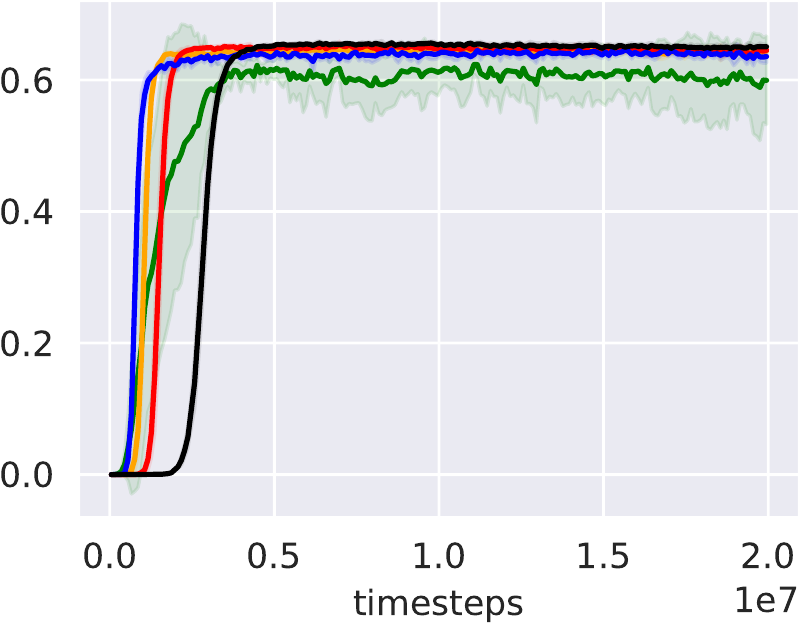}
    \caption{MultiRoom-N7-S8}
  \end{subfigure}%
  \begin{subfigure}[b]{0.25\textwidth}
    \centering
    \includegraphics[width=0.99\textwidth]{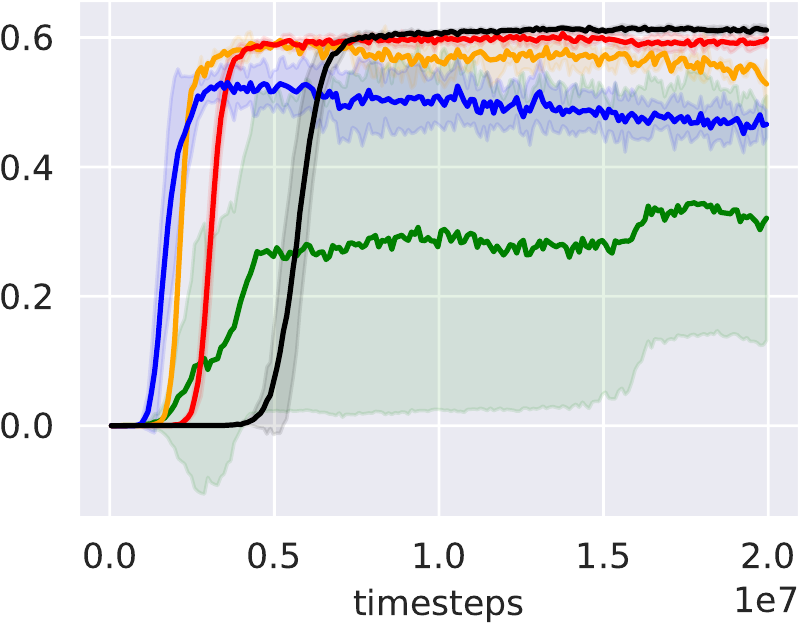}
    \caption{MultiRoom-N10-S10}
  \end{subfigure}%
  
  \begin{subfigure}[b]{0.25\textwidth}
    \centering
    \includegraphics[width=0.99\textwidth]{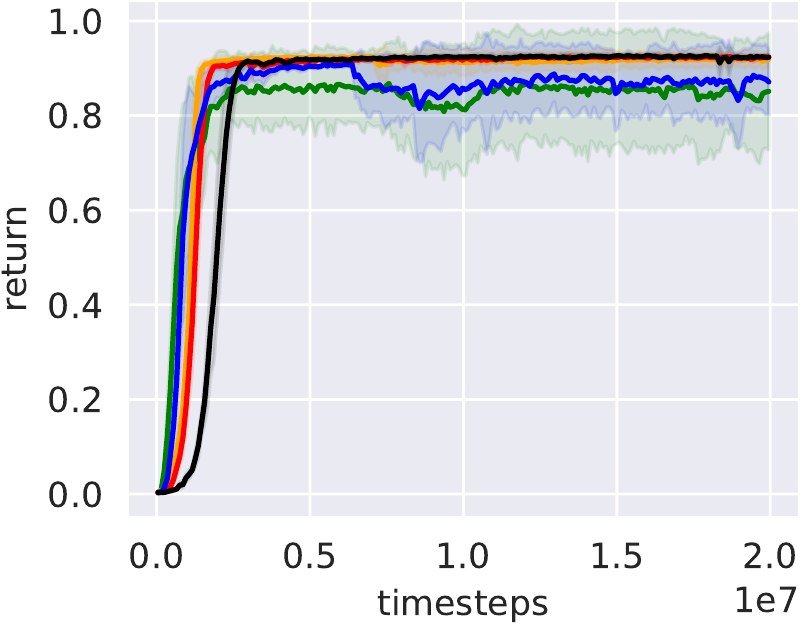}
    \caption{KeyCorridor-S3-R2}
  \end{subfigure}%
  \begin{subfigure}[b]{0.25\textwidth}
    \centering
    \includegraphics[width=0.99\textwidth]{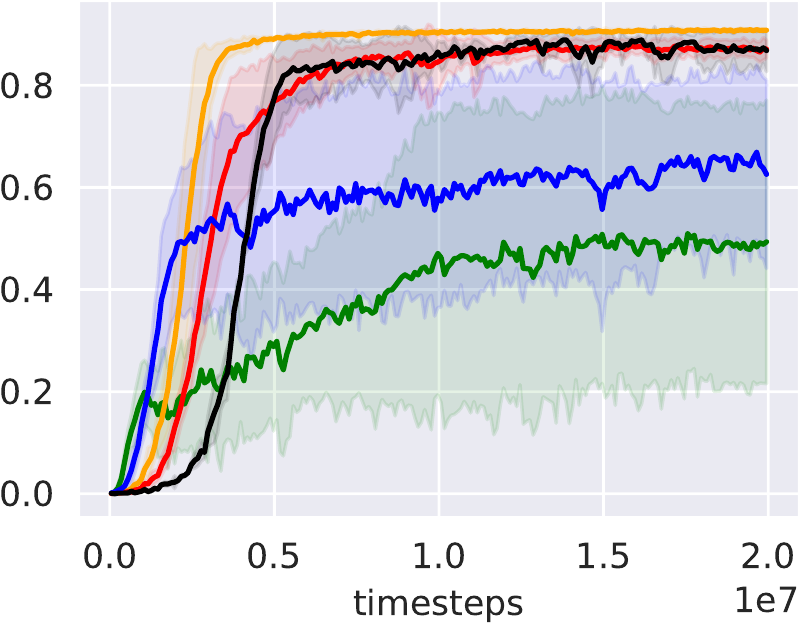}
    \caption{KeyCorridor-S3-R3}
  \end{subfigure}%
  \begin{subfigure}[b]{0.25\textwidth}
    \centering
    \includegraphics[width=0.99\textwidth]{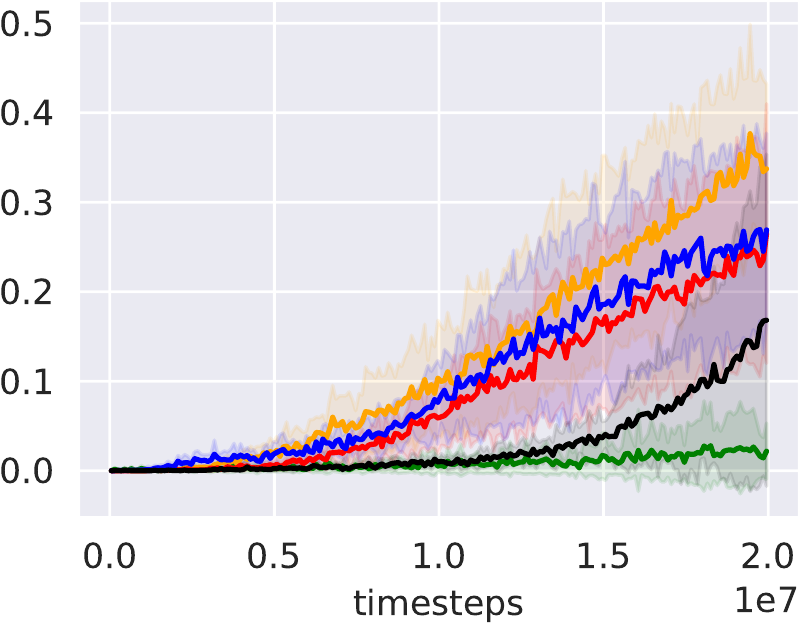}
    \caption{KeyCorridor-S4-R3}
  \end{subfigure}%
  \begin{subfigure}[b]{0.25\textwidth}
    \centering
    \includegraphics[width=0.99\textwidth]{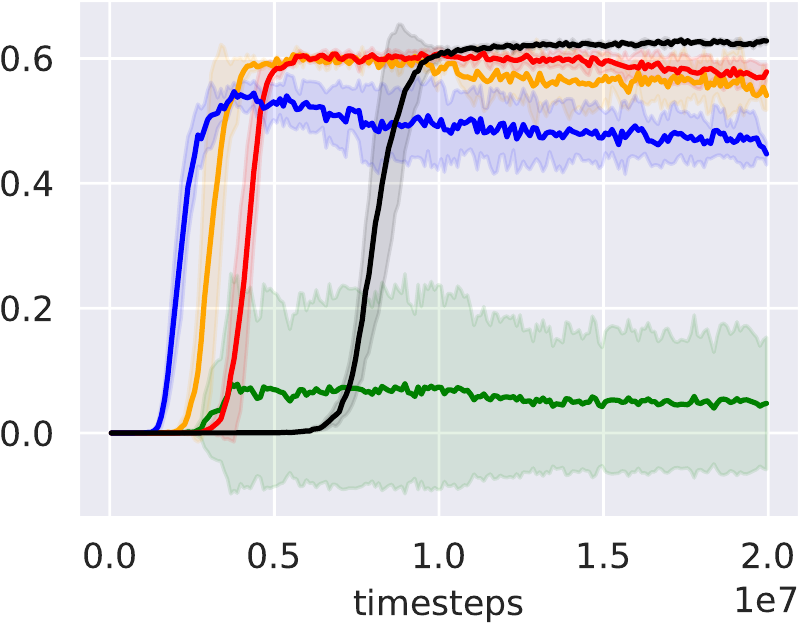}
    \caption{MultiRoom-N12-S10}
  \end{subfigure}%
  \caption{Impact of buffer size $D$ on MiniGrid environments.}
  \label{fig:pureexploration}
\end{figure}

\newpage
\subsection{Pure Exploration on MiniGrid}

\begin{figure}[h!]
  \centering
  \begin{subfigure}[b]{0.85\textwidth}
    \includegraphics[width=1.0\textwidth]{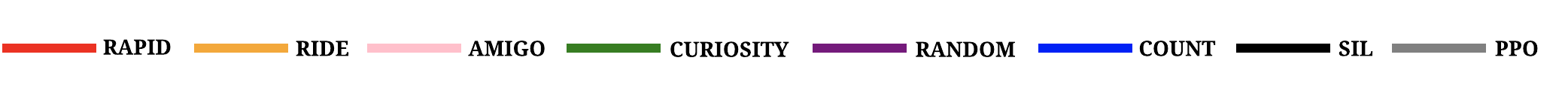}
  \end{subfigure}

  \begin{subfigure}[b]{0.25\textwidth}
    \centering
    \includegraphics[width=0.99\textwidth]{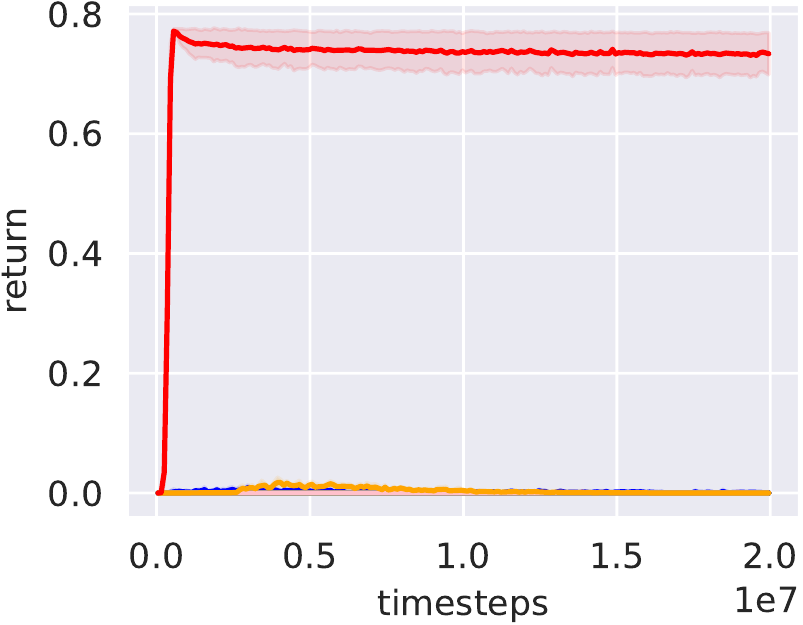}
    \caption{MultiRoom-N7-S4}
  \end{subfigure}%
  \begin{subfigure}[b]{0.25\textwidth}
    \centering
    \includegraphics[width=0.99\textwidth]{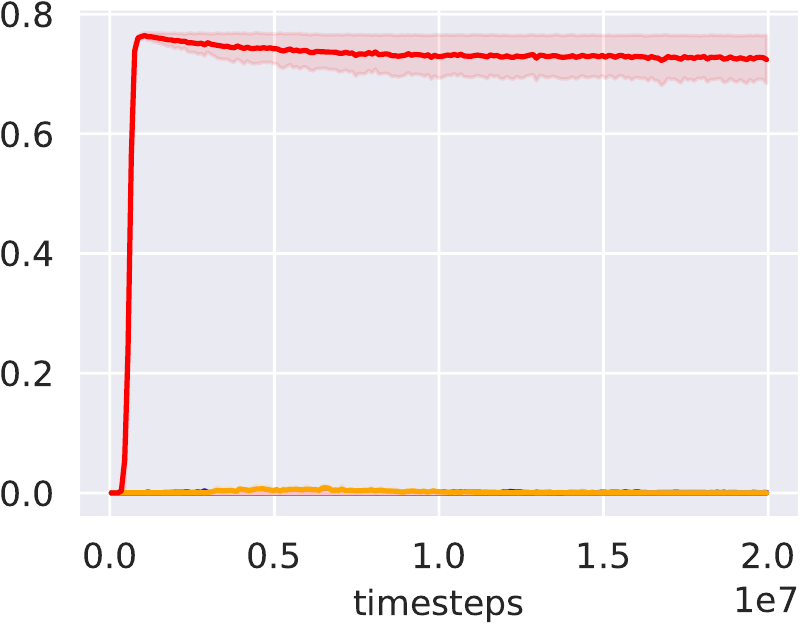}
    \caption{MultiRoom-N10-S4}
  \end{subfigure}%
  \begin{subfigure}[b]{0.25\textwidth}
    \centering
    \includegraphics[width=0.99\textwidth]{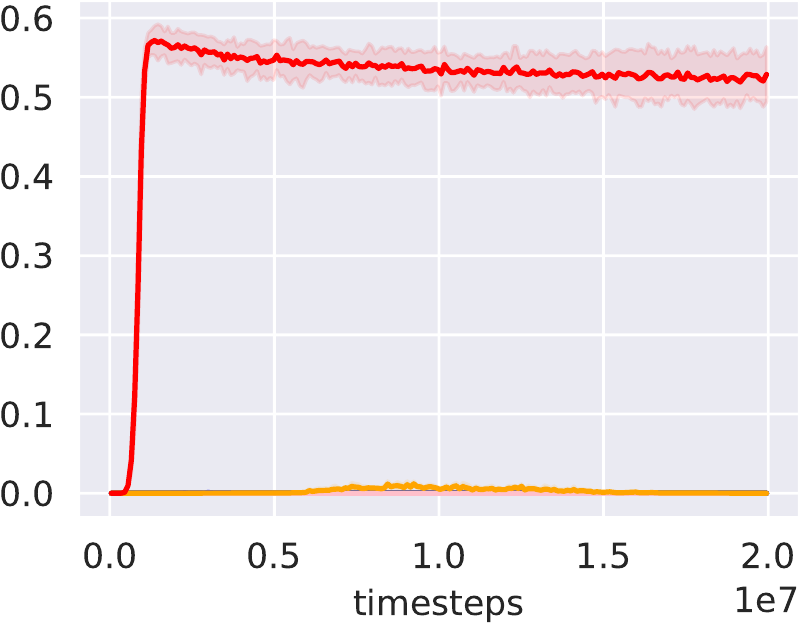}
    \caption{MultiRoom-N7-S8}
  \end{subfigure}%
  \begin{subfigure}[b]{0.25\textwidth}
    \centering
    \includegraphics[width=0.99\textwidth]{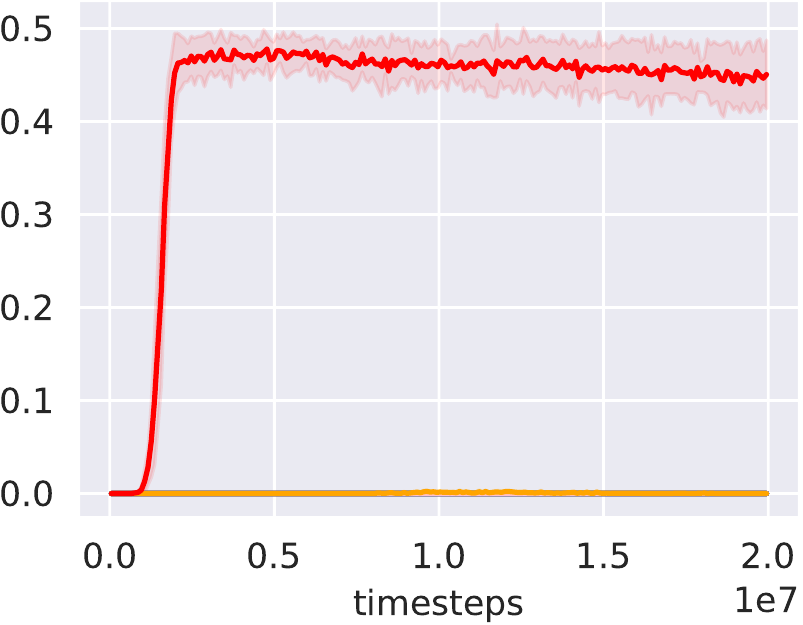}
    \caption{MultiRoom-N10-S10}
  \end{subfigure}%
  
  \begin{subfigure}[b]{0.25\textwidth}
    \centering
    \includegraphics[width=0.99\textwidth]{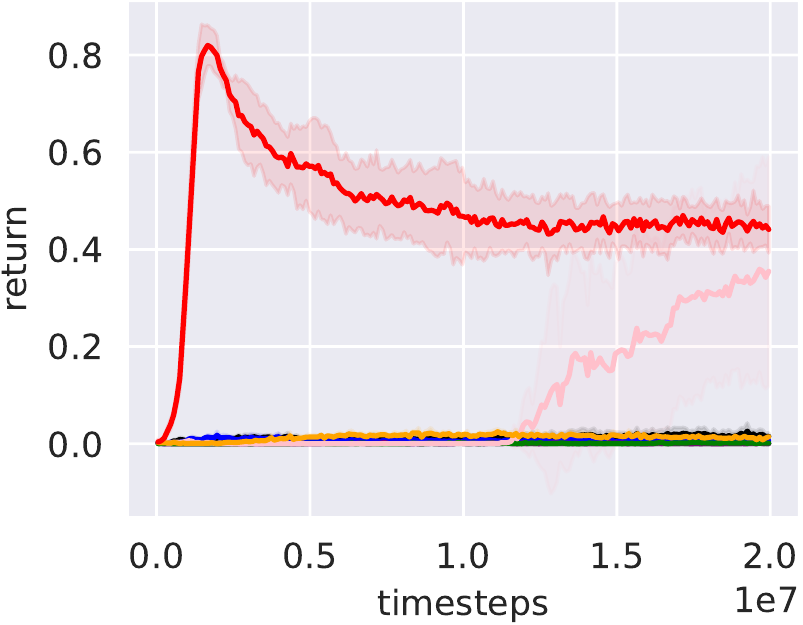}
    \caption{KeyCorridor-S3-R2}
  \end{subfigure}%
  \begin{subfigure}[b]{0.25\textwidth}
    \centering
    \includegraphics[width=0.99\textwidth]{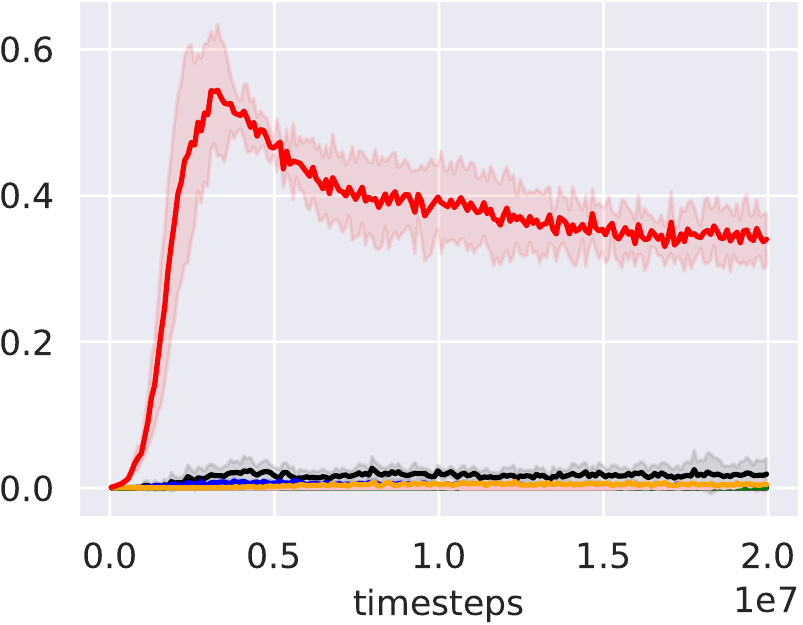}
    \caption{KeyCorridor-S3-R3}
  \end{subfigure}%
  \begin{subfigure}[b]{0.25\textwidth}
    \centering
    \includegraphics[width=0.99\textwidth]{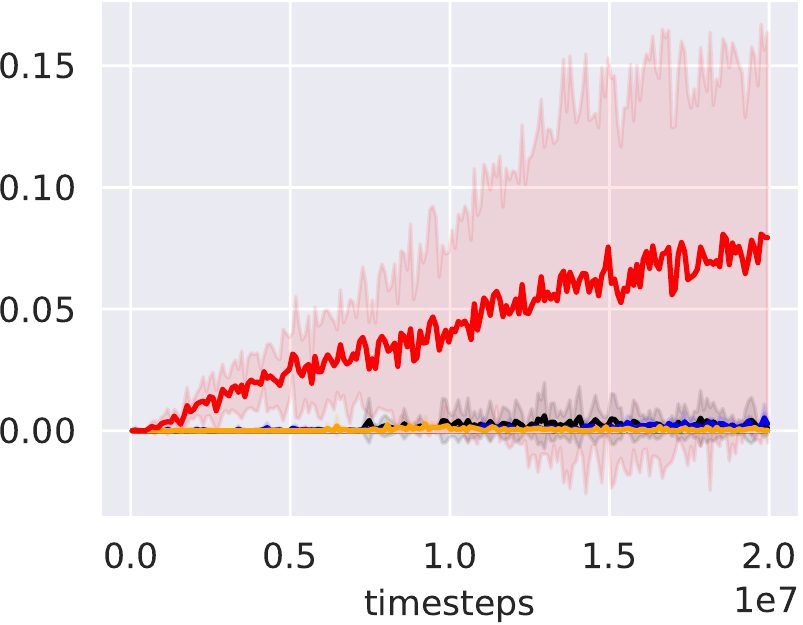}
    \caption{KeyCorridor-S4-R3}
  \end{subfigure}%
  \begin{subfigure}[b]{0.25\textwidth}
    \centering
    \includegraphics[width=0.99\textwidth]{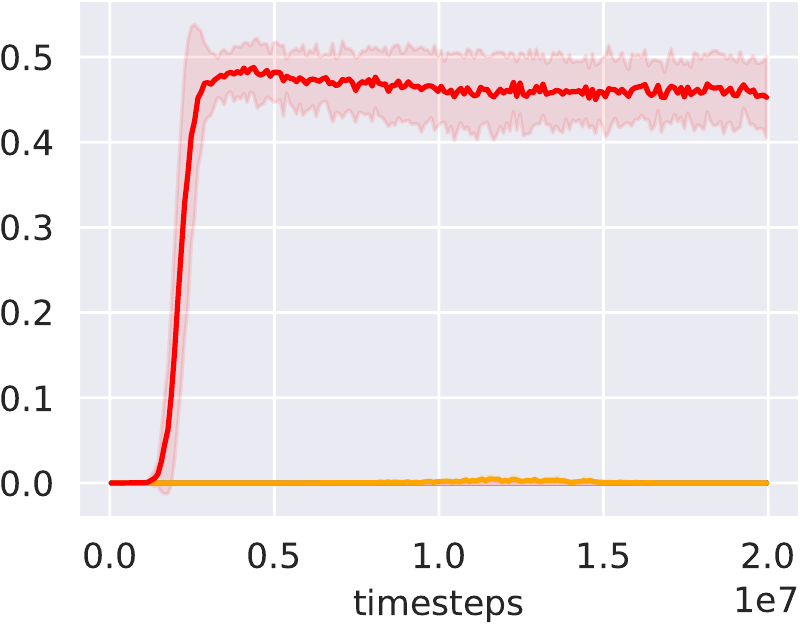}
    \caption{MultiRoom-N12-S10}
  \end{subfigure}%
  \caption{Extrinsic rewards achieved by RAPID and baselines with pure exploration}
\end{figure}

\subsection{Local Exploration Score of Pure Exploration on MiniGrid}

\begin{figure}[h!]
  \centering
  \begin{subfigure}[b]{0.85\textwidth}
    \includegraphics[width=1.0\textwidth]{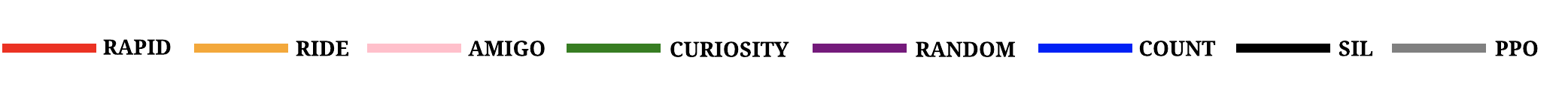}
  \end{subfigure}

  \begin{subfigure}[b]{0.25\textwidth}
    \centering
    \includegraphics[width=0.99\textwidth]{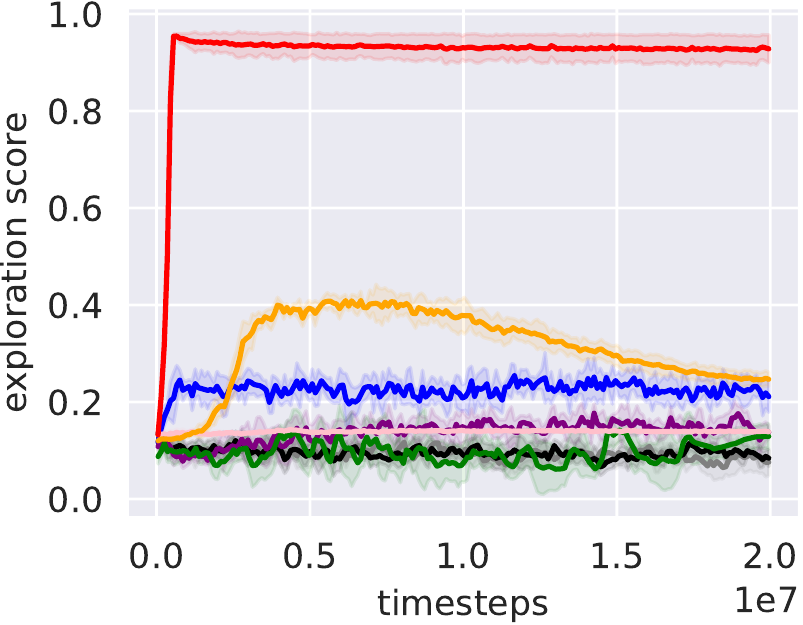}
    \caption{MultiRoom-N7-S4}
  \end{subfigure}%
  \begin{subfigure}[b]{0.25\textwidth}
    \centering
    \includegraphics[width=0.99\textwidth]{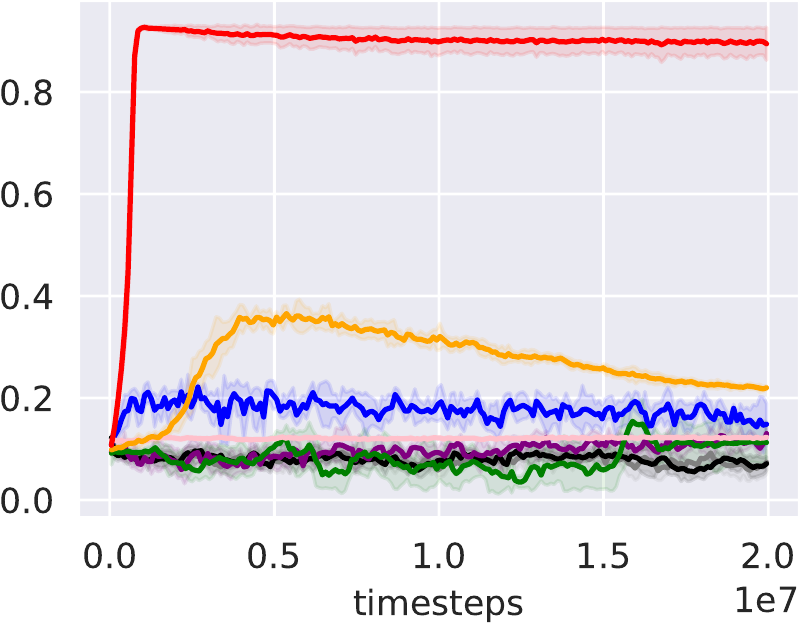}
    \caption{MultiRoom-N10-S4}
  \end{subfigure}%
  \begin{subfigure}[b]{0.25\textwidth}
    \centering
    \includegraphics[width=0.99\textwidth]{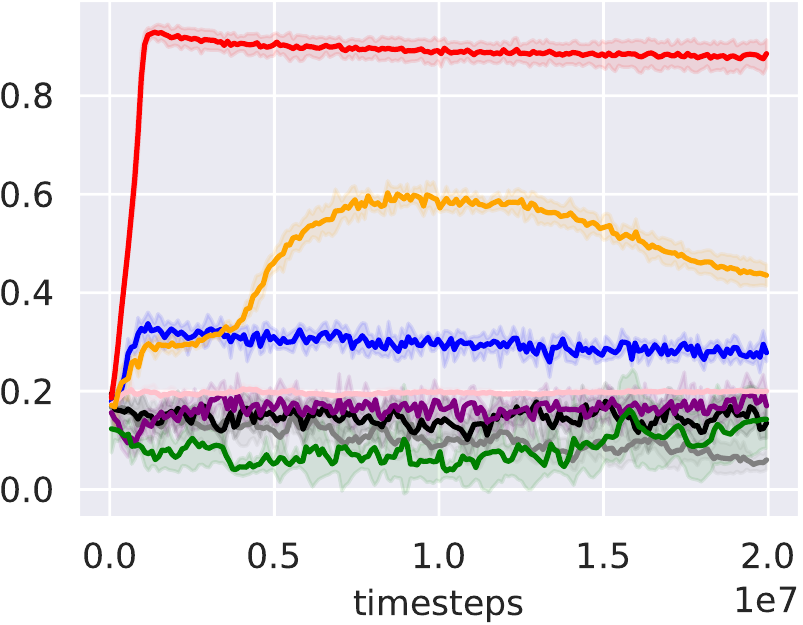}
    \caption{MultiRoom-N7-S8}
  \end{subfigure}%
  \begin{subfigure}[b]{0.25\textwidth}
    \centering
    \includegraphics[width=0.99\textwidth]{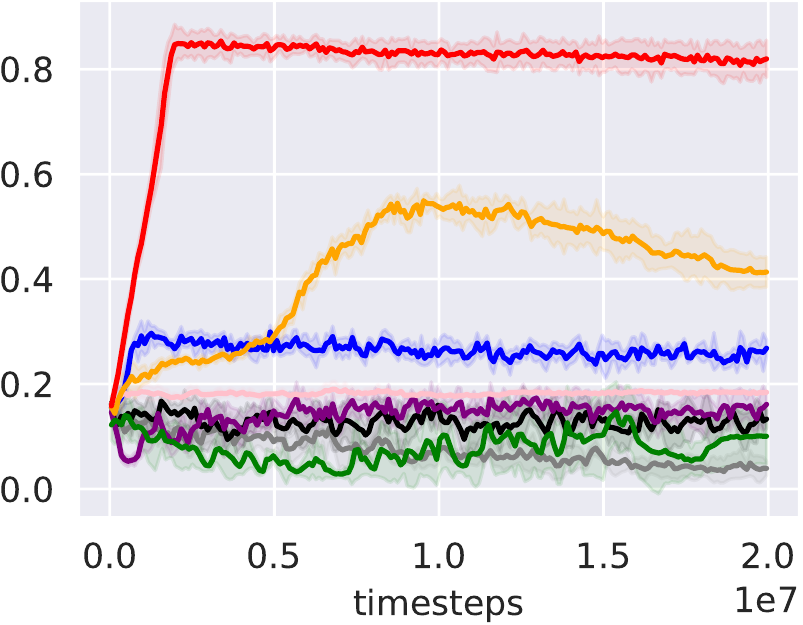}
    \caption{MultiRoom-N10-S10}
  \end{subfigure}%
  
  \begin{subfigure}[b]{0.25\textwidth}
    \centering
    \includegraphics[width=0.99\textwidth]{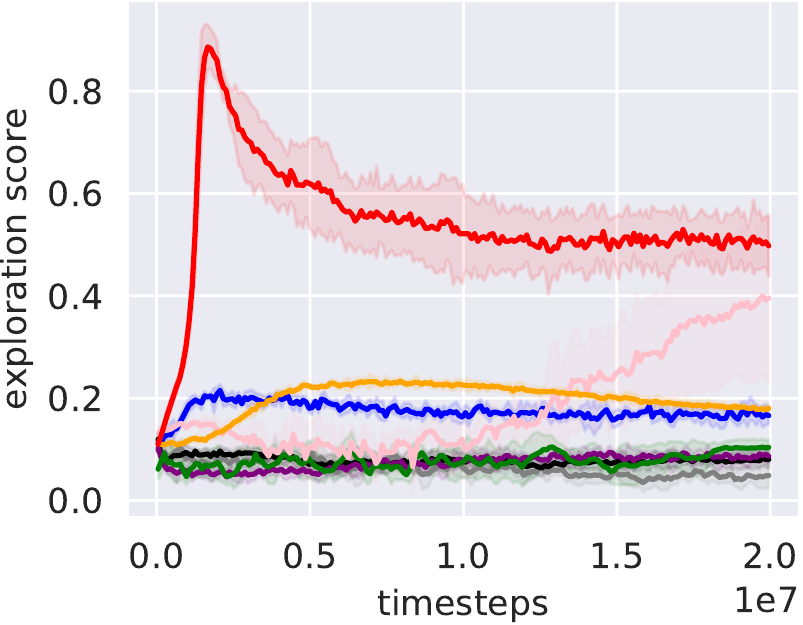}
    \caption{KeyCorridor-S3-R2}
  \end{subfigure}%
  \begin{subfigure}[b]{0.25\textwidth}
    \centering
    \includegraphics[width=0.99\textwidth]{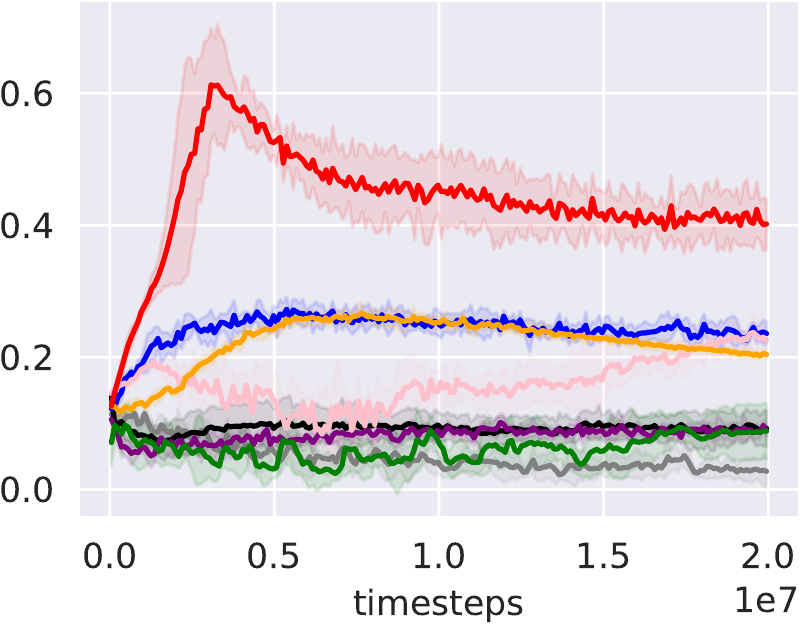}
    \caption{KeyCorridor-S3-R3}
  \end{subfigure}%
  \begin{subfigure}[b]{0.25\textwidth}
    \centering
    \includegraphics[width=0.99\textwidth]{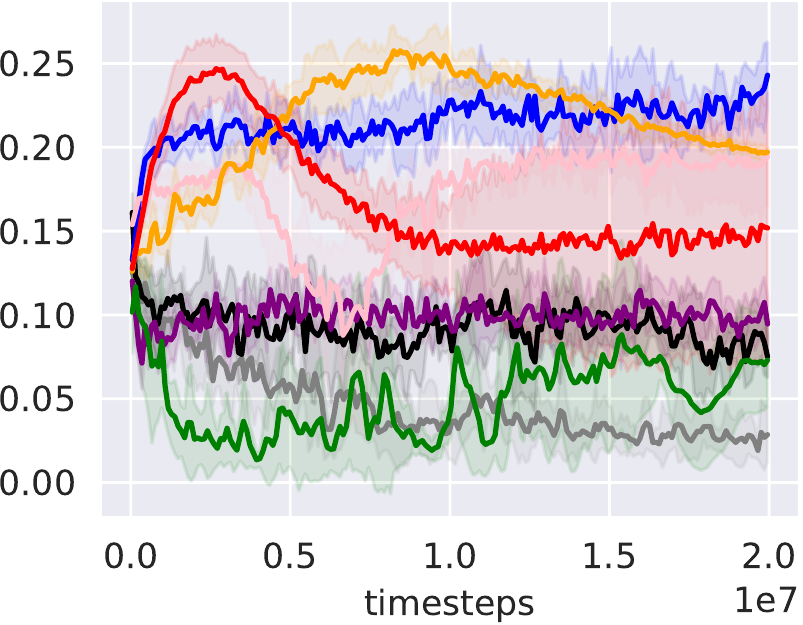}
    \caption{KeyCorridor-S4-R3}
  \end{subfigure}%
  \begin{subfigure}[b]{0.25\textwidth}
    \centering
    \includegraphics[width=0.99\textwidth]{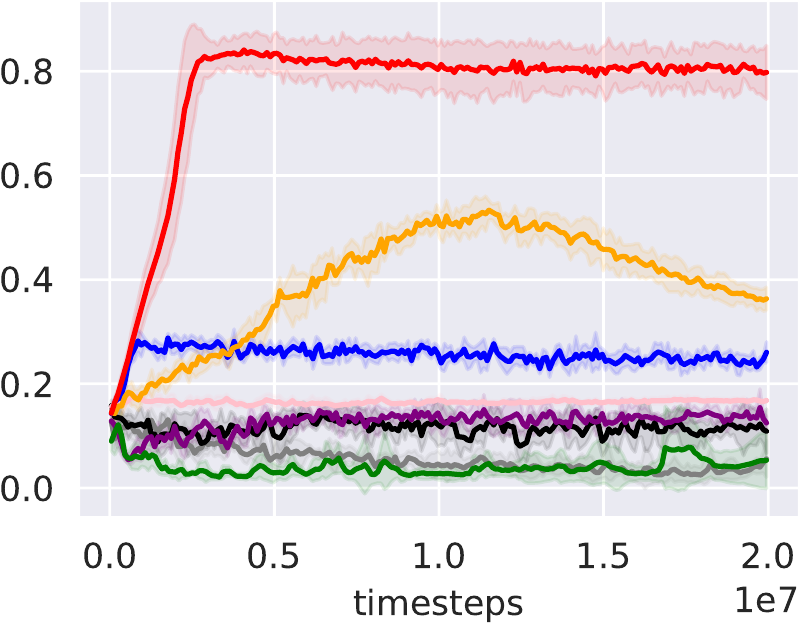}
    \caption{MultiRoom-N12-S10}
  \end{subfigure}%
  \caption{Local exploration scores achieved by RAPID and baselines with pure exploration}
\end{figure}

\newpage
\section{Learning Curves with More Rooms}
\label{sec:E}

\begin{figure}[h!]
  \centering
  \begin{subfigure}[b]{0.2\textwidth}
    \includegraphics[width=1.0\textwidth]{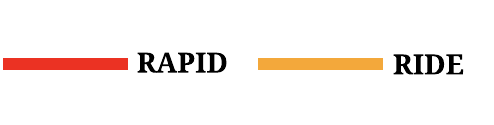}
  \end{subfigure}

  \begin{subfigure}[b]{0.25\textwidth}
    \centering
    \includegraphics[width=0.99\textwidth]{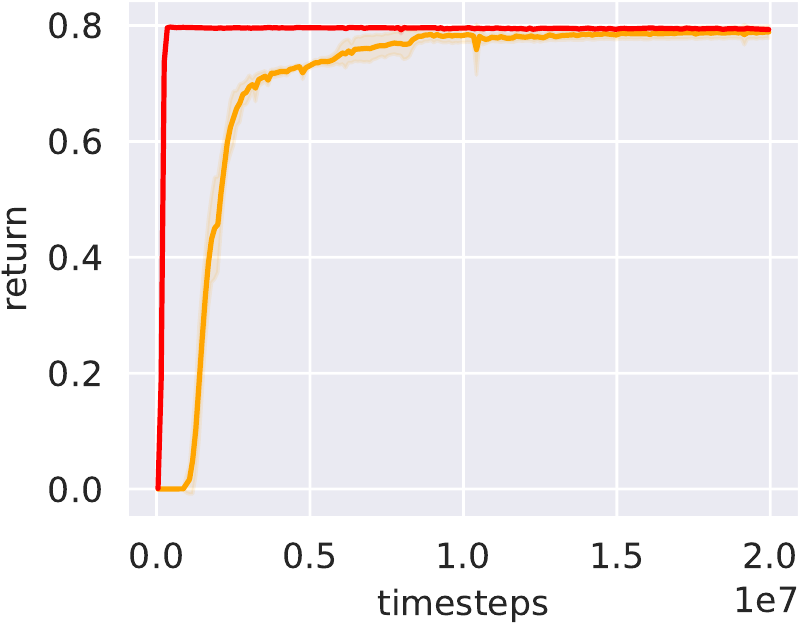}
    \caption{MultiRoom-N4-S4}
  \end{subfigure}%
  \begin{subfigure}[b]{0.25\textwidth}
    \centering
    \includegraphics[width=0.99\textwidth]{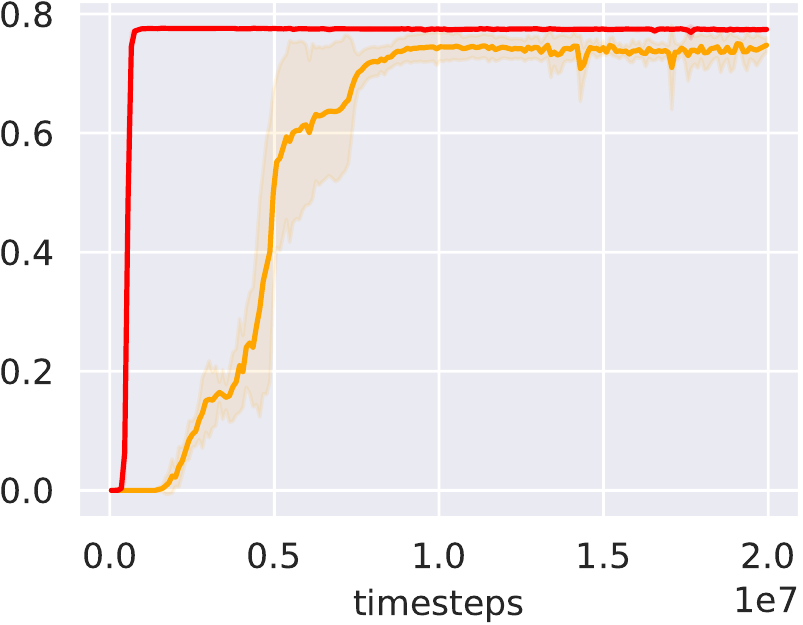}
    \caption{MultiRoom-N8-S4}
  \end{subfigure}%
  \begin{subfigure}[b]{0.25\textwidth}
    \centering
    \includegraphics[width=0.99\textwidth]{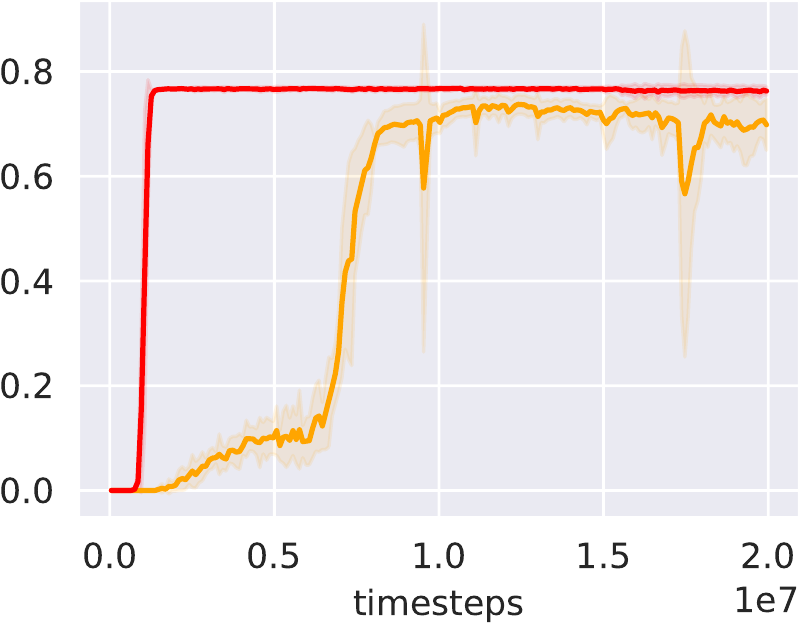}
    \caption{MultiRoom-N12-S4}
  \end{subfigure}%
  \begin{subfigure}[b]{0.25\textwidth}
    \centering
    \includegraphics[width=0.99\textwidth]{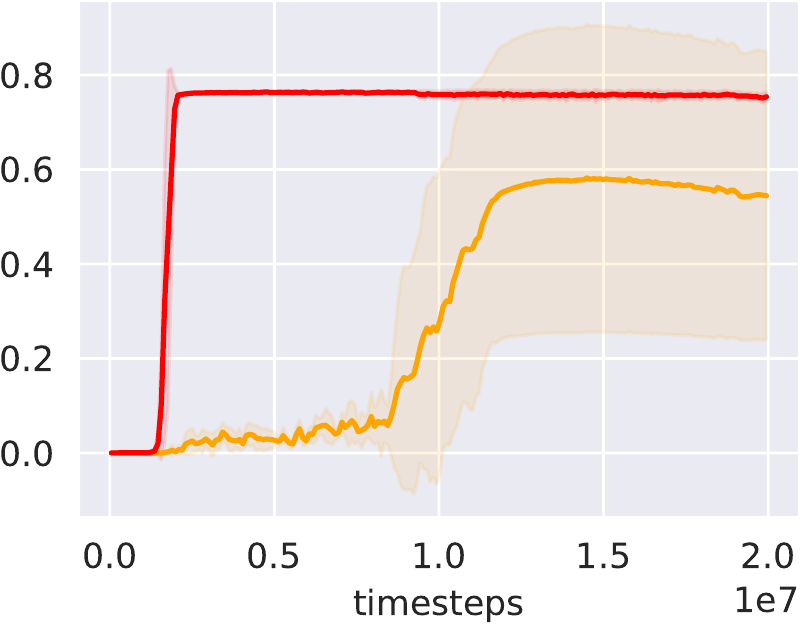}
    \caption{MultiRoom-N16-S4}
  \end{subfigure}%
  
  \begin{subfigure}[b]{0.25\textwidth}
    \centering
    \includegraphics[width=0.99\textwidth]{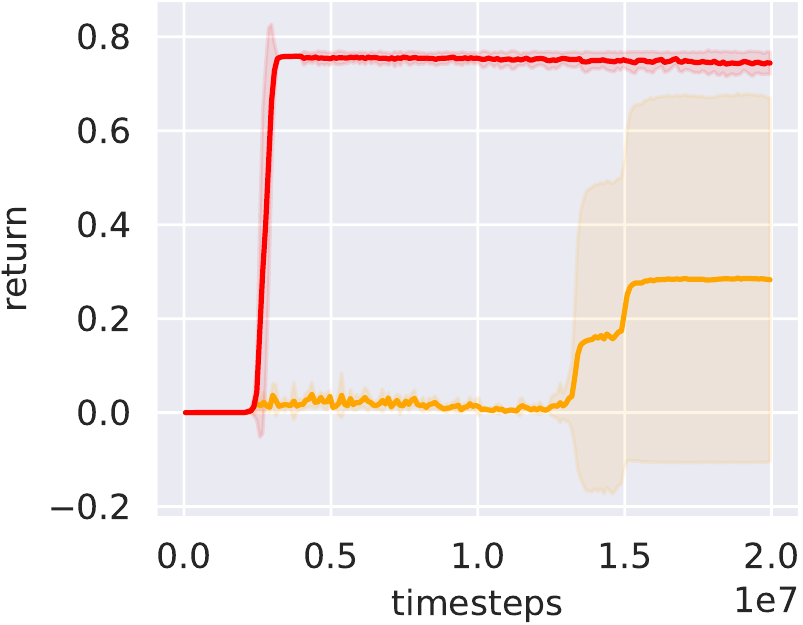}
    \caption{MultiRoom-N20-S4}
  \end{subfigure}%
  \begin{subfigure}[b]{0.25\textwidth}
    \centering
    \includegraphics[width=0.99\textwidth]{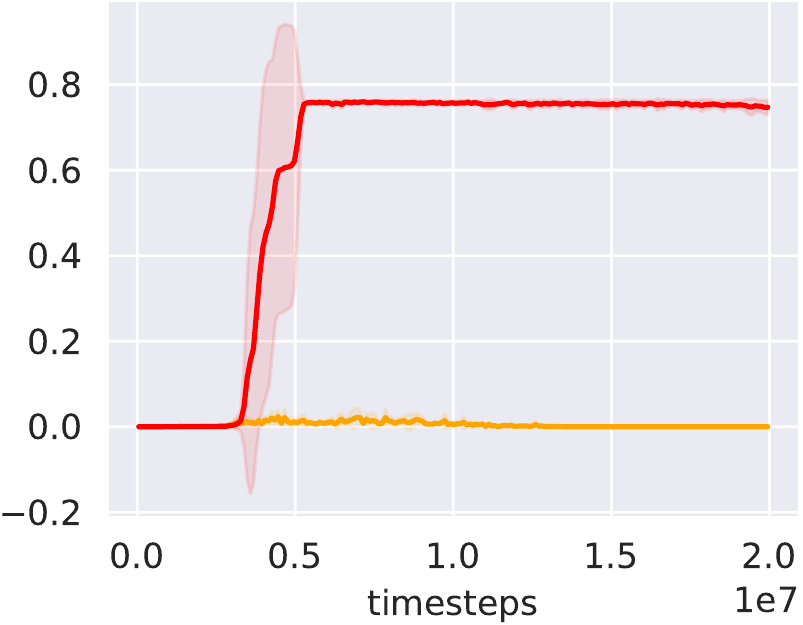}
    \caption{MultiRoom-N24-S4}
  \end{subfigure}%
  \begin{subfigure}[b]{0.25\textwidth}
    \centering
    \includegraphics[width=0.99\textwidth]{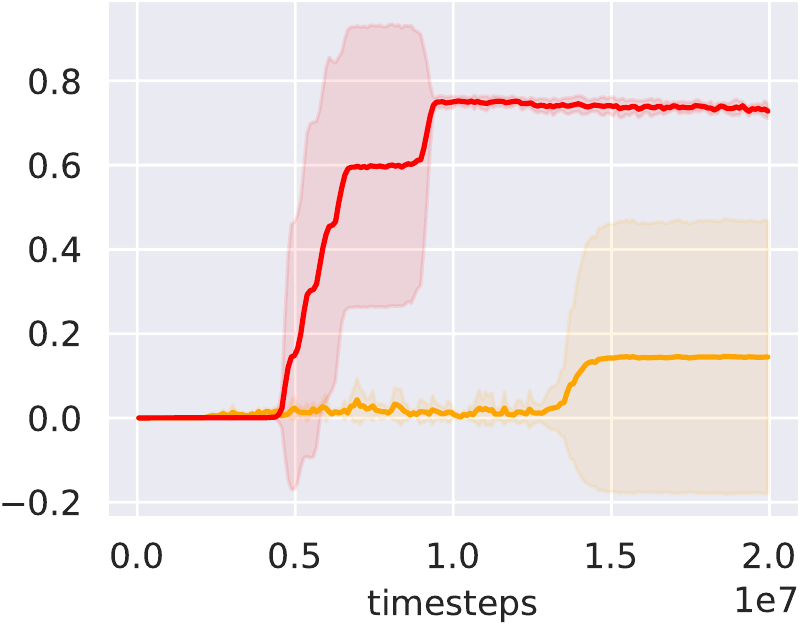}
    \caption{MultiRoom-N28-S4}
  \end{subfigure}%
  \caption{The learning curves with more rooms}
  \label{fig:morerooms}
\end{figure}

\section{Learning Curves with Large Rooms}
\label{sec:F}

\begin{figure}[h!]
  \centering
  \begin{subfigure}[b]{0.2\textwidth}
    \includegraphics[width=1.0\textwidth]{imgs/more_rooms/legend.pdf}
  \end{subfigure}

  \begin{subfigure}[b]{0.25\textwidth}
    \centering
    \includegraphics[width=0.99\textwidth]{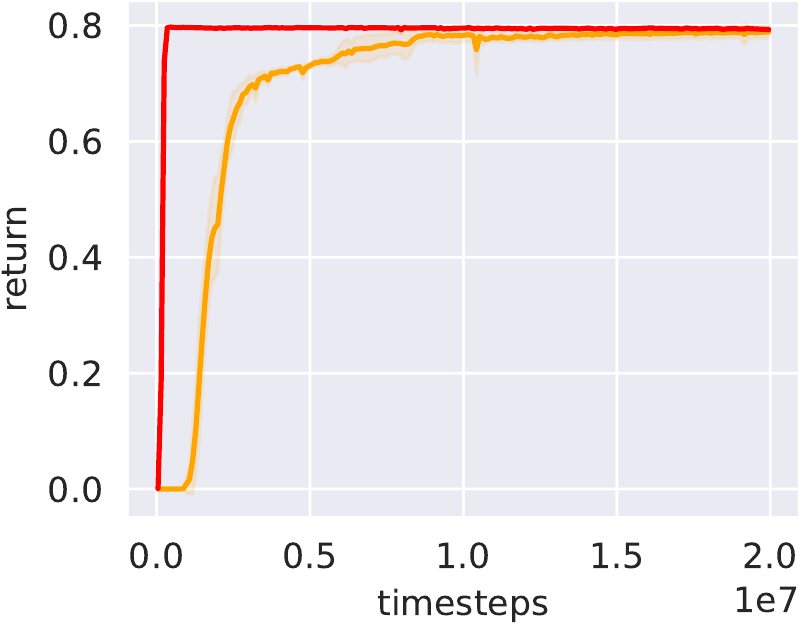}
    \caption{MultiRoom-N4-S4}
  \end{subfigure}%
  \begin{subfigure}[b]{0.25\textwidth}
    \centering
    \includegraphics[width=0.99\textwidth]{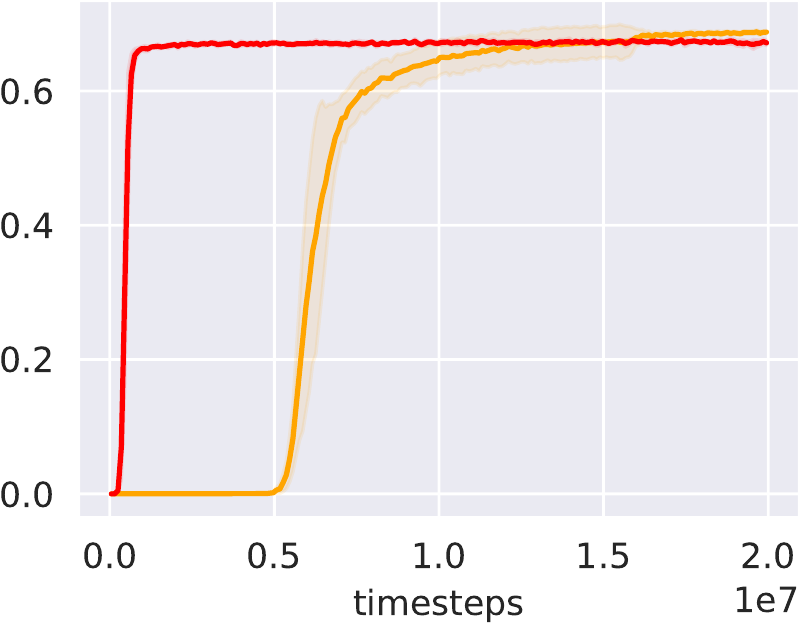}
    \caption{MultiRoom-N4-S8}
  \end{subfigure}%
  \begin{subfigure}[b]{0.25\textwidth}
    \centering
    \includegraphics[width=0.99\textwidth]{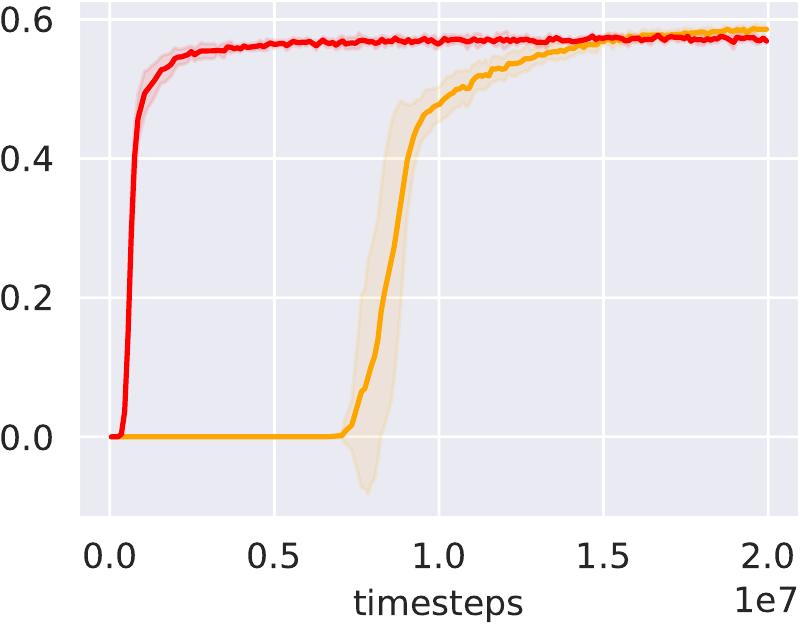}
    \caption{MultiRoom-N4-S12}
  \end{subfigure}%

  \begin{subfigure}[b]{0.25\textwidth}
    \centering
    \includegraphics[width=0.99\textwidth]{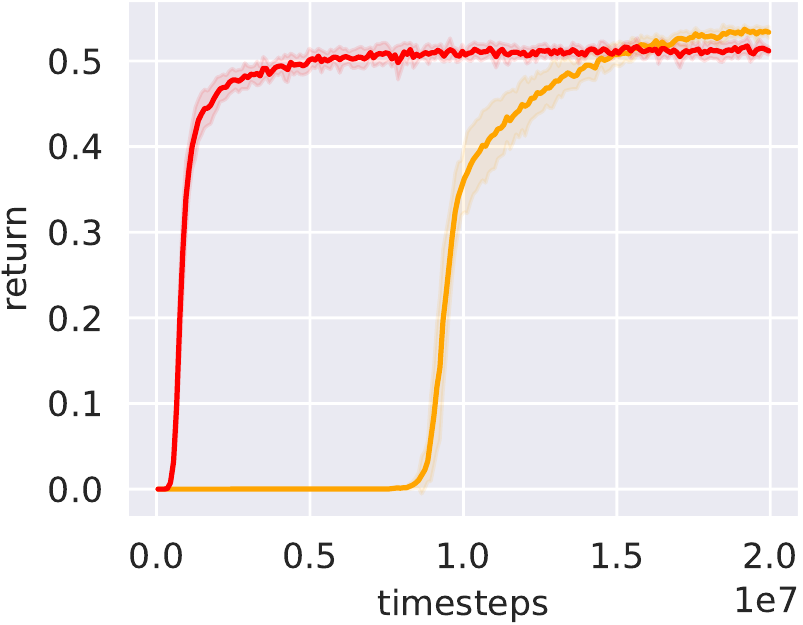}
    \caption{MultiRoom-N4-S16}
  \end{subfigure}%
  \begin{subfigure}[b]{0.25\textwidth}
    \centering
    \includegraphics[width=0.99\textwidth]{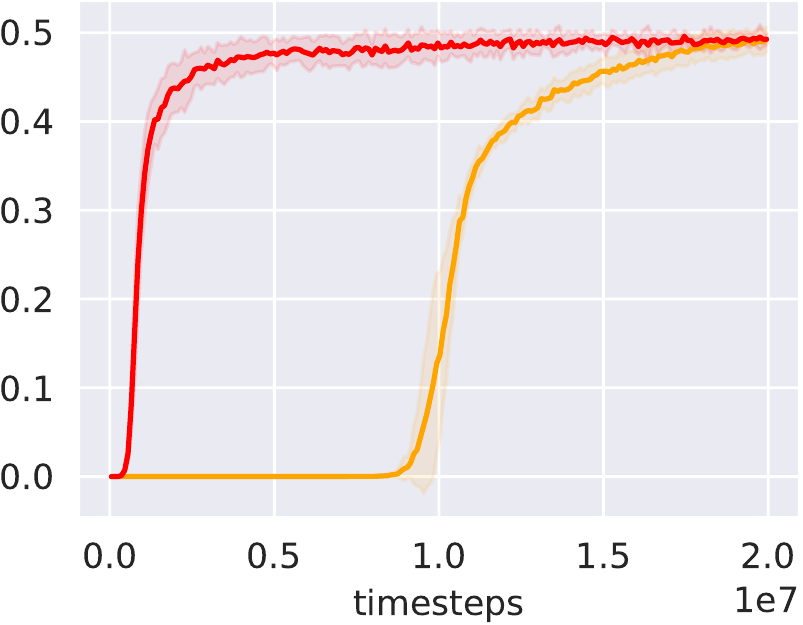}
    \caption{MultiRoom-N4-S20}
  \end{subfigure}%
  \begin{subfigure}[b]{0.25\textwidth}
    \centering
    \includegraphics[width=0.99\textwidth]{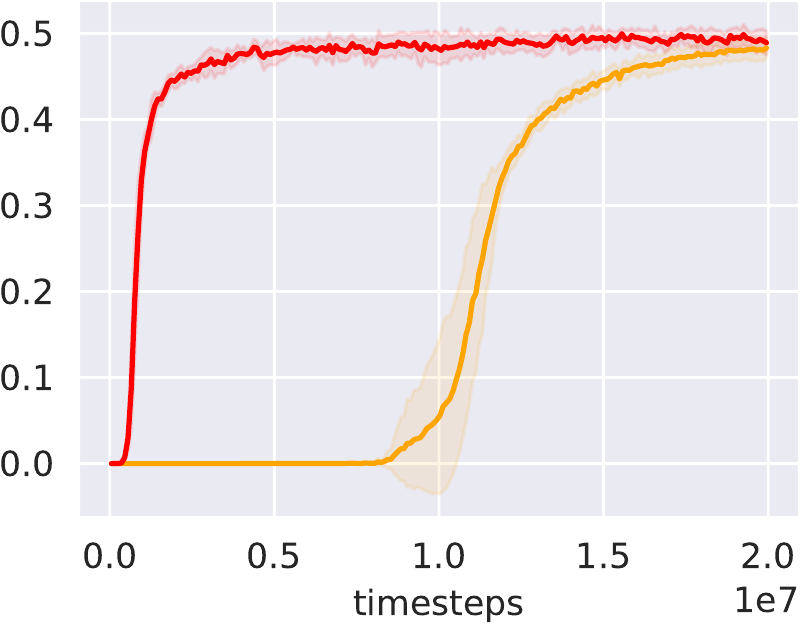}
    \caption{MultiRoom-N4-S24}
  \end{subfigure}%

  \begin{subfigure}[b]{0.25\textwidth}
    \centering
    \includegraphics[width=0.99\textwidth]{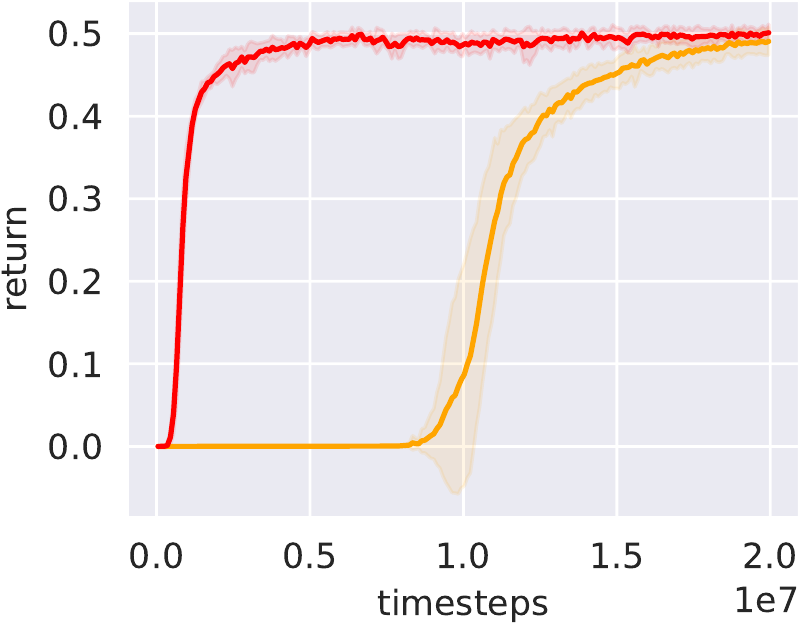}
    \caption{MultiRoom-N4-S28}
  \end{subfigure}%
  \begin{subfigure}[b]{0.25\textwidth}
    \centering
    \includegraphics[width=0.99\textwidth]{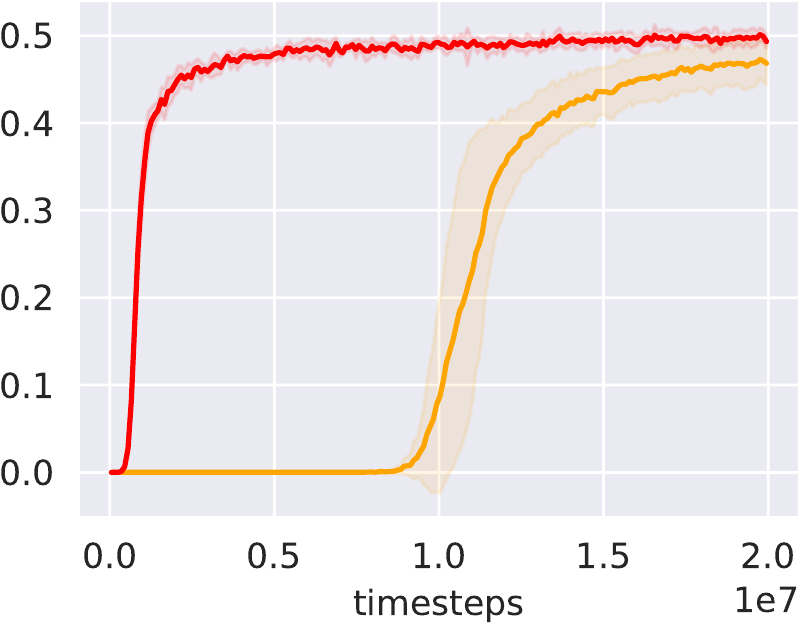}
    \caption{MultiRoom-N4-S32}
  \end{subfigure}%
  \begin{subfigure}[b]{0.25\textwidth}
    \centering
    \includegraphics[width=0.99\textwidth]{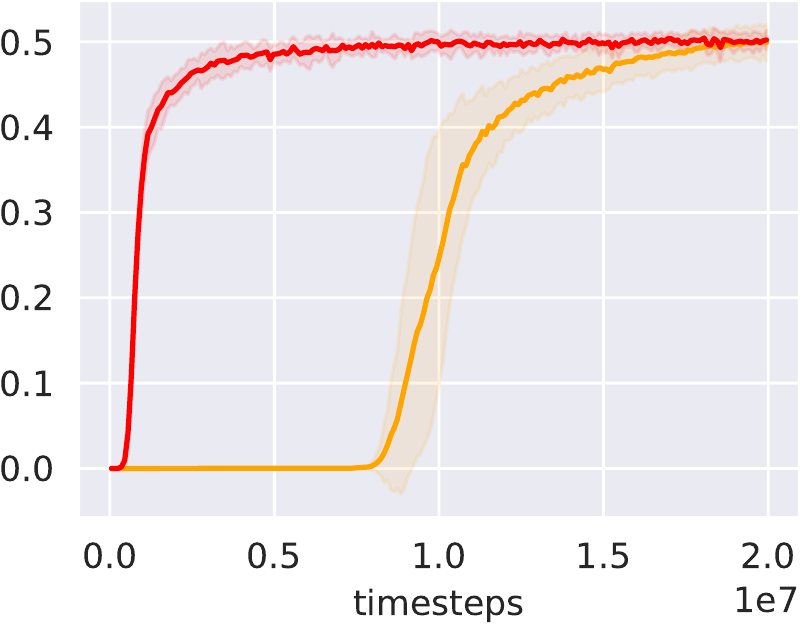}
    \caption{MultiRoom-N4-S36}
  \end{subfigure}%
  \caption{The learning curves with larger rooms sizes.}
\end{figure}

\newpage
\section{Ablation and Analysis of MiniWorld-Maze}
\label{sec:G}

\begin{figure}[h!]
  \centering
    \includegraphics[width=0.5\textwidth]{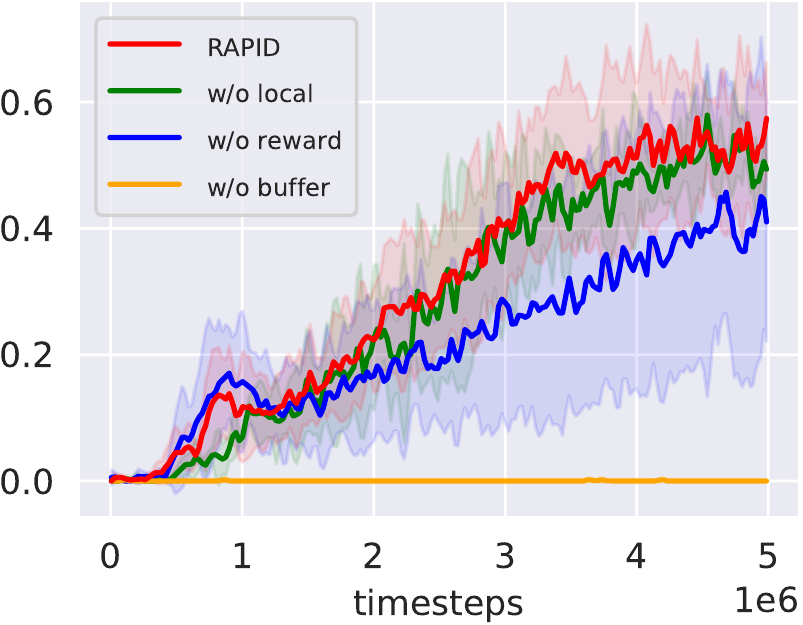}
  \label{fig:miniworldablation}
  \caption{Learning curves of RAPID and the ablations on MiniWorld Maze. We observe minor performance drop when removing the local score, substantial performance drop when removing extrinsic rewards or the buffer.}
\end{figure}

\begin{figure}[h!]
  \centering
    \includegraphics[width=0.5\textwidth]{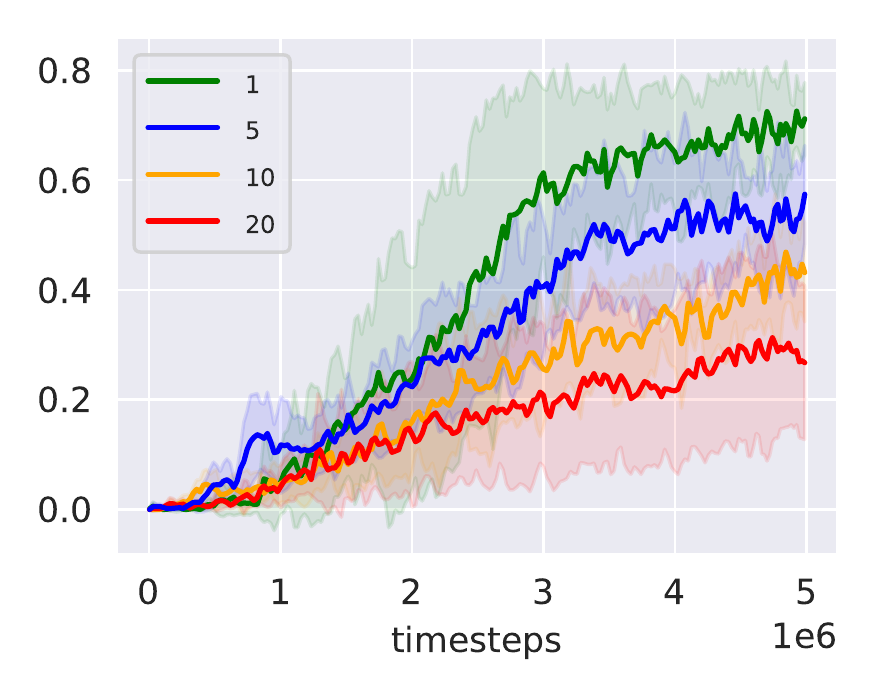}
  \label{fig:miniworldimpactupdate}
  \caption{Impact of training steps on MiniWorld Maze.}
\end{figure}

\begin{figure}[h!]
  \centering
    \includegraphics[width=0.5\textwidth]{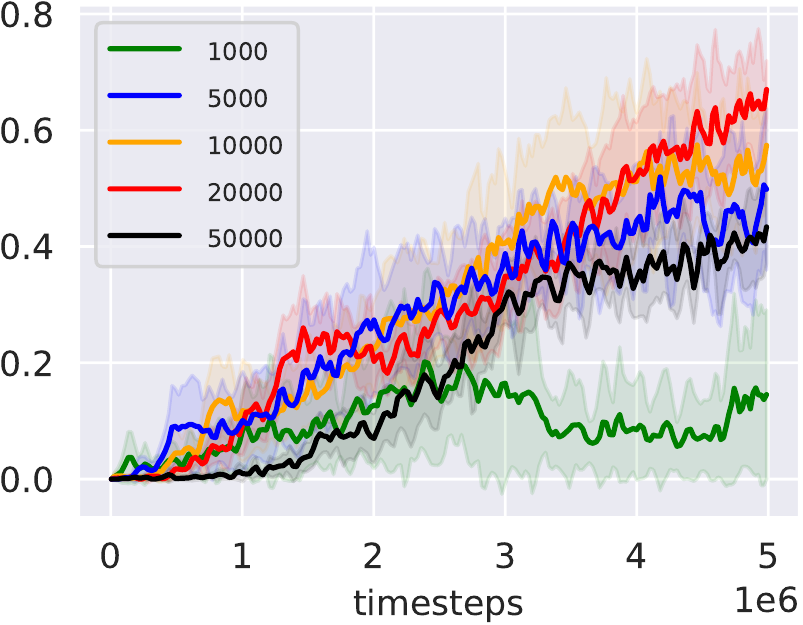}
  \caption{Impact of buffer size on MiniWorld Maze.}
\end{figure}

\begin{figure}[h!]
  \centering
    \includegraphics[width=0.5\textwidth]{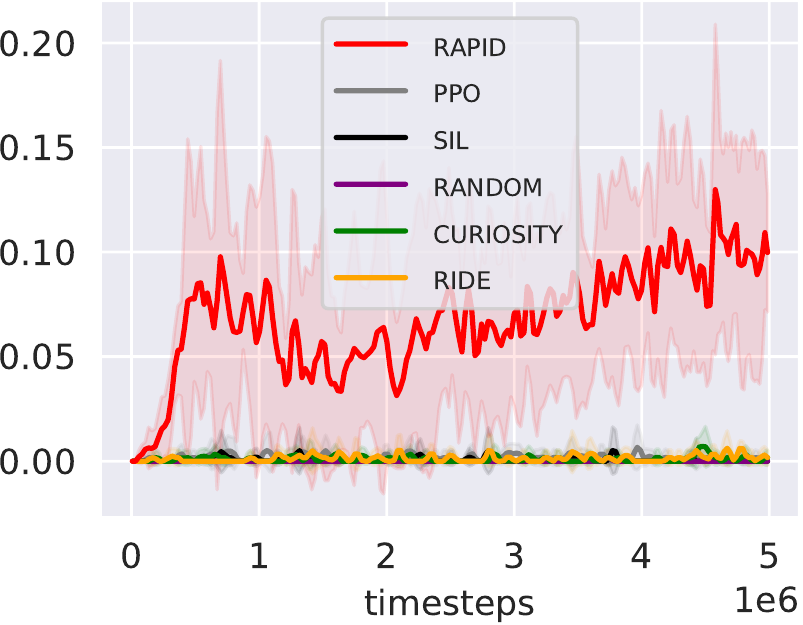}
  \caption{Extrinsic rewards achieved by RAPID and baselines on MiniWorld Maze with pure exploration.}
\end{figure}

\begin{figure}[h!]
  \centering
    \includegraphics[width=0.5\textwidth]{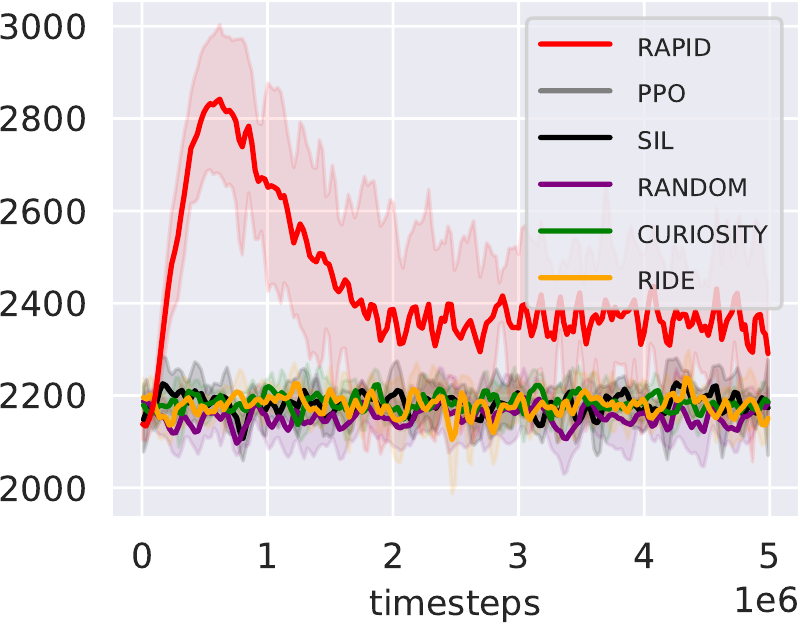}
  \caption{Local exploration scores achieved by RAPID and baselines on MiniWorld Maze with pure exploration.}
\end{figure}

\newpage
\section{Comparison with Storing Entire Episodes in the Buffer}
\label{sec:H}
\begin{figure}[h!]
  \centering
    \includegraphics[width=0.5\textwidth]{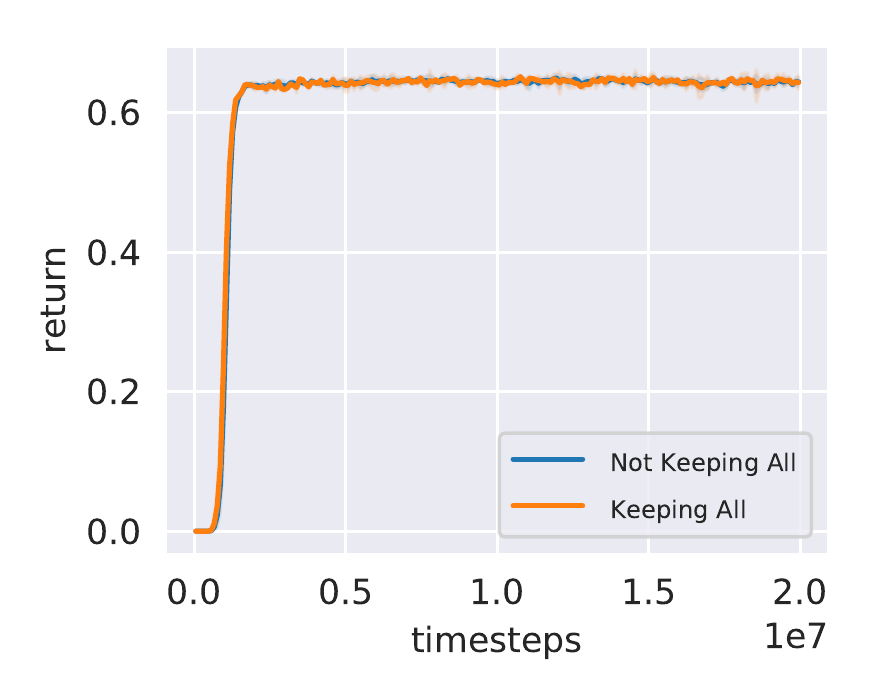}
  \caption{Comparison of keeping all the state-action pairs of an episode and not keeping all the state-action pairs (i.e., a fixed buffer size) on MultiRoom-N7-S8. The experiments are run 5 times with different random seeds. We observe no clear difference in the learning curves of these two implementations.}
  \label{fig:keepall}
\end{figure}

A variant of RAPID is to keep the entire episodes in the buffer instead of state-action pairs. The intuition of keeping the whole episode is that it is possible that a particular state-action pair may only be good in terms of exploration, in the context of the rest of the agent's trajectory in that episode. To test this variant, we force the buffer to keep the entire episode by allowing the episode at the end of the buffer to exceed the buffer size. For example, given that the buffer size is $5000$, if the length of the end episode is $160$ and the number of state-action pairs in the buffer is $5120$, we do not cut off the end episode and exceptionally allow the buffer to store $5120$ state-action pairs. In this way, we ensure that the entire episodes are kept in the buffer. We run this experiment on MiniGrid-MultiRoom-N7-S8-v0~(Figure~\ref{fig:keepall}). We do not observe a clear difference in the learning curves.

\newpage
\section{Results on Procedurally-Generated Swimmer}
\label{sec:I}
\begin{figure}[h!]
  \centering
  \begin{subfigure}[b]{0.3\textwidth}
    \includegraphics[width=\textwidth]{imgs/mujoco/legend.pdf}
  \end{subfigure}
  
  \begin{subfigure}[b]{0.33\textwidth}
    \centering
    \includegraphics[width=0.99\textwidth]{imgs/mujoco/SparseSwimmer-v2.pdf}
    \caption{Swimmer}
  \end{subfigure}%
  \begin{subfigure}[b]{0.33\textwidth}
    \centering
    \includegraphics[width=0.99\textwidth]{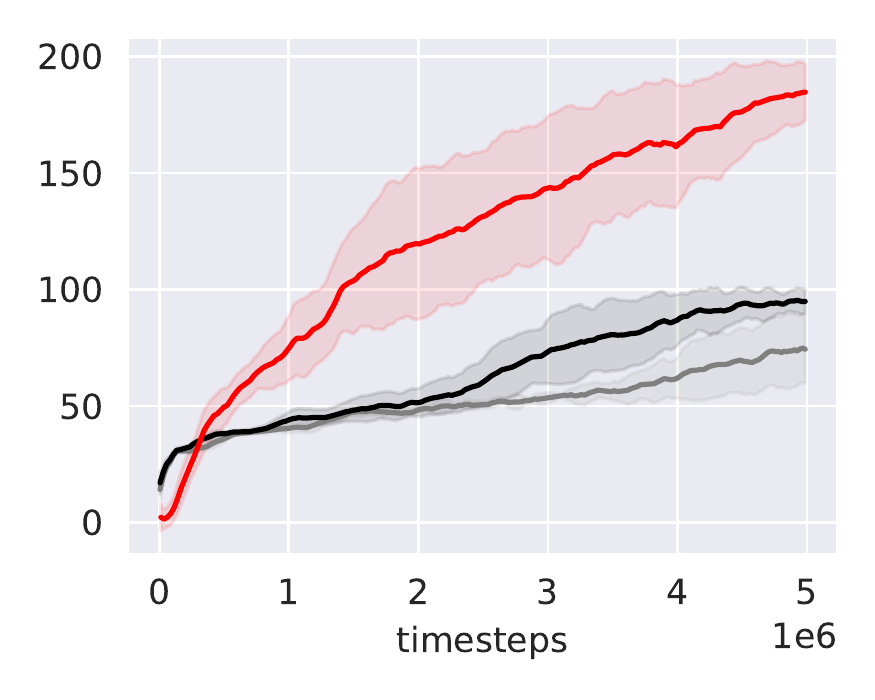}
    \caption{Density-Swimmer}
  \end{subfigure}%
  \begin{subfigure}[b]{0.33\textwidth}
    \centering
    \includegraphics[width=0.99\textwidth]{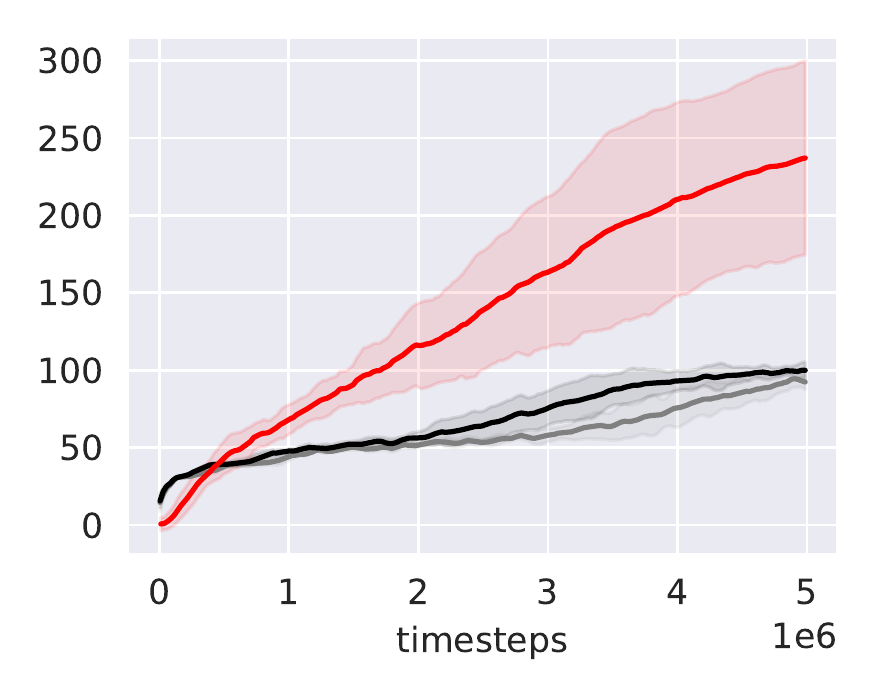}
    \caption{Velocity-Swimmer}
  \end{subfigure}%
  \caption{(a) is the singleton swimmer environment. In (b) the density is procedurally-generated. In (c) the velocity is procedurally-generated. All the methods tend to learn slower in the procedurally-generated settings. Nevertheless, RAPID is still able to discover good policies in these challenging variants.}
  \label{fig:swimmer}
\end{figure}

The standard MuJoCo environments are singleton. Here, we modify the Swimmer-v2 environment to make it procedurally-generated. Specifically, we make the environment in each episode different by modifying the XML configuration file of MuJoCo engine in every new episode. We consider two environmental properties, i.e., density and viscosity. Density is used to simulate lift and drag forces, which scale quadratically with velocity. Viscosity is used to simulate viscous forces, which scale linearly with velocity. In the standard Swimmer-v2 environment, the density and the viscosity are fixed to 4000 and 0.1, respectively. In the first experiment, we uniformly sample the density in $[2000,4000]$ in each new episode to make it procedurally-generated, denoted as Density-Swimmer. Similarly, in the second environment, we uniformly sample the velocity in $[0.1,0.5]$, denoted as Velocity-Swimmer. These two variants make RL training more challenging since the agent needs to learn a policy that can generalize to different densities and velocities. Similarly, we accumulate the rewards to the final timestep of an episode to make the environments sparse. Both environments are provided in our code for reproducibility. The results are reported in Figure~\ref{fig:swimmer}. We observe that all methods learn slower on these two variants. For example, RAPID only achieves around 190 return in Density-Swimmer, while RAPID achieves more than 200 return in Swimmer. Similarly, SIL reaches around 100 return after around $5 \times 10^6$ timesteps in Density-Swimmer, while it achieves around 100 return after around $3 \times 10^6$ timesteps in Swimmer. Nevertheless, RAPID can discover very good policies even in these challenging procedurally-generated settings.

\newpage
\section{Discussions of Annealing}
\label{sec:J}
\begin{figure}[h!]
  \centering
  \begin{subfigure}[b]{0.5\textwidth}
    \centering
    \includegraphics[width=0.99\textwidth]{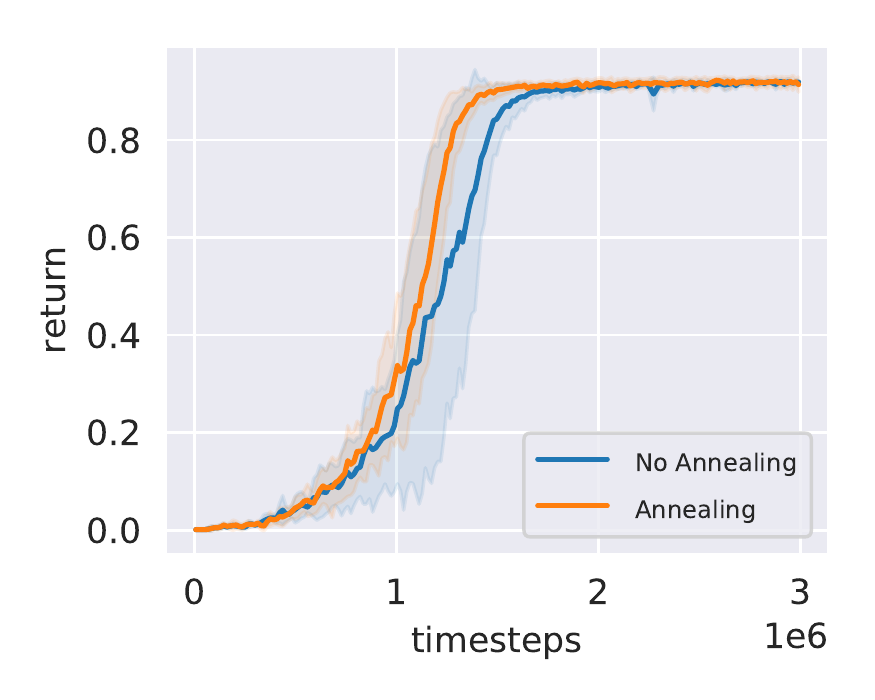}
    \caption{KeyCorridor-S3-R2}
  \end{subfigure}%
  \begin{subfigure}[b]{0.5\textwidth}
    \centering
    \includegraphics[width=0.99\textwidth]{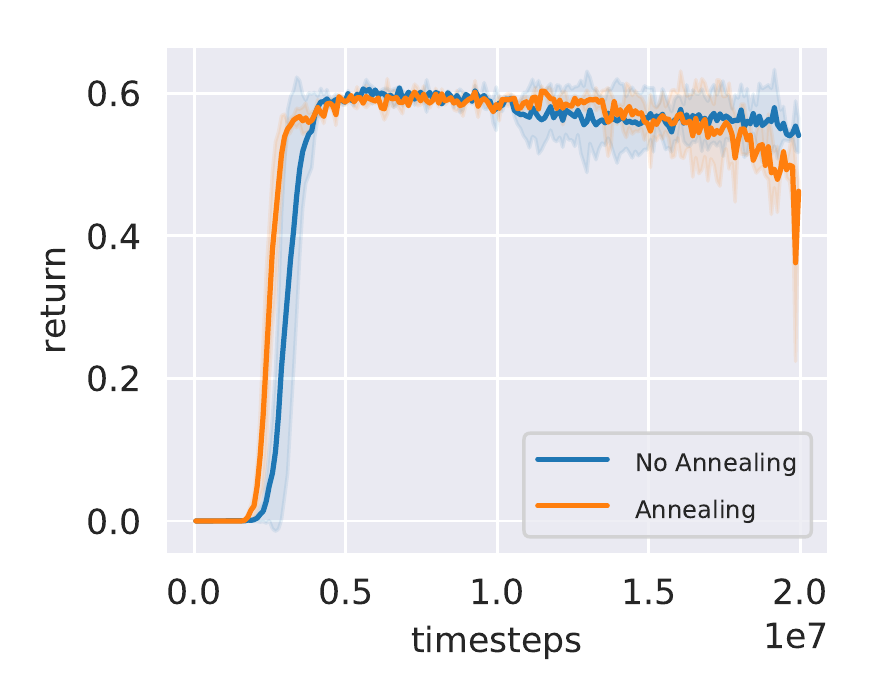}
    \caption{MultiRoom-N12-S10}
  \end{subfigure}%
  \caption{Annealing versus not annealing. Annealing is helpful in KeyCorridor-S3-R2. However, the policy fails to converge with annealing in MultiRoom-N12-S10.}
  \label{fig:annealing}
\end{figure}

In our experiments, we find that annealing the updating step to $0$ sometimes helps improve the sample efficiency. However, in some environments, annealing will have a negative effect on the final performance. To show the effect of annealing, we plot learning curves with or without annealing on MiniGrid-KeyCorridorS3R2-v0 and MiniGrid-MultiRoom-N12-S10-v0 in Figure~\ref{fig:annealing}. We observe that annealing can help in KeyCorridor-S3-R2. The possible explanation is that, after the agent navigates the goal, it has access to the reward signal. the PPO objective is sufficient to train a good policy with the extrinsic rewards. The imitation loss may harm the performance since it could be better to perform exploitation rather than exploration at this stage. However, in MultiRoom-N12-S10, annealing will harm the performance. In particular, the performance drops significantly in the final stage when the updating frequency of imitation loss approaches zero. The possible reason is that MultiRoom-N12-S10 needs very deep exploration to find the goal. Even though the PPO agent can navigate the goal, it may easily suffer from insufficient exploration when it encounters a new environment. As a result, the agent needs to keep exploring the environment in order to generalize. Introducing imitation loss will ensure that the agent keeps exploring the environments.

\end{document}